\documentclass{article}

\PassOptionsToPackage{numbers, compress}{natbib}

\usepackage[preprint]{neurips_2026}
\usepackage{titlesec}
\titlespacing*{\subsection}{0pt}{1.2ex plus 0.3ex minus 0.2ex}{0.6ex plus 0.2ex}
\titlespacing*{\paragraph}{0pt}{1.0ex plus 0.2ex minus 0.1ex}{0.5em}

\usepackage[utf8]{inputenc} 
\usepackage[T1]{fontenc}    
\usepackage{hyperref}       
\usepackage{url}            
\usepackage{booktabs}       
\usepackage{amsfonts}       
\usepackage{nicefrac}       
\usepackage[protrusion=false]{microtype} 
\usepackage{xcolor}         

\usepackage{bm}
\usepackage{makecell}
\usepackage{longtable}
\usepackage{graphicx}
\usepackage{subcaption}
\usepackage{amsmath}
\usepackage{amssymb}
\usepackage{algorithm}
\usepackage{algpseudocode}
\usepackage{xspace}
\usepackage{wrapfig}        
\usepackage{svg}            

\title{Dynamic Quadtree Tokenization and Transformer for Adaptive Mesh PDE Forecasting}

\author{%
Yilin Zhuang \quad Noah Zambrano \quad Karthik Duraisamy\thanks{Corresponding author: \texttt{kdur@umich.edu}}\\
Department of Aerospace Engineering\\
Michigan Institute for Computational Discovery and Engineering\\
University of Michigan, Ann Arbor, MI\\
\texttt{\{ylzhuang, nzamb, kdur\}@umich.edu}
}

\begin{document}

\maketitle

\begin{abstract}
The quadratic attention cost of Vision Transformers (ViTs) forces a trade-off between spatial resolution and rollout horizon, particularly for fine-scale PDEs where shocks, reaction fronts, and material interfaces occupy small, evolving regions of the domain. Conventional neural surrogates also lack mechanisms to adapt resolution dynamically. We propose WAMRViT, a ViT that tokenizes inputs as balanced quadtrees using a wavelet-inspired refinement criterion, jointly encodes position and refinement level with 3D rotary positional embeddings, and regrids in cell space during inference for stable long-horizon rollouts. A multi-scale variant retains each leaf at its native source resolution and lets the model learn across resolution levels. Unlike fixed-budget adaptive-tokenization methods, WAMRViT imposes no predetermined token count and supports fully adaptive topology throughout autoregressive rollout. To our knowledge, it is the first machine-learning surrogate to natively tokenize multi-level Adaptive Mesh Refinement (AMR) data. On uniform-grid benchmarks, uniform-patch WAMRViT improves finest-level region-of-interest VRMSE over a finest-patch uniform ViT while using substantially fewer tokens. The multi-scale variant achieves the lowest first-step full-field RMSE and VRMSE on both benchmarks and improves rollout-averaged full-field and refined-region accuracy at long horizons. With parallelized regridding, its end-to-end rollout cost lies between finest-patch and approximately token-matched coarser-patch ViTs. On a complex AMR combustion problem whose finest features cannot be represented natively by the evaluated uniform-grid baselines, WAMRViT operates on adaptive cells built from the simulator hierarchy and substantially reduces finest-level error at matched transformer capacity. Code: \url{https://github.com/tonyzyl/wamrvit}.
\end{abstract}

\section{Introduction}
Deep-learning surrogates have become a practical alternative to classical PDE solvers in computational fluid dynamics, combustion, and shock hydrodynamics \citep{raissi2019pinn,lu2021deeponet,li2021fno,alkin2024upt}. Among these, Vision Transformers (ViTs) are attractive because self-attention directly represents long-range correlations that can be difficult for local convolutional or graph operators to capture \citep{kolesnikov2021vit}. The price is quadratic cost in the number of tokens. This cost can impose an accuracy ceiling: on standard vision benchmarks, ViTs improve as patch size shrinks and, for a fixed backbone, can perform best when individual pixels become tokens \citep{nguyen2024pixels}. For fine-scale PDE forecasts, a finest-grid ViT can exceed practical compute and memory budgets as spatial resolution and rollout horizon grow.

This bottleneck is not inherent. In the complex PDE regimes where transformer surrogates are most valuable, such as detonation fronts and multi-material shocks, the dynamics are spatially sparse, and a uniform tokenization strategy squanders the $\mathcal{O}(N^2)$ self-attention budget on dynamically quiescent regions where the physical fields are largely homogeneous. Classical numerical analysis addresses this mismatch via Adaptive Mesh Refinement (AMR) \citep{berger1984amr,berger1989amr}: the solver commits compute to where it is needed and coarsens elsewhere, producing a hierarchical mesh.

Adaptive-patching ViTs already exploit spatially varying resolution \citep{xu2025amrtransformer, zhang2024apf, choudhury2025apt, zhang2024matey, wang2026iap}. Bringing this idea closer to AMR practice calls for combining three capabilities: (i) a recursive, neighbor-balanced hierarchy with mean-conservative coarsening and explicit inter-scale communication; (ii) direct ingestion of simulator-native AMR data; and (iii) repeated topology updates and state transfer during autoregressive forecasting, where errors compound and active regions move.

\paragraph{Contributions.}
We introduce \textbf{WAMRViT}, a transformer surrogate whose tokens are the leaves of a 2:1-balanced quadtree adapted to each input field. The hierarchy and token count evolve with predicted features, allowing computation to follow local structure throughout rollout. Our contributions address the three capabilities above.

\textbf{Hierarchical, balanced tokenization.} Our tokenizer employs a 2:1-balanced quadtree driven by a wavelet-inspired, non-learned local-range refinement criterion (Section~\ref{sec:tokenizer}). The tree supplies explicit face-neighbor relations and mean-conservative coarsening. Spatial positioning is encoded via a 3D rotary embedding \citep{su2024roformer} extended to the coordinates $(x, y, \ell)$. Crucially, the refinement level $\ell$ acts as a soft attention bias rather than a hard mask, allowing tokens at different levels to interact across coarse--fine boundaries where cross-level communication is necessary.

\textbf{Native AMR ingestion and resolution-preserving uniform-grid forecasting.} Uniform-patch WAMRViT reads a simulator's multi-level AMR hierarchy through a shared leaf interface. For uniform-grid inputs, the multi-scale variant retains each leaf's full source region and resolves inter-level resolution differences within the model through per-level patch embedders, resolution-specific decoders, and level-conditioned AdaLN. Both paths produce one token per leaf, supporting adaptive forecasting on native AMR hierarchies and uniform-grid fields.

\textbf{Long-horizon adaptive rollout.} To our knowledge, WAMRViT is the first machine-learning surrogate to perform autoregressive forecasting on a native multi-level AMR hierarchy while dynamically updating its topology and transferring predicted states between successive grids. Refinement and coarsening track moving features rather than freezing the grid, and the token count follows the evolving field rather than a prescribed budget. Parallel CPU-array regridding and a hybrid CPU/GPU path that minimize data exchange make this evolving-grid workflow practical.

\paragraph{Related work.}
The closest prior methods are adaptive-patching ViTs for PDE surrogates: AMR-Transformer \citep{xu2025amrtransformer}, MATEY \citep{zhang2024matey}, and IAP \citep{wang2026iap}. Overtone \citep{mukhopadhyay2025controllable} instead trains one ViT to operate at several uniform-grid patch sizes and cycles those sizes across rollout steps. Earlier adaptive neural-operator work targets chemical kinetics rather than the full flow field \citep{Zanardi_2023}. These methods do not evaluate natively multi-resolution AMR data under a moving autoregressive topology. MATEY's reported experiments use two-level partitioning and do not evaluate hierarchy depth. IAP releases only partial code, and we did not include a matched rollout implementation. WAMRViT addresses the resulting systems gap between adaptive tokenization and AMR practice: native multilevel input, balanced hierarchical topology, conservative inter-level transfer, variable token counts without a prescribed budget, and inference-time regridding during autoregressive prediction. Appendix~\ref{app:related-work} provides a comprehensive comparison.

\section{Method}
\label{sec:method}

\begin{figure}[!ht]
\centering
\includegraphics[width=\linewidth]{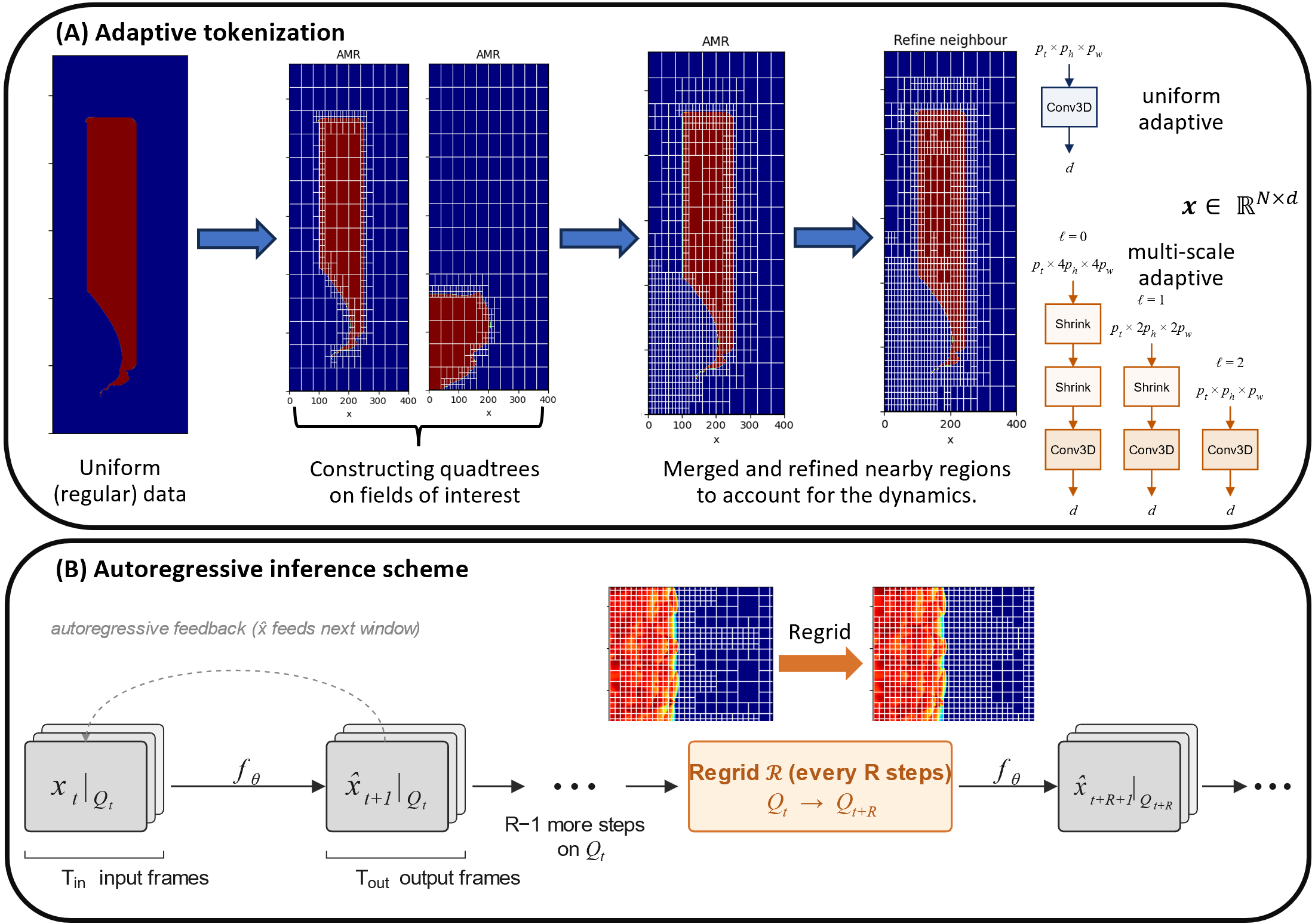}
\caption{\textbf{Overview of WAMRViT.} (a) Adaptive tokenization: the input field is partitioned into a 2:1-balanced quadtree by a wavelet-inspired local-range criterion, with one token per leaf. (b) Autoregressive inference: every $R$ steps a refine-protect-coarsen update revises the quadtree in cell space so the topology tracks moving features.}
\label{fig:overview}
\end{figure}

Let $Q_t$ be the current 2:1-balanced quadtree with $N_t$ leaves, meaning face-neighboring leaves differ by at most one refinement level. In the default uniform-patch path, the $T_\text{in}$ input frames are represented on $Q_t$ as $x_{t-T_\text{in}+1:t}|_{Q_t} \in \mathbb{R}^{N_t \times C \times T_\text{in} \times p_h \times p_w}$, where each leaf carries a fixed-size spatial raster. A shared tubelet convolution~\citep{arnab2021vivit} with temporal patch size $p_t = T_\text{in}$ spans the temporal window and leaf raster to produce one $d$-dimensional token per leaf. The transformer processes this variable-length sequence using each leaf's center and level $(x,y,\ell)$ through 3D Rotary Positional Embeddings (RoPE)~\citep{su2024roformer}, and a shared output projection decodes every token into $T_\text{out}$ predicted frames on the same leaves.

A model forward pass advances the field while leaving the topology unchanged. Prediction and periodic topology update are distinct operations:
\begin{equation}
\begin{aligned}
   \hat x_{t+1:t+T_\text{out}}|_{Q_t}
   &= f_\theta\!\left(x_{t-T_\text{in}+1:t}|_{Q_t}\right), \\
   \left(Q_{t+R},\, \hat x|_{Q_{t+R}}\right)
   &= \mathcal{R}\!\left(Q_t,\, \hat x|_{Q_t}\right)
      \quad \text{every $R$ rollout steps}.
\end{aligned}
\label{eq:operator}
\end{equation}
where, in the regridding relation, $\hat x$ denotes the state retained after advancing $R$ rollout steps on $Q_t$, including the history needed for the next input window, and $Q_{t+R}$ is the topology updated at that boundary. Regridding transfers this state without advancing time. Section~\ref{sec:rollout} summarizes the refine-protect-coarsen operator $\mathcal{R}$, and Appendix~\ref{app:regrid} gives the complete algorithm.

\subsection{Adaptive Tokenization and Positional Encoding}
\label{sec:tokenizer}
\label{sec:rope}

The spatial domain is tiled by root cells that may split recursively into four children up to level $L$. For uniform-grid inputs, a leaf is marked for refinement when the local range of any selected adaptation channel exceeds $\varepsilon$ times that channel's global range. Per-channel decisions are unioned. This non-learned local-extremum criterion bounds localized Haar-wavelet detail~\citep{haar1909theorie}. Native adaptive inputs instead start $Q_t$ from the simulator's refinement hierarchy, capped at level $L$, and then adjust it on the selected adaptation channels with the regridding procedure of Section~\ref{sec:rollout} (heat release rate for AMReX, Appendix~\ref{app:amrex}). After either construction path, 2:1 balance is restored by refining any leaf whose face neighbor is more than one level finer. The uniform-patch tokenizer stores a fixed $(C,p_h,p_w)$ raster per leaf and frame, obtained by exact block averaging of uniform-grid data or by sampling the finest native level covering each pixel.

Each training window shares one topology, built from its last input frame and used for all input and target frames, because $f_\theta$ predicts on the leaves of its input partition (Eq.~\eqref{eq:operator}). Building it from the newest observed frame matches rollout, where regridding uses the newest predicted frame. For native adaptive inputs, this requirement rules out the simulator's own grids, which change between snapshots. Because the partition stays fixed while features move, refined regions are dilated by $\delta$ face-neighbor layers, pre-allocating fine leaves into which nearby features can move. Deterministic training topologies can be cached per window and reused across epochs, with lazy value loading outside the topology cache (Appendix~\ref{app:train-tokenize}). Appendix~\ref{app:quadtree} gives the geometry, balancing procedure, and value-assignment rules.

Each token is associated with its leaf center $(x,y)$ and refinement level $\ell$. Our axial 3D RoPE assigns disjoint query/key dimensions to the three coordinates and applies an independent rotary embedding on each axis. The spatial scales convert normalized centers to finest-cell index units, matching the positional convention of the regular-grid ViT controls. We ablate the level-axis scale $s_\ell$ to evaluate how the strength of refinement-level encoding affects forecasting accuracy. Appendix~\ref{app:rope-config} gives the dimension partition, frequencies, scale formulas, and geometric interpretation.

\subsection{Transformer Backbone}
\label{sec:backbone}

\begin{wrapfigure}{r}{0.40\textwidth}
\centering
\includegraphics[width=\linewidth]{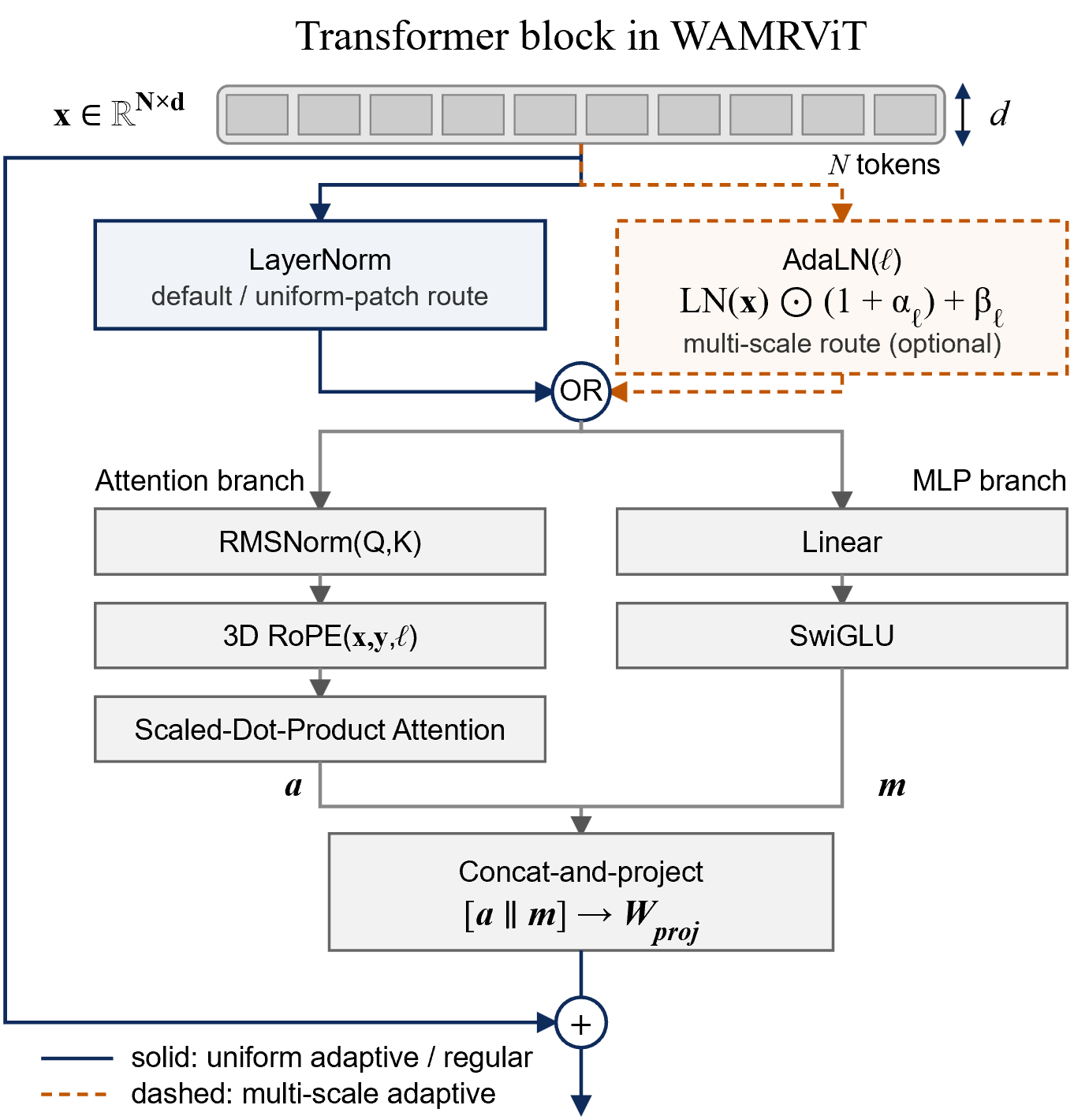}
\caption{\textbf{Parallel-layout transformer block.} Attention and MLP branches are concatenated and projected into the residual stream.}
\label{fig:transformer-block}
\end{wrapfigure}
Each parallel-layout block~\citep{dehghani2023vit22b} applies a single pre-LayerNorm and feeds the normalized tokens to an attention branch and a SwiGLU MLP branch. The branch outputs are concatenated, linearly projected, and added to the residual stream. Figure~\ref{fig:transformer-block} illustrates this architecture. Default WAMRViT and ViT-finest/mid use the same block stack, holding the transformer backbone fixed across adaptive and regular-grid representations.

For uniform-grid inputs, multi-scale WAMRViT retains each leaf's full source region rather than block-averaging it before the model. Coarse-region detail is reduced inside a learned encoder rather than discarded during data fill, while attention still processes one token per leaf. A level-$\ell$ leaf carries a raster of shape $(C,T_\text{in},p_h r_\ell,p_w r_\ell)$, where $r_\ell=2^{L-\ell}$. Its per-level encoder applies $L-\ell$ learned stride-$2$ reduction stages followed by a tubelet projection, producing one $d$-dimensional token regardless of the source raster size. Per-level decoders restore predictions at the retained resolution through corresponding expansion stages. Level-conditioned AdaLN replaces the per-block pre-norm and post-stack output norm, applying learned scale and shift parameters indexed by refinement level. The decoded leaves are reassembled on the finest uniform grid for the full-field objective below. These components define a separate architectural variant. Appendices~\ref{app:multiscale-method} and~\ref{app:multiscale-arch} give the complete encoder, decoder, and normalization specifications.

\subsection{Training Objectives}
\label{sec:objectives}

For uniform-patch WAMRViT, we train $f_\theta$ with the absolute $L_2$ norm of the prediction residual, averaged over leaves and channels,
\begin{equation}
   \ell(\hat y, y) = \frac{1}{N_t C}\sum_{i=1}^{N_t}\sum_{c=1}^{C}
   \left\|\hat y_{i,c}-y_{i,c}\right\|_2,
   \label{eq:loss}
\end{equation}
where $i$ and $c$ index leaves and channels, respectively, and the norm is taken over the $T_\text{out}\times p_h\times p_w$ spatiotemporal raster of each leaf. Leaf losses are weighted equally, regardless of the physical areas they represent.

Let $\hat y,y\in\mathbb{R}^{C\times T_\text{out}\times H\times W}$ denote full-field predictions and targets. Multi-scale WAMRViT uses a differentiable scatter onto the finest source grid. Regular-grid baselines use their respective output grids. The channel-mean objective takes each norm over the full spatiotemporal grid:
\begin{equation}
   \ell_{\mathrm{full}}(\hat y,y)
   = \frac{1}{C}\sum_{c=1}^{C}
   \left\|\hat y_c-y_c\right\|_2.
   \label{eq:loss-dense}
\end{equation}

Push-forward training unrolls $K$ model applications and averages their single-step losses, $\mathcal{L}(\theta)=K^{-1}\sum_{k=1}^{K}\ell(\hat y^{(k)},y^{(k)})$~\citep{brandstetter2022pf}. The full-field variants substitute $\ell_{\mathrm{full}}$ for $\ell$. Appendix~\ref{app:training-specs} gives the numerical stabilizer and complete push-forward recursion.

\subsection{Autoregressive Rollout and In-Cell-Space Regridding}
\label{sec:rollout}

Every $R$ predicted steps, regridding updates the topology from the newest predicted frame and transfers the entire input window onto it. The model calls in between keep the topology fixed. Because the source grid is unavailable during rollout, the refinement tests operate on leaf rasters. Uniform-patch regridding has three phases. (1) Refine leaves whose within-cell variation or face-boundary jump exceeds the threshold, bilinearly populate their children, and restore 2:1 balance. The jump test detects features between internally smooth leaves, which within-cell variation alone would miss. (2) Protect active leaves and a buffer of $\delta$ face-neighbor layers from immediate coarsening. This buffer keeps moving features on refined leaves, so the topology can stay fixed for $R$ steps instead of being regridded after every prediction. (3) Coarsen four unprotected siblings into their parent by exact $2{\times}2$ block averaging, but only when their variation is below the hysteresis threshold and a look-ahead confirms that merging preserves 2:1 balance. Multi-scale native storage instead partitions and concatenates the retained source rasters exactly (Appendix~\ref{app:multiscale-method}). Stopping conditions and complete pseudocode are given in Appendix~\ref{app:regrid}. The array backend parallelizes topology updates over packed leaf metadata using CPU kernels. For PLI and TRL, the small input windows are transferred to the CPU for regridding. For AMReX, the larger input windows remain on the GPU. Only the adaptation channels of the newest frame are transferred to the CPU for topology decisions, and compact split/merge records are returned to the GPU for state updates.

\section{Experiments}
\label{sec:experiments}

\subsection{Problems}
\label{sec:datasets}

We evaluate uniform-grid fields with two contrasting spatial distributions and a simulator-native AMR hierarchy. Appendix~\ref{app:qualitative-overview} provides a visual overview of all three problems.

\paragraph{PLI multimaterial shock propagation.}
The PLI subset of the HEAT benchmark~\citep{banesh2025heat} contains cylindrically symmetric shocks moving through perturbed material interfaces and ambient air. Its $10$-channel observations comprise $1000$ trajectories of $101$ snapshots on a $1120\times400$ uniform grid. Large quiescent regions surround localized shocks and interfaces, making PLI the clearest test of whether adaptive tokenization can preserve refined-region accuracy without paying for finest patches everywhere.

\paragraph{TRL radiative mixing.}
TRL is from The Well~\citep{ohana2024well} with $90$ trajectories of $101$ snapshots, $4$ channels, and a $128\times384$ uniform grid. Its cold--hot interface rolls into dense filaments under shear, leaving substantially less quiescent area than PLI. TRL therefore tests adaptivity where a regular grid is less obviously wasteful and where the available training set is smaller.

\paragraph{AMReX combustion.}
The native-AMR case follows high-fidelity H$_2$/air detonation simulations with detailed finite-rate chemistry~\citep{raman2025combustion}. Its $17$-channel, $396$-snapshot trajectory occupies $264$ GB and contains six coexisting refinement levels concentrated around the moving shock and reaction zone. To fit training memory, WAMRViT uses the five coarsest ($L=4$, finest grid $4096\times24576$). A dense finest-grid representation is not stored. WAMRViT instead builds its quadtree from each reference snapshot's AMReX hierarchy and samples leaf values directly from the AMReX data. Shock discontinuities, stiff chemistry, and solver-sensitive soot-foil morphology make forecasting this hierarchy a demanding test of adaptive surrogates. Full channel lists and provenance are in Appendix~\ref{app:datasets}. Representative full-domain PLI and AMReX predictions are in Figures~\ref{fig:app-pli-qualitative} and~\ref{fig:app-amrex-qualitative}.

\subsection{Baselines}
\label{sec:baselines}

ViT-finest uses the default WAMRViT parallel block stack with the finest patch size on a regular grid, providing a backbone-matched regular-grid baseline. ViT-mid uses the same stack with a coarser patch size chosen to approximately match WAMRViT's token count. Multi-scale WAMRViT is evaluated separately as a resolution-preserving architectural variant. On the native-AMR combustion problem, a finest-grid ViT would require approximately $98$k tokens, while the validated multi-scale training objective additionally materializes the finest uniform grid to compute its full-field loss. These finest-grid approaches are therefore impractical under the current training protocol. The feasible regular-grid baselines instead project the AMR hierarchy onto a coarser grid, yet still use about $14\%$ more tokens than uniform WAMRViT on average over the accuracy rollouts.

SwinV2~\citep{liu2022swinv2} provides an architecture-family control with conventional shifted-window blocks and a refined multiscale decoder. We approximately match its transformer capacity to WAMRViT, rather than its total parameter count or computational cost. This comparison tests an alternative architecture family, not a complete PDE foundation-model system. All models share the data splits and per-dataset optimization protocol. Full architectural specifications are provided in Appendix~\ref{app:training-specs}.

\subsection{Training and Evaluation}
\label{sec:training}

Uniform-patch WAMRViT and the two ViT controls share a six-layer backbone with $12$ attention heads of dimension $d_h=192$ ($d=2304$), held fixed across datasets. Models predict one frame from two input frames on PLI/AMReX. TRL uses four inputs, following The Well's reference protocol~\citep{ohana2024well}. All variants use AdamW with a cosine learning-rate schedule, BF16 mixed precision, gradient clipping, EMA weight averaging, and $L_2$ prediction-error objectives, evaluated per leaf for uniform-patch WAMRViT and over the full field for the multi-scale variant and regular-grid baselines. We denote the push-forward training depth by $K$: training unrolls $K$ model applications and averages their per-step losses, with $K=1$ for PLI and TRL and $K=4$ for AMReX. To accommodate variable token counts, adaptive runs use micro-batch-one training. Section~\ref{sec:objectives} defines the per-leaf and full-field losses. Appendix~\ref{app:training-specs} gives the full hyperparameters.

PLI uses $800$ training and $200$ test trajectories; TRL uses The Well's official train/test split~\citep{ohana2024well}; AMReX uses the first $80\%$ of its trajectory for training and the final $20\%$ for chronological testing. We use no separate validation set. Per-dataset adaptation settings, including initial refinement and regrid tolerances and protection radii, are selected qualitatively to align refinement with the physically active regions while keeping the training and inference topology distributions consistent. Their values are reported in Appendix~\ref{app:training-specs}.

Evaluation uses shared held-out windows: PLI is evaluated on sparsely sampled $20$-step rollouts, and AMReX on four $20$-step rollouts from the chronological test segment, with WAMRViT regridding every $R=4$ prediction steps in both cases. TRL instead uses one-step prediction over all test windows because its snapshot interval makes longer autoregressive comparisons uninformative. No inference-time regridding is performed in this setting. We use ``rollout step'' to denote the evaluation horizon. Main accuracy comparisons use three training seeds. Table~\ref{tab:perlevel} reports first-step means and sample standard deviations. Full-field metrics compare predictions on the common finest uniform grid and are available for PLI and TRL. Cell-space metrics use regular predictions mapped onto paired adaptive leaves by block averaging or nearest-neighbor upsampling. ``Finest-level'' means $\ell=L$. Because regridding changes that leaf set, later finest-level values describe the currently refined population rather than a fixed set of pixels. Within either scope, RMSE and variance-normalized RMSE (VRMSE) are computed per channel and then averaged over channels. VRMSE normalizes with the ground-truth spatial variance of the full frame or individual leaf, respectively. Metric formulas are given in Appendix~\ref{app:metrics}.

\subsection{Results}
\label{sec:results}

\begin{figure}[!htb]
\centering
\begin{subfigure}{\linewidth}
\centering
\includegraphics[width=\linewidth]{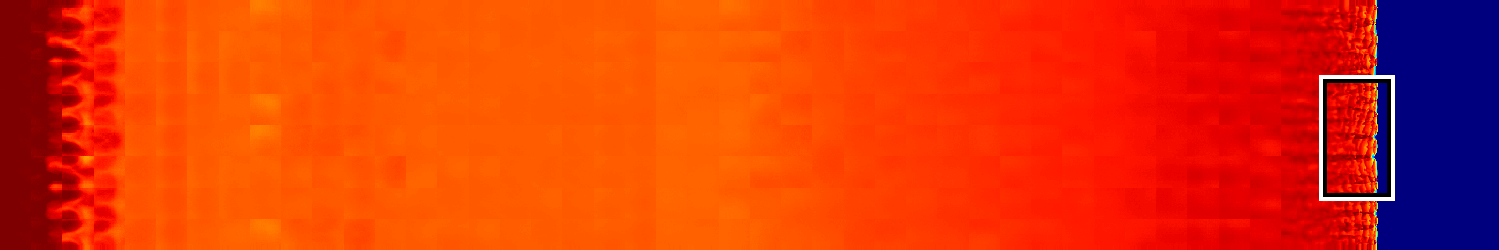}
\caption{Full-domain temperature prediction.}
\label{fig:amrex-qualitative-overview}
\end{subfigure}

\vspace{0.4em}
\begin{subfigure}{\linewidth}
\centering
\includegraphics[width=\linewidth]{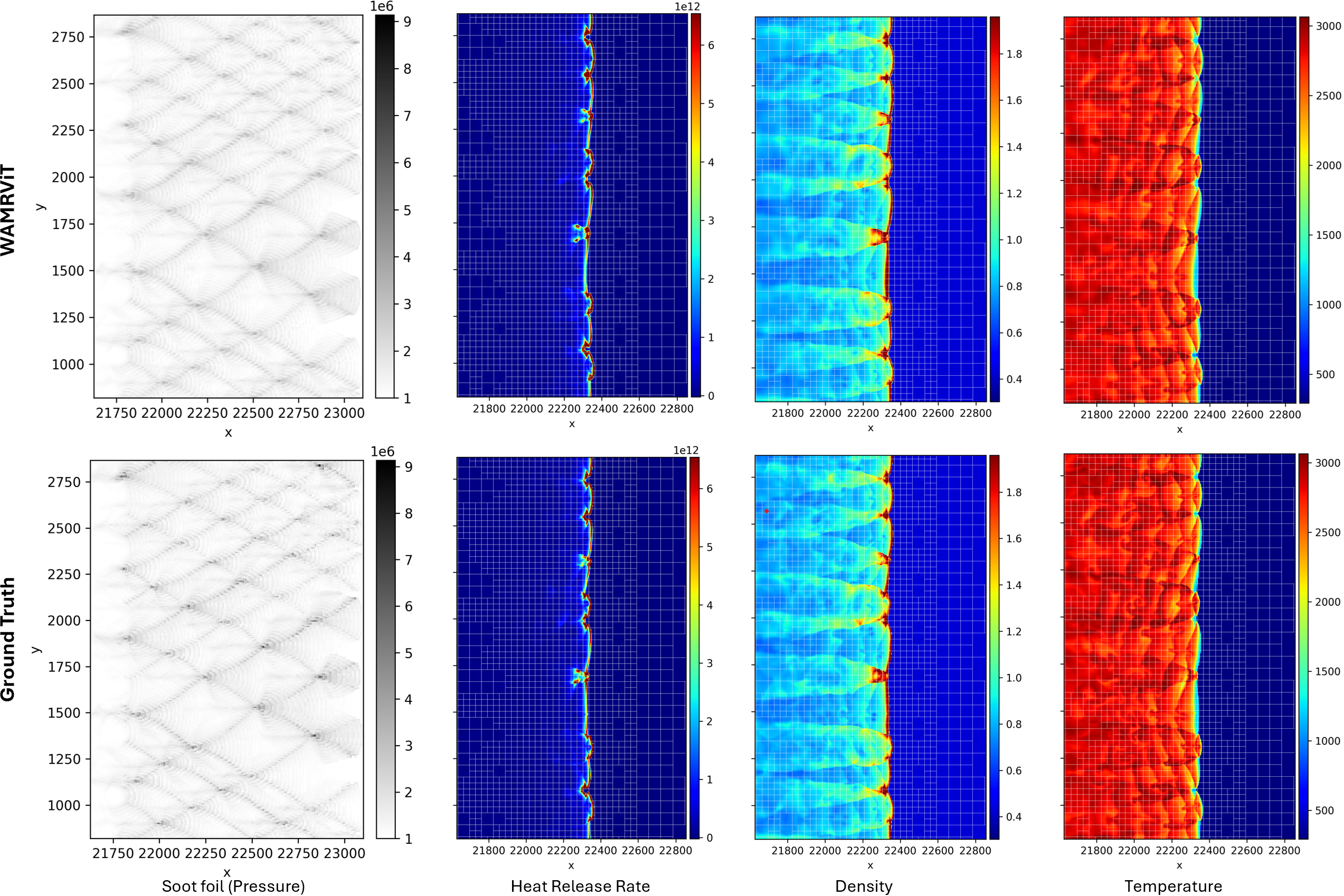}
\caption{Zoomed qualitative examples.}
\label{fig:amrex-qualitative-detail}
\end{subfigure}
\caption{\textbf{AMReX combustion prediction on the chronological holdout.} (a) Full-domain WAMRViT prediction at rollout step $25$. The white box marks the spatial extent of the zoomed field examples. (b) Predictions (top row) and ground truth (bottom row): peak-pressure soot foils over $50$ rollout steps (left), and heat-release rate, density, and temperature at step $25$ (right). These longer qualitative diagnostics extend beyond the $20$-step quantitative protocol. Ground-truth grids are shown for reference. Additional full-domain comparisons are shown in Appendix~\ref{subsec:qualitative_rollout}. Animations are provided in the \href{https://github.com/tonyzyl/wamrvit/tree/main/assets/animations}{repository}.}
\label{fig:amrex-qualitative}
\end{figure}%

\begin{table}[!htb]
\centering
\caption{\textbf{First-step VRMSE and rollout-mean token counts.} Accuracy is mean $\pm$ sample standard deviation over seeds $\{0,1,42\}$. Token counts $\bar N$ are averaged over evaluation windows, rollout steps ($20$ for PLI/AMReX, $1$ for TRL), and seeds. SwinV2 reports first-stage tokens. Finest-level metrics use the regions refined to level $L$ in the initial adaptive topology. IAP$^{*}$~\citep{wang2026iap} reports performance at the two token budgets comparable to ours, $1{,}020$ and $2{,}040$, respectively.}
\label{tab:perlevel}
\scriptsize
\setlength{\tabcolsep}{2.6pt}
\begin{tabular}{@{}lcccccc@{}}
 \toprule
 & \makecell{WAMRViT\\(uniform)} & \makecell{WAMRViT\\(multi-scale)} & \makecell{ViT\\(finest)} & \makecell{ViT\\(mid)} & SwinV2 & IAP$^{*}$~\citep{wang2026iap} \\
 \midrule
 PLI, full-field & \makecell{0.0661\\{\scriptsize $\pm$1.67E-3}} & \makecell{\textbf{0.0389}\\{\scriptsize $\pm$2.52E-4}} & \makecell{\underline{0.0539}\\{\scriptsize $\pm$6.44E-3}} & \makecell{0.0553\\{\scriptsize $\pm$6.68E-4}} & \makecell{0.0920\\{\scriptsize $\pm$2.32E-4}} & / \\
 PLI, finest ($\ell\!=\!3$) & \makecell{\underline{2.6839}\\{\scriptsize $\pm$0.166}} & \makecell{\textbf{2.5000}\\{\scriptsize $\pm$0.046}} & \makecell{3.4125\\{\scriptsize $\pm$0.305}} & \makecell{3.3401\\{\scriptsize $\pm$0.121}} & \makecell{6.3348\\{\scriptsize $\pm$0.054}} & / \\
 PLI, tokens $\bar N$ & 3,327 & 3,339 & 8,960 & 3,500 & 8,960 & / \\
 \midrule
 TRL, full-field & \makecell{0.1502\\{\scriptsize $\pm$2.81E-4}} & \makecell{\textbf{0.1476}\\{\scriptsize $\pm$5.40E-4}} & \makecell{\underline{0.1490}\\{\scriptsize $\pm$6.61E-4}} & \makecell{0.1629\\{\scriptsize $\pm$4.87E-4}} & \makecell{0.1955\\{\scriptsize $\pm$6.46E-4}} & 0.1577 / 0.1478 \\
 TRL, finest ($\ell\!=\!2$) & \makecell{\textbf{9.0411}\\{\scriptsize $\pm$0.665}} & \makecell{18.8458\\{\scriptsize $\pm$0.699}} & \makecell{\underline{16.5135}\\{\scriptsize $\pm$0.351}} & \makecell{22.1965\\{\scriptsize $\pm$0.334}} & \makecell{69.8422\\{\scriptsize $\pm$0.503}} & / \\
 TRL, tokens $\bar N$ & 1,597 & 1,597 & 3,072 & 1,536 & 3,072 & 1,020 / 2,040 \\
 \midrule
 AMReX, finest ($\ell\!=\!4$) & \makecell{\textbf{1.581E10}\\{\scriptsize $\pm$4.87E8}} & / & / & \makecell{\underline{6.306E12}\\{\scriptsize $\pm$2.53E11}} & \makecell{7.779E12\\{\scriptsize $\pm$2.76E11}} & / \\
 AMReX, tokens $\bar N$ & 5,399 & / & / & 6,144 & 6,144 & / \\
 \bottomrule
 \end{tabular}
\end{table}

\begin{figure}[!htb]
\centering
\begin{subfigure}{0.48\linewidth}
\includegraphics[width=\linewidth]{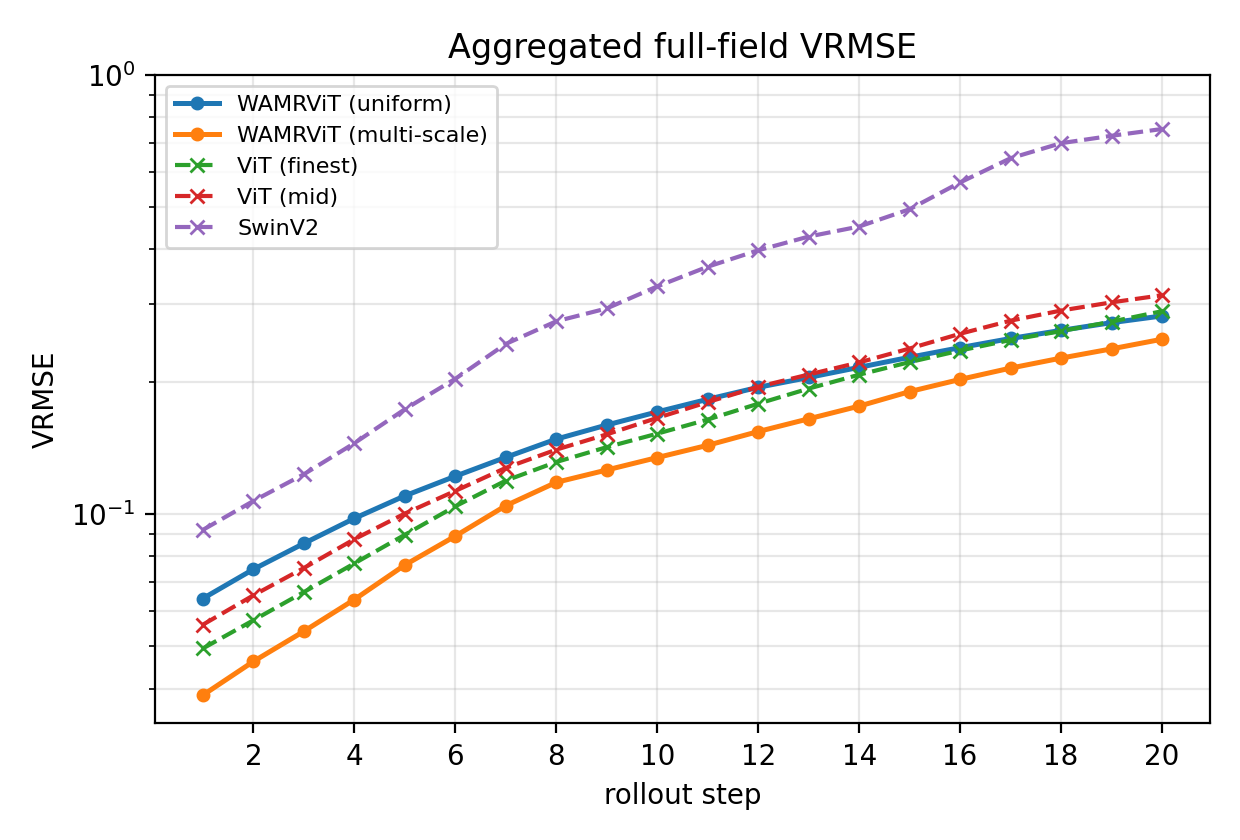}
\caption{PLI full-field VRMSE.}
\end{subfigure}\hfill
\begin{subfigure}{0.48\linewidth}
\includegraphics[width=\linewidth]{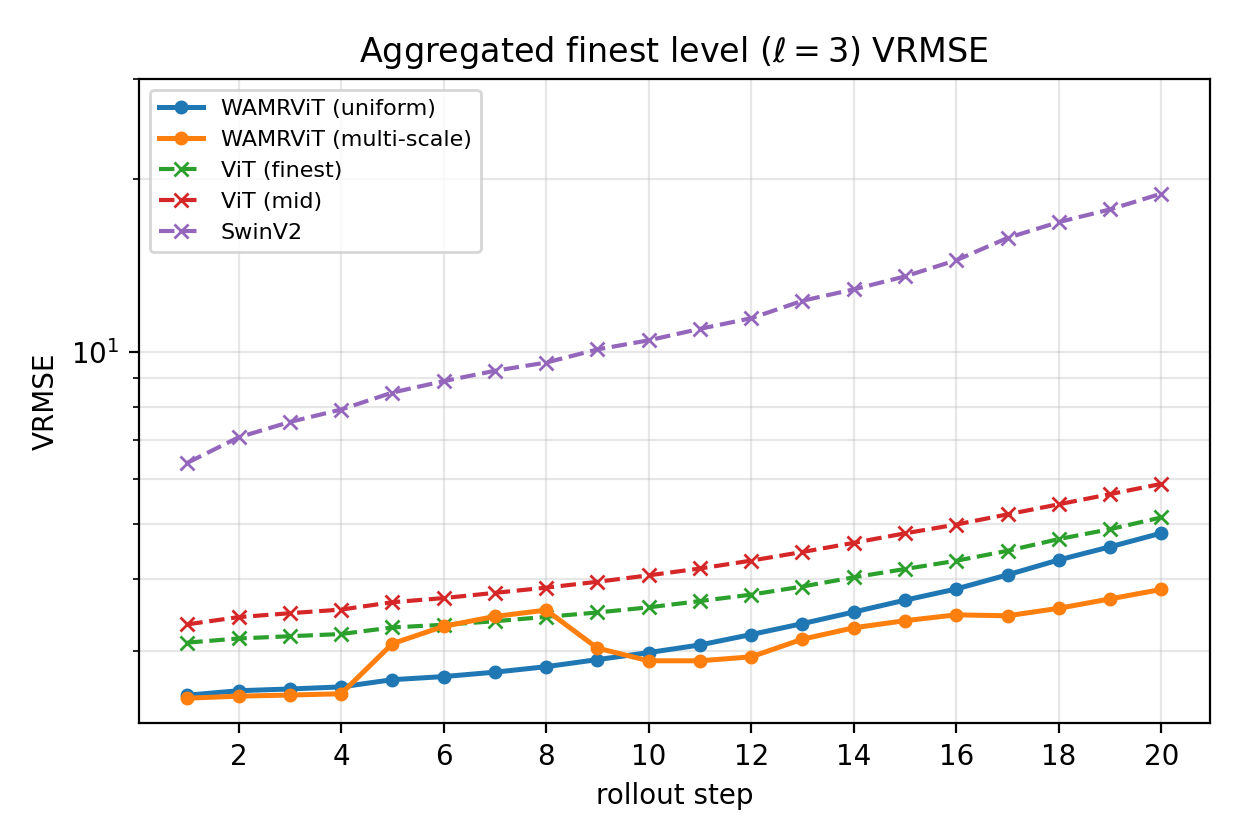}
\caption{PLI finest-level VRMSE.}
\end{subfigure}\\[0.4em]
\begin{subfigure}{0.48\linewidth}
\includegraphics[width=\linewidth]{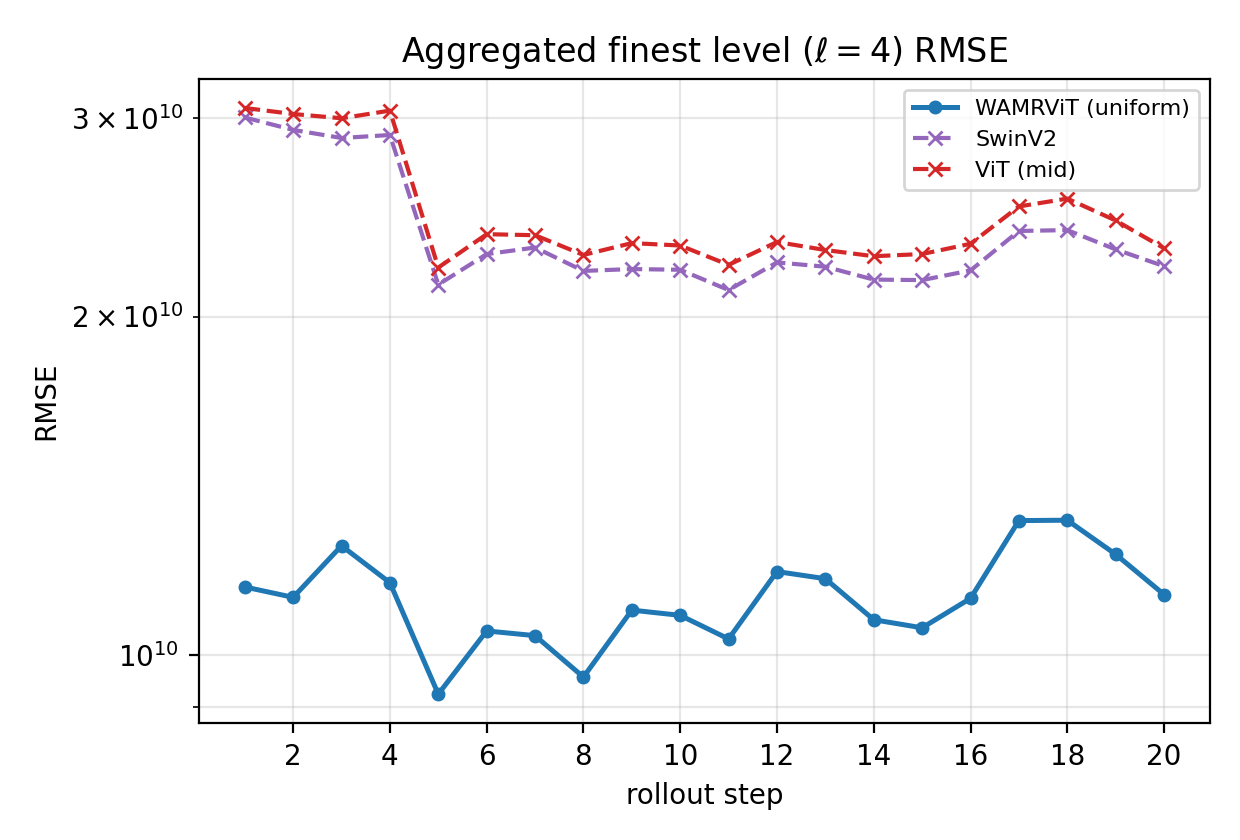}
\caption{AMReX finest-level RMSE.}
\end{subfigure}\hfill
\begin{subfigure}{0.48\linewidth}
\includegraphics[width=\linewidth]{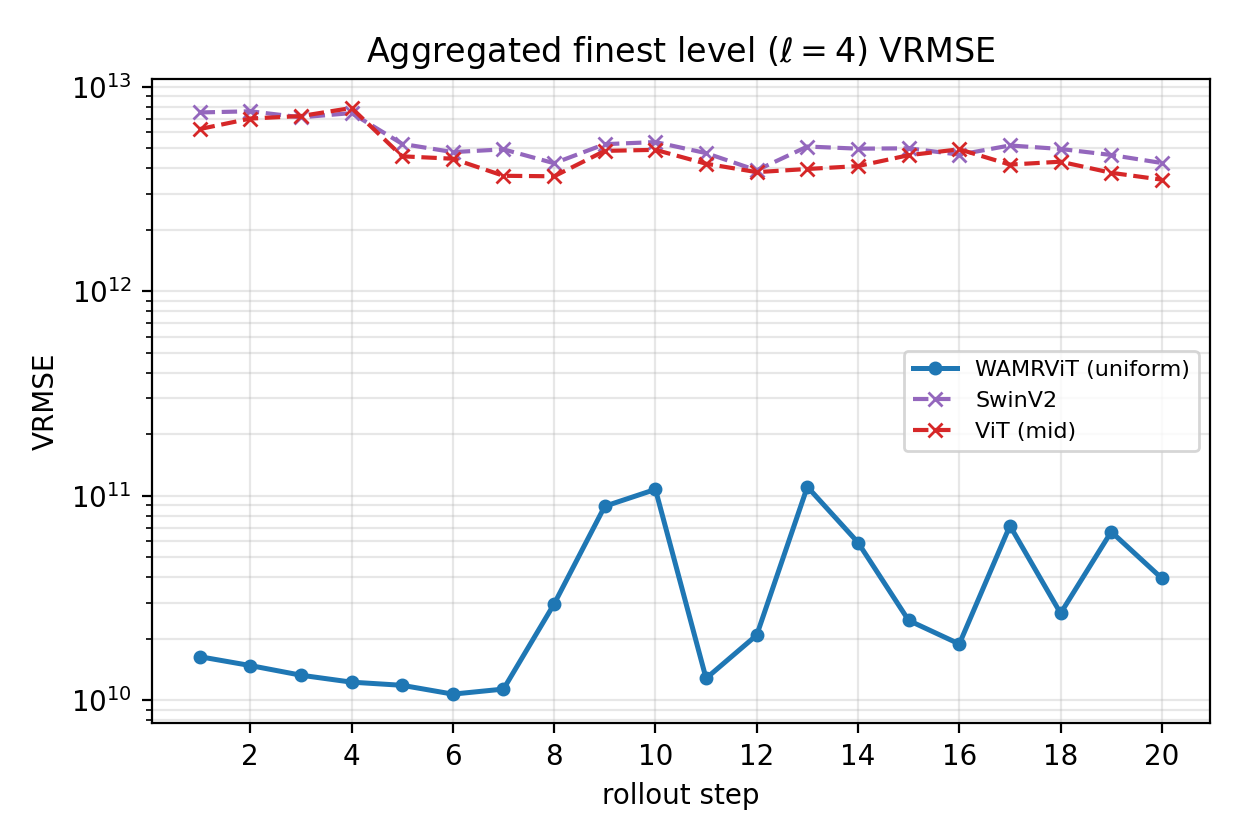}
\caption{AMReX finest-level VRMSE.}
\end{subfigure}
\caption{\textbf{Aggregated rollout error.} PLI full-field and finest-level VRMSE for $K=1$ models (top), and AMReX finest-level RMSE and VRMSE for $K=4$ models (bottom). WAMRViT regrids every four rollout steps. At the PLI finest level, the single ViT-finest curve is evaluated on uniform WAMRViT's refined region. Multi-scale WAMRViT uses its own region. AMReX errors are dominated by heat-release rate (HRR). Per-field results are reported in Appendix~\ref{app:perfield-amrex-fields}. Three-seed first-step statistics are given in Table~\ref{tab:perlevel}.}
\label{fig:rollout}
\end{figure}

\paragraph{PLI.}
At the first rollout step (Table~\ref{tab:perlevel}), multi-scale WAMRViT has the lowest VRMSE in both scopes, $27.7\%$ below ViT-finest over the full field. Uniform WAMRViT also outperforms ViT-finest in the finest-level region but trails it over the full field. This is consistent with its tokenizer block-averaging coarse leaves before the model, which discards sub-patch detail that full-field metrics still measure. Multi-scale WAMRViT retains that detail, reduces it in a learned encoder, and is trained with the full-field loss, at essentially the same token count as uniform WAMRViT, about $37\%$ of ViT-finest's. These advantages narrow but persist over the rollout. Averaged over all $20$ steps and seeds, uniform and multi-scale WAMRViT reduce finest-level VRMSE by $14.2\%$ and $20.1\%$ relative to ViT-finest evaluated on each variant's own finest-level region, and multi-scale WAMRViT reduces full-field VRMSE by $9.6\%$. RMSE counterparts and complete curves are in Appendix~\ref{app:additional_rollout}.

\paragraph{TRL.}
TRL is a one-step test with matched initial tolerances for the two adaptive variants and no inference-time regrid. Both variants use approximately half as many tokens as ViT-finest. Over the full field, both are within about $1\%$ of ViT-finest in VRMSE, with multi-scale WAMRViT slightly lower and uniform WAMRViT slightly higher. At the finest level, uniform WAMRViT reduces VRMSE by $45.3\%$ relative to ViT-finest, whereas multi-scale WAMRViT is $14.1\%$ higher. RMSE instead slightly favors multi-scale WAMRViT over ViT-finest in both scopes (Table~\ref{tab:perlevel-rmse}), illustrating sensitivity to local variance normalization. The published IAP~\citep{wang2026iap} results in Table~\ref{tab:perlevel} remain external references with a different training protocol. TRL's spatially widespread mixing leaves less quiescent area than PLI, reducing the opportunity to concentrate tokens on a small active region. Together with its coarse temporal sampling, this may contribute to multi-scale WAMRViT's smaller and metric-dependent gains over ViT-finest.

\paragraph{AMReX combustion.}
Only cell-space comparisons are feasible at the finest tokenized level ($\ell=4$). WAMRViT reduces first-step finest-level RMSE by $60.8\%$ relative to ViT-mid, with a similar margin over SwinV2. Averaged over seeds and all $20$ rollout steps, the reduction remains above $50\%$ against both. The per-leaf VRMSE aggregate is dominated by HRR, so we also report RMSE and the per-field breakdown in Appendix~\ref{app:perfield-amrex-fields}. Topology updates also change the evaluated leaf population, so fluctuations reflect both forecast evolution and changes in the set of refined cells. Figure~\ref{fig:amrex-qualitative} and Appendix~\ref{subsec:qualitative_rollout} show the dynamic grid following the reaction front and reproducing the soot-foil morphology. The full-domain views nevertheless reveal field deviations in coarse post-shock regions (Figure~\ref{fig:amrex-qualitative-overview} and Appendix Figure~\ref{fig:app-amrex-qualitative}). Uniform-patch WAMRViT weights leaf losses equally rather than by physical area. In the quantitative AMReX rollouts, the coarsest level covers about $84\%$ of the domain but accounts for only $6\%$ of the tokens. This imbalance may place less training emphasis on broad coarse regions than on the finely resolved reaction front. These are forecasts of held-out future states, testing the surrogate on coupled shock and reaction dynamics.

Table~\ref{tab:perlevel-rmse} reports first-step RMSE from the same three-seed evaluations. Multi-scale WAMRViT has the lowest full-field RMSE on PLI and TRL, and the AMReX finest-level advantage holds under both metrics. Finest-level rankings can differ between the two metrics because finest-level VRMSE divides each leaf's error by that leaf's ground-truth standard deviation before averaging (Appendix~\ref{app:metrics}). Nearly constant leaves thus carry large weight in VRMSE, whereas RMSE is dominated by the large absolute errors at high-contrast features. On PLI, for example, uniform WAMRViT's finest-level RMSE is comparable to ViT-finest's, while its VRMSE is lower.

\paragraph{End-to-end efficiency.}
Table~\ref{tab:walltime} reports training duration, timing-cohort token counts, and inference cost, including complete array-based topology updates and value transfers, with AMReX leaf values kept on the GPU (Section~\ref{sec:rollout}). Uniform/multi-scale WAMRViT is $3.03\times/2.57\times$ faster than ViT-finest on PLI and approximately $1.21\times$ faster on TRL. Regridding accounts for $5.4\%$--$17.2\%$ of adaptive inference time. On AMReX, it offsets the forward-pass advantage over ViT-mid, making total time $8.6\%$ higher. SwinV2 is fastest and uses the least peak allocated memory across all three datasets. Appendix~\ref{app:walltime} gives the timing protocol and regrid settings.

\begin{table}[!htb]
\centering
\caption{\textbf{Per-model training and inference cost.} Inference uses ten matched starting windows, batch one, and $20$ model calls on one H100, including four complete regrids for adaptive models. Forward and regrid are mean per-window sums. Peak is maximum PyTorch-allocated CUDA memory, and $\bar N$ averages tokens per call over the timing cohort. Train reports total minutes for the schedules in Appendix Table~\ref{tab:training-specs}. }
\label{tab:walltime}
\small
\setlength{\tabcolsep}{3pt}
\begin{tabular}{@{}llrrrrrr@{}}
\toprule
Dataset & Variant & \makecell{Train\\(min)} & $\bar N$ & \makecell{Forward\\(s)} & \makecell{Regrid\\(s)} & \makecell{Total\\(s)} & \makecell{Peak\\(GiB)} \\
\midrule
 PLI   & WAMRViT (uniform)        & 188 & 3601 & 1.454 & 0.126 & $1.580\pm0.328$ & 1.584 \\
       & WAMRViT (multi-scale)    & 333 & 3606 & 1.601 & 0.259 & $1.860\pm0.314$ & 1.802 \\
       & ViT-finest               & 236 & 8960 & 4.789 & -- & $4.789\pm0.008$ & 1.979 \\
       & ViT-mid                  & 128 & 3500 & 1.301 & -- & $1.301\pm0.003$ & 1.523 \\
       & SwinV2                   & 283 & 8960 & 0.469 & -- & $0.469\pm0.001$ & 1.343 \\
 \midrule
 TRL   & WAMRViT (uniform)        & 136 & 2111 & 0.774 & 0.159 & $0.933\pm0.095$ & 1.354 \\
       & WAMRViT (multi-scale)    & 223 & 2199 & 0.888 & 0.050 & $0.938\pm0.102$ & 1.384 \\
       & ViT-finest               & 155 & 3072 & 1.132 & -- & $1.132\pm0.002$ & 1.385 \\
       & ViT-mid                  & 129 & 1536 & 0.517 & -- & $0.517\pm0.003$ & 1.252 \\
       & SwinV2                   & 443 & 3072 & 0.453 & -- & $0.453\pm0.002$ & 1.180 \\
 \midrule
 AMReX & WAMRViT (uniform)        & 1371 & 5303 & 3.162 & 0.658 & $3.820\pm0.233$ & 6.094 \\
       & ViT-mid                  & 1805 & 6144 & 3.519 & -- & $3.519\pm0.006$ & 5.281 \\
       & SwinV2                   & 2086 & 6144 & 0.508 & -- & $0.508\pm0.002$ & 4.682 \\
 \bottomrule
\end{tabular}
\end{table}

\subsection{Ablations}
\label{sec:ablations}

\begin{table}[!htb]
\centering
\caption{\textbf{Level-axis RoPE ablation: first-step channel-mean VRMSE.} Uniform-patch WAMRViT on all three datasets; multi-scale WAMRViT on PLI. Training uses $K=1$ for PLI/TRL and $K=4$ for AMReX. Further results are in Appendix~\ref{app:level-scale-ablation}.}
\label{tab:level-scale-step1}
\small
\setlength{\tabcolsep}{3pt}
\begin{tabular}{@{}lcc|cc|cc|c@{}}
\toprule
 & \multicolumn{2}{c|}{\makecell{PLI\\uniform}} & \multicolumn{2}{c|}{\makecell{PLI\\multi-scale}} & \multicolumn{2}{c|}{\makecell{TRL\\uniform}} & \makecell{AMReX\\uniform} \\
\cmidrule(lr){2-3}\cmidrule(lr){4-5}\cmidrule(lr){6-7}\cmidrule(lr){8-8}
$s_\ell$ & full-field & finest ($\ell\!=\!3$) & full-field & finest ($\ell\!=\!3$) & full-field & finest ($\ell\!=\!2$) & finest ($\ell\!=\!4$) \\
\midrule
 $1$ (default) & 0.0642 & \underline{2.5107} & \textbf{0.03868} & \textbf{2.4804} & 0.1515 & 9.7444 & \underline{1.631E10} \\
 $2$ & \textbf{0.0630} & 2.5250 & 0.03889 & 2.7372 & \underline{0.1505} & \textbf{8.9478} & 1.805E10 \\
 $4$ & \underline{0.0632} & \textbf{2.5042} & \underline{0.03874} & \underline{2.5213} & 0.1506 & 9.8886 & \textbf{1.544E10} \\
 $8$ & 0.0666 & 2.5394 & 0.03967 & 2.6803 & \textbf{0.1502} & \underline{9.1519} & 1.660E10 \\
 \bottomrule
\end{tabular}
\end{table}

\paragraph{Push-forward depth.}
On PLI, WAMRViT and ViT-finest are each trained at $K\in\{1,2,4\}$ to test the effect of $K$ on rollout performance. At rollout step $20$, WAMRViT's finest-level VRMSE is below ViT-finest's at both $K=1$ and $K=4$, with the margin growing to $23.2\%$ at $K=4$. The intermediate $K=2$ adaptive run is unstable at this scope. At full field with $K=4$, WAMRViT is lower than ViT-finest at step $20$ but higher when averaged over the rollout. The effect of push-forward depth depends on the evaluation scope and horizon. At step $20$, WAMRViT's finest-level advantage holds for $K=1$ and $K=4$, but reverses at $K=2$. Complete curves and values are in Appendix~\ref{app:pf-ablation}.

\paragraph{Topology resampling.}
We compare uniform WAMRViT trained with and without random coarsening/refinement on PLI and AMReX. On PLI, resampling degrades both scopes: averaged over the $20$-step rollout, VRMSE rises by $18.5\%$ at full field and $32.2\%$ at the finest level. On AMReX, over the same horizon, it lowers finest-level VRMSE by $8.8\%$ but raises coarsest-level VRMSE by $15.7\%$. Its effect is therefore dataset-dependent, and all Table~\ref{tab:perlevel} models train without it (Appendix~\ref{app:resample-ablation}).

\paragraph{Level-axis scale.}
Table~\ref{tab:level-scale-step1} compares $s_\ell\in\{1,2,4,8\}$ for uniform-patch and multi-scale WAMRViT on PLI and uniform-patch WAMRViT on TRL and AMReX. For uniform-patch models, the preferred scale varies with dataset, scope, metric, and horizon; neither $s_\ell=2$ (perimeter ratio) nor $s_\ell=4$ (area ratio) is consistently best. On AMReX, $s_\ell=4$ leads most VRMSE and $s_\ell=8$ most RMSE comparisons; the default leads none. Multi-scale WAMRViT on PLI favors $s_\ell=1$ in first-step full-field and finest-level VRMSE and rollout-mean full-field RMSE and VRMSE. We retain $s_\ell=1$ in main experiments (details in Appendix~\ref{app:level-scale-ablation}).

\paragraph{Mesh-policy sensitivity.}
With fixed PLI checkpoints, we vary initial and runtime mesh settings (Appendix~\ref{app:pli-mesh-sensitivity}). Both variants are relatively robust to the tested tolerance rescalings, with rollout-mean full-field errors changing by at most about $3.3\%$. Neighborhood changes have larger effects, but the multi-scale advantage over uniform-patch WAMRViT persists across all tested policies.

\section{Limitations}
\label{sec:limitations}

Variable leaf counts currently require micro-batch-one adaptive training, while inference regridding adds topology updates and value transfers between model calls. AMReX training is limited by data-pipeline materialization rather than accelerator compute, as each leaf carries a $17$-channel $32\times32$ raster per input frame. Regridding accounts for $5.4$--$17.2\%$ of adaptive inference time. On AMReX, WAMRViT remains $8.6\%$ slower end to end than ViT-mid under the batch-one protocol, and SwinV2 is faster on every dataset. A dense finest-grid AMReX baseline and multi-scale full-field training are infeasible under the current materialization protocol, and training memory limits the AMReX quadtree to five of the six native levels. Padding, packed attention, and GPU-resident topology operations offer potential routes to higher throughput (Appendix~\ref{app:future-work}).

Adaptive regridding introduces deployment hyperparameters, including refinement tolerance, protection radius, and regrid interval. Rollout topologies may differ from those seen during training but should remain sufficiently similar. The current refinement criterion is non-learned and has not been compared with learned alternatives. Uniform-patch WAMRViT weights leaf losses equally, potentially underweighting broad coarse regions. The AMReX full-domain views show deviations in coarse post-shock regions. On TRL, multi-scale WAMRViT has lower full-field RMSE and VRMSE than ViT-finest but higher finest-level VRMSE. All experiments are two-dimensional, leaving generalization to three dimensions and broader physics untested. The AMReX study uses a single trajectory with a chronological holdout. Multi-trajectory training and evaluation remain future work.

\section{Conclusion}
\label{sec:conclusion}

We introduced WAMRViT, an autoregressive PDE surrogate that combines balanced quadtree tokenization, level-aware 3D RoPE, and in-cell-space regridding to adapt its hierarchy as predicted features evolve. Its uniform-patch variant builds its quadtree from simulator-native AMR hierarchies, while its multi-scale variant preserves source-grid resolution through learned level-specific encoding and decoding. On PLI, uniform and multi-scale WAMRViT reduce $20$-step rollout-mean finest-level VRMSE by $14.2\%$ and $20.1\%$, respectively, relative to the same ViT-finest baseline evaluated on the finest-level leaves of each variant's own regridded topology, while using approximately $37\%$ as many tokens. The multi-scale variant also reduces full-field VRMSE by $9.6\%$. On one-step TRL, the comparison depends on scope and metric: uniform WAMRViT reduces finest-level VRMSE by $45.3\%$ relative to ViT-finest, whereas the multi-scale variant is $14.1\%$ higher there. On native AMReX combustion data, WAMRViT reduces first-step finest-level RMSE by $60.8\%$ relative to the feasible ViT-mid projection, with a similar margin over SwinV2. Both variants achieve faster end-to-end inference than ViT-finest on the uniform-grid benchmarks, whereas on AMReX WAMRViT is $8.6\%$ slower end-to-end than ViT-mid. Together, these results demonstrate the potential of dynamically evolving token hierarchies to concentrate resolution around moving features while supporting autoregressive forecasting on simulator-native AMR data.

\begin{ack}
This work was supported by Los Alamos National Laboratory through the project ``Algorithm/Software/Hardware Co-design for High Energy Density Applications'' at the University of Michigan, and by the Center for Prediction, Reasoning, and Intelligence for Multiphysics Exploration (C-PRIME), a PSAAP-IV project funded by the U.S. Department of Energy under Grant No. DE-NA0004264.
\end{ack}

\bibliographystyle{unsrtnat}
\bibliography{refs}

\newpage

\appendix

\section{Qualitative overview of the forecasting problems}
\label{app:qualitative-overview}
\label{subsec:qualitative_rollout}

The following full-domain snapshots introduce the spatial structures that motivate adaptive forecasting before the technical appendices. PLI combines moving shocks and material interfaces with large quiescent regions; TRL develops filamentary mixing under shear and radiative cooling; AMReX resolves coupled shocks and reaction zones on a native refinement hierarchy. Each figure shows rollout step 20 on held-out data. Quantitative comparisons are reported in Tables~\ref{tab:perlevel} and~\ref{tab:perlevel-rmse}.

\subsection{PLI multimaterial shock propagation}
\begin{figure}[H]
\centering
\includegraphics[width=0.80\linewidth]{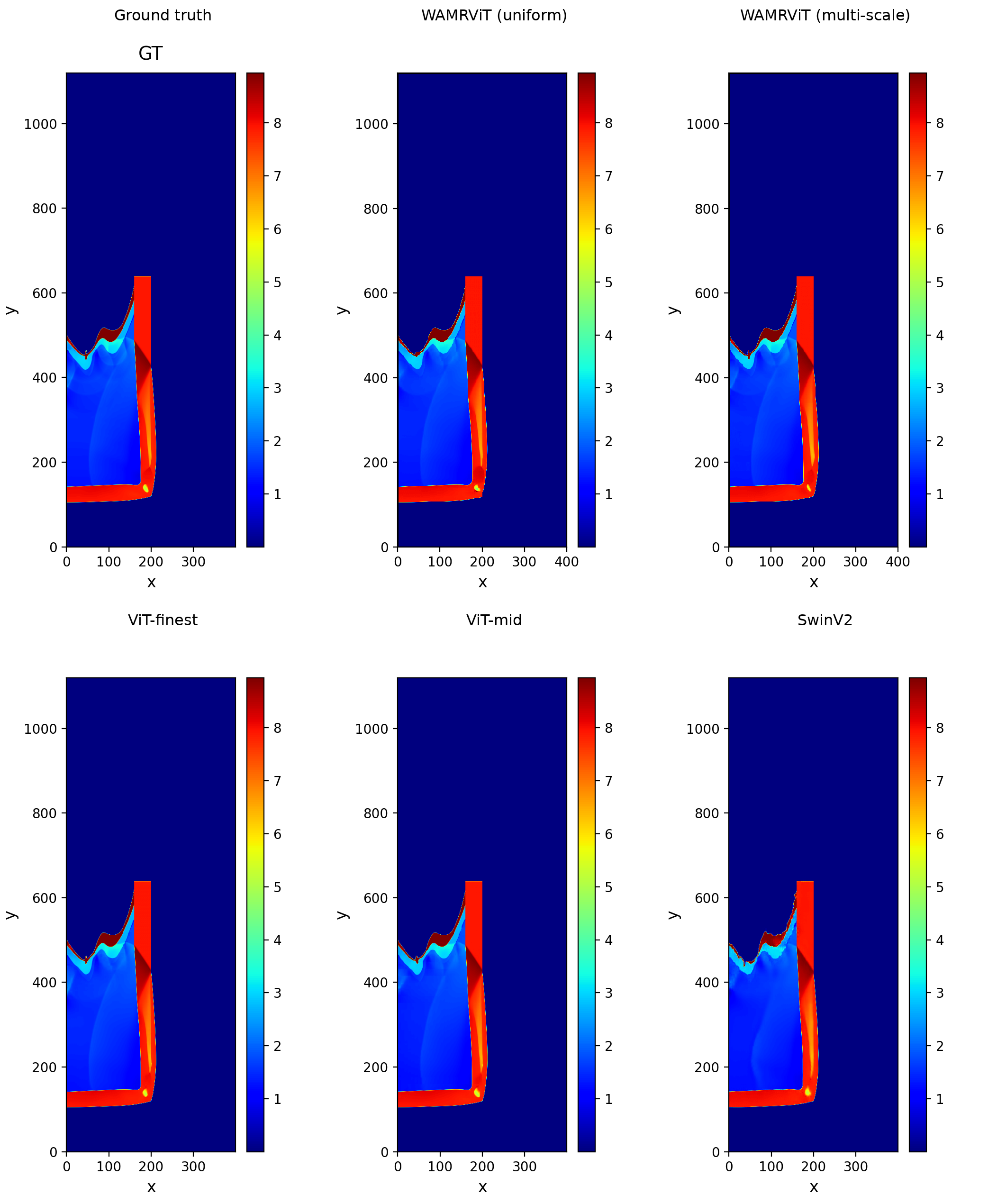}
\caption{\textbf{Qualitative rollout snapshot on PLI.} Density at rollout step 20, (trajectory index 950, in test set), starting frame 20.}
\label{fig:app-pli-qualitative}
\end{figure}

Figure~\ref{fig:app-pli-qualitative} compares density at rollout step 20 on trajectory index 950 (starting frame 20). All five models retain the large-scale material interface, while differences remain around the curved front and lower casing.

\subsection{TRL radiative mixing}
\begin{figure}[H]
\centering
\includegraphics[width=\linewidth]{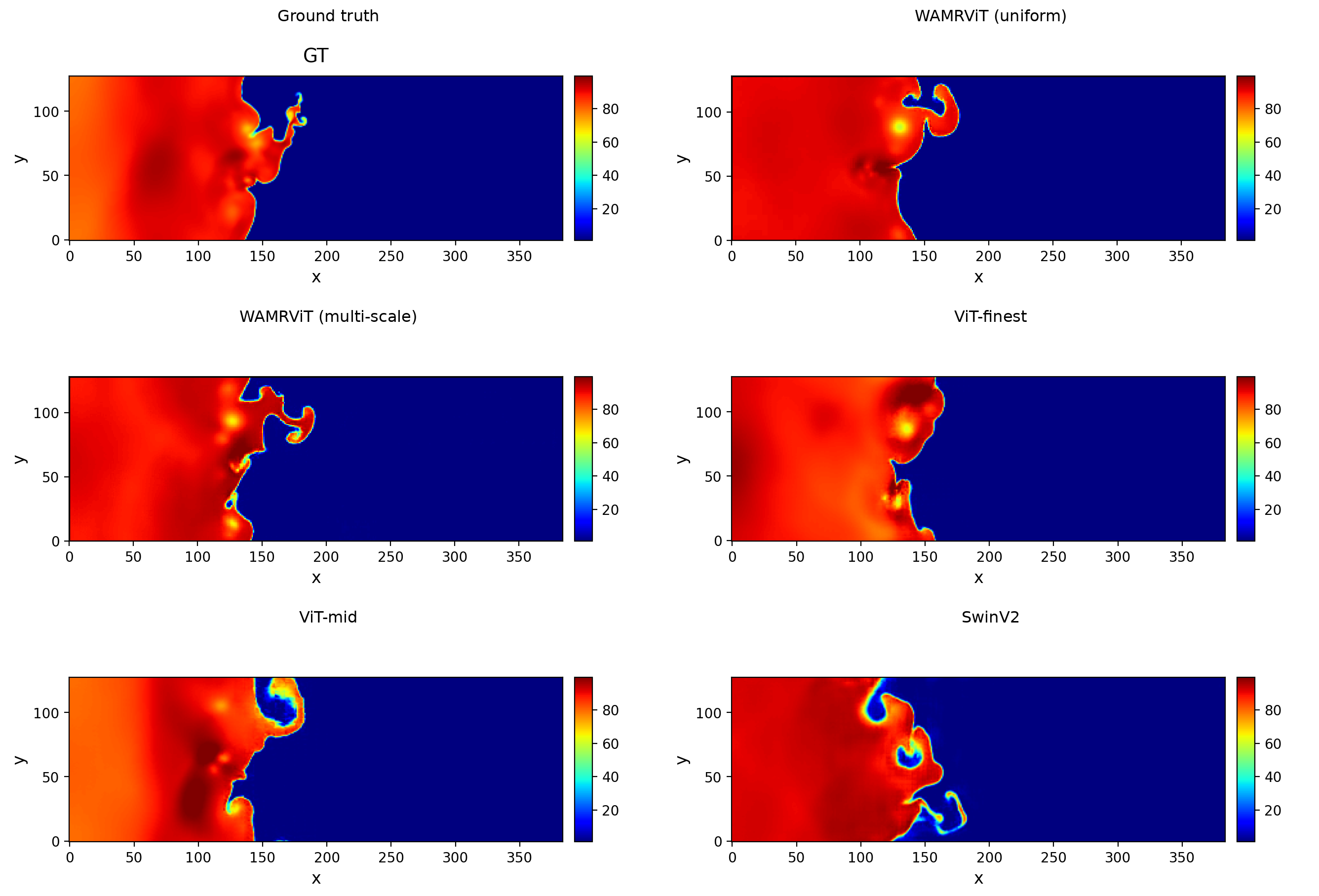}
\caption{\textbf{Qualitative rollout snapshot on TRL.} Density at rollout step 20 on trajectory 0 (starting frame 20). Both adaptive variants use initial and regrid tolerance $0.1$ per channel.}
\label{fig:app-trl-qualitative}
\end{figure}

Figure~\ref{fig:app-trl-qualitative} compares density predictions at rollout step 20 on test trajectory 0 (starting frame 20). This is a qualitative stress test beyond the one-step quantitative TRL protocol. The five learned models show different interface positions and small-scale structures at this horizon. ViT-mid and ViT-finest produce visibly different states despite sharing the same backbone, illustrating sensitivity to patch granularity.

\subsection{AMReX combustion}
\begin{figure}[H]
\centering
\includegraphics[width=0.85\linewidth]{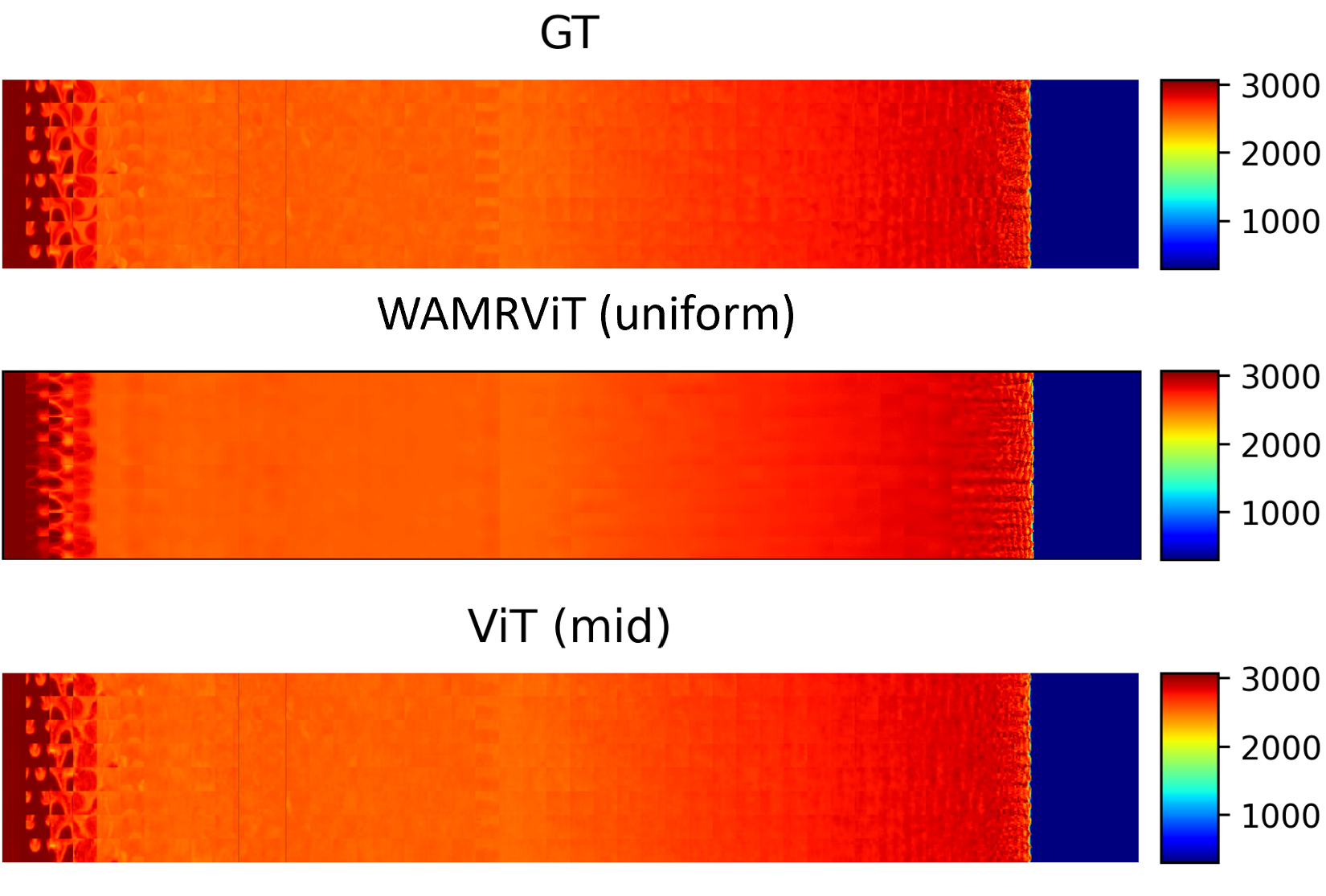}
\par
\includegraphics[width=0.88\linewidth]{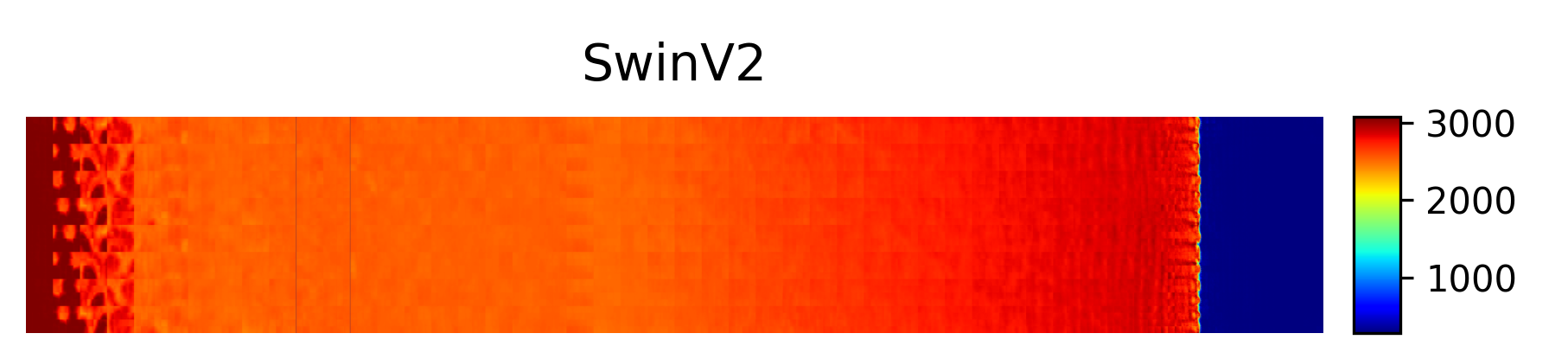}
\caption{\textbf{Qualitative rollout snapshot on AMReX combustion, full domain.} Temperature at rollout step 20 in the chronological test region, starting from frame 330. Top to bottom: ground truth, WAMRViT (uniform), ViT-mid, and SwinV2. Figure~\ref{fig:amrex-qualitative} shows a zoomed view of the leading shock region.}
\label{fig:app-amrex-qualitative}
\end{figure}

Figure~\ref{fig:app-amrex-qualitative} compares predictions at rollout step 20 starting from frame 330, within the chronological test region. Unlike the zoomed view in Figure~\ref{fig:amrex-qualitative}, these panels render the full domain at capped resolution because each AMReX snapshot averages about $0.67$~GB. All three models track the leading combustion front at this scale. Uniform WAMRViT shows residual post-shock artifacts in coarse regions. Its quantitative benefit is concentrated on the dynamically refined leaves and is therefore clearer in the finest-level metrics and zoomed view.

\section{Datasets}
\label{app:datasets}

This appendix supplements Section~\ref{sec:datasets} with full channel lists and simulation provenance.

\paragraph{Perturbed-Layer-Interface (PLI) multimaterial shock propagation.}
The PLI subset of the HEAT (High Explosives \& Affected Targets) benchmark \citep{banesh2025heat} consists of cylindrically symmetric ($R,Z$) simulations in which a high-explosive main charge detonates at $t=0$ and drives a shock through six fixed material slots: the high explosive, a stainless-steel case, a polymer cushion, an aluminum striker, a copper throw plate, and surrounding air. The per-trajectory degrees of freedom are six spline coefficients defining four perturbed interface-shape splines (cushion-striker anchor $sa$, striker thickness $st$, cushion thickness $ct$, and throw thickness $tt$ in the dataset notation). We use the first $1000$ released trajectories, each stored at fp16 with $101$ snapshots on a uniform $1120\times400$ Eulerian grid. From the solver fields, our $10$-channel observation retains velocity components $(U,W)$, cell-averaged density $\rho$, the sum of the six material-energy channels, $E_{\mathrm{sum}}=\sum_m E_m$, and six volume-of-fluid indicators $\phi_{\mathrm{case}}$, $\phi_{\mathrm{cushion}}$, $\phi_{\mathrm{maincharge}}$, $\phi_{\mathrm{outside\_air}}$, $\phi_{\mathrm{striker}}$, and $\phi_{\mathrm{throw}}$. The first $800$ trajectories are used for training and the remaining $200$ for evaluation.

\paragraph{Turbulent radiative layer 2D (TRL).}
TRL is drawn from The Well benchmark suite \citep{ohana2024well}, which packages the simulations of \citet{fielding2020multiphase}. It contains two-dimensional simulations of a radiatively cooling shear layer in which a cool, dense slab is sheared against a hot, diffuse medium while radiative cooling drives rapid interface transitions. Trajectories vary in the ratio of cooling time to shear time, which controls the filamentary mixing structure. Each snapshot occupies a uniform $128\times384$ grid. The four observation channels are density $\rho$, pressure $p$, and velocity components $(u_x,u_y)$. We adapt on all four channels with a common tolerance because each resolves the same interface structure. We use The Well's official train/test split ($72$ and $9$ trajectories; its $9$ validation trajectories are unused) and follow its reference forecasting protocol, predicting one future frame from four input frames ($4\to1$). One-step accuracy is evaluated over all $873$ windows from nine held-out trajectories.

\paragraph{AMReX combustion: stratified-detonation turbulent reacting layer.}
The trajectory is generated from a direct numerical simulation of an H$_2$/air detonation~\citep{raman2025combustion}, solving the compressible reactive Navier--Stokes equations with detailed chemistry (9 species and 21 reactions) following \citet{mueller1999flow}. Block-structured AMR resolves shocks and reaction zones at micrometer scales while maintaining tractable simulation cost. The native hierarchy contains six coexisting levels, indexed $0$--$5$, with refinement dynamically tracking the detonation front. To fit training memory, WAMRViT uses levels $0$--$4$, whose finest grid is $4096\times24576$ (Appendix~\ref{app:amrex}).

The $17$-channel state comprises heat release rate (HRR), temperature $T$, Mach number, pressure, density $\rho$, total energy density $\rho E$, momentum components $(\rho u_x,\rho u_y)$, and nine density-weighted species mass fractions $\rho Y_X$ for $X\in\{\mathrm{H},\mathrm{H}_2,\mathrm{H}_2\mathrm{O},\mathrm{H}_2\mathrm{O}_2,\mathrm{HO}_2,\mathrm{N}_2,\mathrm{O},\mathrm{O}_2,\mathrm{OH}\}$. A dense finest-grid representation is neither stored nor practical because the finest level occupies only a localized fraction of the domain. Uniform controls therefore use a projection onto the AMR level-$2$ grid ($1024\times6144$), whereas WAMRViT samples each adaptive leaf from the finest AMReX level covering it (Appendix~\ref{app:amrex}).

\section{Quadtree Construction and Value Assignment}
\label{app:quadtree}

This appendix describes the quadtree data structure, the refinement and coarsening primitives, and the procedures used to populate leaf cells with field values from both uniform-grid arrays and natively adaptive AMR data. Unless stated otherwise, the spatial value formulas describe one frame in the default uniform-patch path. All frames in a retained input window share one topology, and spatial value transfers are applied independently to each frame. The native-storage split/stitch operations are given in Appendix~\ref{app:multiscale-method}.

\subsection{Domain tiling and cell geometry}

The physical domain $\Omega = [x_{\min}, x_{\max}] \times [y_{\min}, y_{\max}]$ is partitioned into a regular grid of \emph{root tiles}, each of physical size $(\tau_w, \tau_h)$. The numbers of tiles in each direction are
\[
   n_x = \frac{W}{\tau_w}, \qquad n_y = \frac{H}{\tau_h},
 \]
where $W = x_{\max} - x_{\min}$ and $H = y_{\max} - y_{\min}$, and both ratios must be integers. Each root tile is a leaf cell at level $\ell = 0$ with center $(c_x, c_y)$ and half-extents $(h_x, h_y) = (\tau_w/2, \tau_h/2)$.

A cell at level $\ell$ can be subdivided into four children at level $\ell + 1$, up to maximum depth $L$. The four children are indexed by quadrant $q \in \{\mathrm{SW}, \mathrm{SE}, \mathrm{NW}, \mathrm{NE}\}$ with centers and half-extents
\begin{align}
   c_x^{(q)} &= c_x \pm \tfrac{h_x}{2},
   &
   c_y^{(q)} &= c_y \pm \tfrac{h_y}{2},
   \label{eq:child-center} \\
   h_x^{(q)} &= \tfrac{h_x}{2},
   &
   h_y^{(q)} &= \tfrac{h_y}{2},
   \label{eq:child-half}
 \end{align}
where the sign in~\eqref{eq:child-center} is determined by the quadrant (e.g.\ SW $\to$ $(-,-)$, NE $\to$ $(+,+)$). Throughout this appendix we use a $y$-axis-up convention, so smaller $c_y$ corresponds to the southern children and array storage is indexed accordingly. Each leaf stores a raster $v \in \mathbb{R}^{C \times p_h \times p_w}$ at a fixed patch shape $(p_h, p_w)$ shared across all levels.

\subsection{Refinement (bilinear upsampling)}
\label{app:refine}

When a leaf at level $\ell$ with value $v \in \mathbb{R}^{C \times p_h \times p_w}$ is refined under uniform-patch storage, its patch is split at the integer midpoints $h_m=\lfloor p_h/2\rfloor$ and $w_m=\lfloor p_w/2\rfloor$:
\begin{align}
   v_{\mathrm{SW}} &= v[:,\, 0{:}h_m,\; 0{:}w_m], \\
   v_{\mathrm{SE}} &= v[:,\, 0{:}h_m,\; w_m{:}p_w], \\
   v_{\mathrm{NW}} &= v[:,\, h_m{:}p_h,\; 0{:}w_m], \\
   v_{\mathrm{NE}} &= v[:,\, h_m{:}p_h,\; w_m{:}p_w].
 \end{align}
The source quadrants may differ by one pixel when a patch dimension is odd. Each quadrant is independently bilinearly resized to the common child storage shape $(C,p_h,p_w)$, while the four children retain equal physical extents. Align-corners interpolation aligns the corner pixels of the output with those of the source quadrant. For each channel $c$, given a quadrant $v_q \in \mathbb{R}^{h_q \times w_q}$ with its actual height $h_q$ and width $w_q$ and target size $(p_h,p_w)$, the upsampled value at output position $(i,j)$ is
\begin{equation}
   \hat{v}_q[i, j] = (1-w_y)(1-w_x)\,v_q[i_0, j_0]
     + (1-w_y)w_x\,v_q[i_0, j_1]
     + w_y(1-w_x)\,v_q[i_1, j_0]
     + w_y w_x\,v_q[i_1, j_1],
   \label{eq:bilinear}
 \end{equation}
with $(i_0, j_0)$, $(i_1, j_1)$ the floor and ceil source indices and $w_y, w_x \in [0,1]$ the fractional offsets. The four children are then assigned as leaves at level $\ell + 1$ with these upsampled patches.

\paragraph{Training-vs.-inference role.}
At training time, the bilinear values produced during refinement are immediately overwritten by the data-fill step (Appendix~\ref{app:datafill}), so the interpolation choice has no effect on the training objective. At inference, regridding (Section~\ref{sec:rollout}) carries the model's predictions through topology changes via this bilinear operation.

\subsection{Coarsening ($2\times 2$ block-mean)}
\label{app:coarsen}

Coarsening merges four sibling leaves into their parent: given an internal node whose four children are all leaves with values $v_{\mathrm{SW}}, v_{\mathrm{SE}}, v_{\mathrm{NW}}, v_{\mathrm{NE}} \in \mathbb{R}^{C \times p_h \times p_w}$, we stitch them into a $(C, 2p_h, 2p_w)$ super-patch in spatial layout:
\begin{equation}
   V_{\mathrm{stitched}} =
   \begin{bmatrix}
     v_{\mathrm{NW}} & v_{\mathrm{NE}} \\
     v_{\mathrm{SW}} & v_{\mathrm{SE}}
   \end{bmatrix}
   \in \mathbb{R}^{C \times 2p_h \times 2p_w}.
   \label{eq:stitch}
 \end{equation}
The matrix is the spatial layout under our $y$-up convention. In array storage $V_{\mathrm{stitched}}[:,\, 0{:}p_h,\, :]$ holds the southern children $\{v_{\mathrm{SW}}, v_{\mathrm{SE}}\}$ and $V_{\mathrm{stitched}}[:,\, p_h{:}2p_h,\, :]$ the northern. The super-patch is then downsampled by exact $2\times 2$ block averaging:
\begin{equation}
   v_{\mathrm{parent}}[c, i, j]
   = \frac{1}{4}\!\sum_{di \in \{0,1\}} \sum_{dj \in \{0,1\}}
     V_{\mathrm{stitched}}[c,\, 2i + di,\, 2j + dj].
   \label{eq:block-avg}
 \end{equation}
The operation is mean-conservative: $\mathrm{mean}(v_{\mathrm{parent}}) = \mathrm{mean}(V_{\mathrm{stitched}})$ for each stored channel. The four children are removed and the parent becomes a leaf with value $v_{\mathrm{parent}}$.

\subsection{2:1 balance enforcement}
\label{app:balance}

After any topology modification (refinement or coarsening), we enforce the 2:1 balance constraint: for every leaf $C_i$ at level $\ell_i$ and every face neighbor $C_j$, we require $|\ell_i - \ell_j| \le 1$. The enforcement loop iterates until convergence:
\begin{enumerate}
\item For each leaf $C_i$, check all four face directions. If any neighbor $C_j$ satisfies $\ell_j - \ell_i > 1$ and $\ell_i < L$, add $C_i$ to a pending set.
\item Refine every cell in the pending set.
\item Repeat until the pending set is empty.
\end{enumerate}
Because each iteration only refines cells (never coarsens), and the maximum depth $L$ is finite, the loop terminates in at most $L$ iterations.

\subsection{Value assignment from uniform-grid arrays}
\label{app:datafill}

Given a field $x \in \mathbb{R}^{C \times N_y \times N_x}$ defined on a regular pixel grid of $N_y$ rows and $N_x$ columns covering $\Omega$, and a quadtree $Q$ with a fixed topology, each leaf $C_i$ at level $\ell_i$ is populated as follows.

The pixel spacing is $\Delta x = W / N_x$, $\Delta y = H / N_y$, where $W, H$ are the physical extents introduced above. The leaf's physical extent maps to a rectangular region of the source array:
\begin{align}
   j_{\mathrm{start}} &= \lfloor (c_x - h_x - x_{\min}) / \Delta x \rfloor,
   &
   j_{\mathrm{end}} &= \lceil (c_x + h_x - x_{\min}) / \Delta x \rceil, \\
   i_{\mathrm{start}} &= \lfloor (c_y - h_y - y_{\min}) / \Delta y \rfloor,
   &
   i_{\mathrm{end}} &= \lceil (c_y + h_y - y_{\min}) / \Delta y \rceil.
 \end{align}
For this block-average assignment, the source grid satisfies $N_x=n_xp_w2^L$ and $N_y=n_yp_h2^L$, so leaf boundaries align with source-pixel boundaries. The extracted region $R_i = x[:, i_{\mathrm{start}}:i_{\mathrm{end}}, j_{\mathrm{start}}:j_{\mathrm{end}}]$ therefore has shape $(C, p_h r_{\ell_i}, p_w r_{\ell_i})$, where $r_{\ell_i} = 2^{L - \ell_i}$. Each stored raster pixel is obtained by averaging an $r_{\ell_i}\times r_{\ell_i}$ source block:
\begin{equation}
   v[c, a, b] = \frac{1}{r_{\ell_i}^2} \sum_{da=0}^{r_{\ell_i}-1} \sum_{db=0}^{r_{\ell_i}-1}
     R_i[c,\, r_{\ell_i} a + da,\, r_{\ell_i} b + db].
   \label{eq:assign-block-avg}
 \end{equation}
For the finest level ($\ell_i = L$), $r_L = 1$ and the assignment is a direct copy with no averaging. This operation involves no interpolation (every pixel in the source array contributes to exactly one leaf's block average), so the assignment is exact and conservative.

\subsection{Value assignment from AMReX plotfiles}
\label{app:amrex}

AMReX~\citep{Zhang2019} simulations store data on a hierarchy of overlapping grids: level-0 grids cover the entire domain at coarse resolution, and successively finer levels cover subregions with refinement ratio $r$ (typically $r = 2$). Where grids from different levels overlap, the finest available level takes precedence. The challenge for our quadtree tokenizer is to resolve this ownership and extract values at each leaf's resolution from the correct AMR level.

\paragraph{Topology construction.}
AMReX regrids between snapshots, so its grids cannot serve directly as the shared input/target partition of a training window. Instead, each window's quadtree is built from the AMR hierarchy of its reference frame (its last input frame):

\begin{enumerate}
\item \textbf{Coordinate mapping.} AMReX grids are stored in physical coordinates. We map all grid extents to an index-space $[0, n_x 2^L] \times [0, n_y 2^L]$ via a linear transformation, so that the finest-level cell has unit side length. The native hierarchy has six levels. We cap the quadtree at $L=4$ to fit training memory, so native level-$5$ regions map to level-$4$ leaves whose pixels sample the level-$5$ data at level-$4$ spacing (Eq.~\eqref{eq:amrex-sample}).

\item \textbf{Refinement predicate.} A leaf $C_i$ at level $\ell_i$ is marked for refinement if any AMR grid at a level finer than $\ell_i$ has a non-trivial intersection with $C_i$'s spatial extent:
\begin{equation}
       C_i \text{ is refined} \iff \exists\, g : \ell_g > \ell_i
       \;\wedge\; (C_i \cap g) \ne \varnothing.
       \label{eq:amrex-refine}
     \end{equation}
This predicate is evaluated via a Numba-accelerated \citep{lam2015numba} scan over all AMR grids.

\item \textbf{Iterative refinement.} The predicate is applied across all leaves in a loop until no further refinements occur. The result is a quadtree whose leaf-level distribution mirrors the AMR refinement pattern.

\item \textbf{2:1 balance.} After refinement converges, the 2:1 balance restoration of Appendix~\ref{app:balance} is run until stable.

\item \textbf{Detector adjustment and dilation.} Using HRR values sampled from the reference frame by the pixel-level extraction rule below, we adjust the hierarchy-derived tree with Algorithm~\ref{alg:regrid} at tolerance $\varepsilon=0.01$, omitting the protection phase. We then dilate its refined regions by $\delta=2$ face-neighbor layers, as for uniform-grid sources.
\end{enumerate}

\paragraph{Pixel-level value extraction.}
Once the topology is fixed, each leaf's $(C, p_h, p_w)$ raster is populated by a Numba-accelerated kernel. For each pixel $(i_y, i_x)$ within a leaf's patch, the kernel computes the pixel's physical center
\begin{equation}
   (p_x, p_y) = \bigl(x_{\min}^{(C_i)} + (i_x + \tfrac{1}{2})\,\Delta x_{\mathrm{leaf}},\;\;
                       y_{\min}^{(C_i)} + (i_y + \tfrac{1}{2})\,\Delta y_{\mathrm{leaf}}\bigr),
   \label{eq:pixel-center}
 \end{equation}
where $\Delta x_{\mathrm{leaf}} = 2 h_x / p_w$ and $\Delta y_{\mathrm{leaf}} = 2 h_y / p_h$. Among all AMR grids that contain $(p_x, p_y)$, the grid at the highest AMR level is selected, and the field value at the nearest grid cell center is copied:
\begin{equation}
   v[c, i_y, i_x] = g^*_{\mathrm{data}}\!\bigl[c,\; \lfloor(p_y - y_{\min}^{g^*}) / \Delta y_{g^*}\rfloor,\;
                                                        \lfloor(p_x - x_{\min}^{g^*}) / \Delta x_{g^*}\rfloor\bigr],
   \label{eq:amrex-sample}
 \end{equation}
where $g^* = \arg\max_{g \ni (p_x,p_y)} \ell_g$. This ``highest-level wins'' rule correctly resolves the AMReX overlap convention: where a fine grid exists, its data takes precedence over the coarser grid underneath.

\paragraph{Contrast with uniform-grid projection.}
A uniform-grid alternative would materialize the entire AMR hierarchy as a dense array. At the finest resolution represented by the capped quadtree, this array has shape $(C, H_{\max}, W_{\max})$, where $H_{\max} = n_y\, p_h\, 2^L$ and $W_{\max} = n_x\, p_w\, 2^L$. Here, $L$ is the quadtree cap, not the simulator's highest AMR level. For problems with localized refinement (such as our AMReX combustion dataset, where refinement follows the flame front), the projected array is dominated by coarse regions that have been trivially interpolated to fine resolution, wasting memory and compute. Our quadtree tokenizer avoids this materialization entirely: it reads values from the finest AMR level available at each pixel location, and stores them in a variable-$N$ token set whose size reflects the actual refinement structure.

\subsection{Tensor conversion}
\label{app:tensor}

For one frame, topology construction and value assignment yield the serialized tensor pair $(X, \mathbf{c})$. Leaves are enumerated in Morton (Z-curve) order and their rasters are stacked:
\begin{align}
   X &\in \mathbb{R}^{N \times C \times p_h \times p_w}, \\
   \mathbf{c} &\in \mathbb{R}^{N \times 3},
 \end{align}
where $N$ is the number of leaves, and $\mathbf{c}[i] = (\tilde c_x^{(i)}, \tilde c_y^{(i)}, \ell_i)$ are the normalized centers and level index. Center normalization maps $(c_x, c_y)$ from $[x_{\min}, x_{\max}] \times [y_{\min}, y_{\max}]$ to $[0, 1]^2$. The level index $\ell_i$ is stored as an integer and used directly as the third RoPE coordinate (see Section~\ref{sec:rope}). Stacking the $T_\text{in}$ per-frame tensors in this shared leaf order gives the $(N, C, T_\text{in}, p_h, p_w)$ input consumed by the transformer. At time $t$, $N=N_t$.

\subsection{Random resampling augmentation}
\label{app:resample}

The wavelet refinement criterion creates a deterministic correlation between refinement level and spatial complexity: coarse tokens see smooth patches while fine tokens carry sharp features. We hypothesized that, if smooth patches also change slowly, this distribution could bias coarse-level predictions toward the identity map, which would be harmful after the inference-time regridder (Section~\ref{sec:rollout}) coarsens a previously refined region whose underlying physics is still active.

To weaken this correlation, we perturb the quadtree topology at training time between topology construction and data fill.

\paragraph{Procedure.}
Given ratios $(\rho_c, \rho_r) \in [0,1]^2$, three phases are applied to the quadtree $Q$:
\begin{enumerate}
\item \textbf{Coarsen.} Collect all internal nodes whose four children are leaves with $\ell > 0$ (root tiles are preserved), yielding $n_c$ candidates after deduplication. Sample $\max(1, \lfloor \rho_c n_c \rfloor)$ of them uniformly at random and coarsen them by the block-mean rule of Appendix~\ref{app:coarsen}.
\item \textbf{Refine.} Collect all current leaves with $\ell_i < L$, yielding $n_r$ candidates. Sample $\max(1, \lfloor \rho_r n_r \rfloor)$ uniformly at random and refine them by the bilinear-upsample rule of Appendix~\ref{app:refine}. Candidates are re-collected after the coarsen phase, so freshly coarsened cells remain eligible for refinement.
\item \textbf{Restore balance.} Iterate the 2:1 balance restoration of Appendix~\ref{app:balance} to a fixed point. Termination in at most $L$ passes follows from the monotone-refining argument of Appendix~\ref{app:balance}.
\end{enumerate}

\paragraph{Settings and scope.}
The main Table~\ref{tab:perlevel} runs disable this augmentation. When enabled in the PLI and AMReX ablations of Appendix~\ref{app:resample-ablation}, it uses $\rho_c = 0.1$ and $\rho_r = 0.15$. Resampling is training-only. Inference-time topology follows the regridding procedure of Section~\ref{sec:rollout}.

\subsection{Training-time tokenization pipeline}
\label{app:train-tokenize}
Algorithm~\ref{alg:train-tokenize} composes the primitives of Appendices~\ref{app:refine}--\ref{app:tensor} to convert a single uniform-grid reference frame into the per-frame $(X, \mathbf{c})$ pair. For a training window, the topology is constructed once from its reference frame. Data fill and serialization then reuse that topology for each input and target frame, with the input tensors stacked as in Appendix~\ref{app:tensor}. The per-channel refinement loop (lines~3--14) is the wavelet-inspired local-range criterion of Section~\ref{sec:tokenizer}. Independent per-channel trees are unioned, and thresholding each channel relative to its own range $R_c$ keeps $\varepsilon$ a single global hyperparameter. The balance restoration after the union follows the monotone-refining procedure of Appendix~\ref{app:balance}. We then expand the refined regions by $\delta$ face-neighbor layers, refining coarser neighboring leaves and restoring 2:1 balance after each expansion. Random resample augmentation (Appendix~\ref{app:resample}) is applied before data fill so that augmented topology inherits the block-average assignment rule, which guarantees that augmentation does not leak ground-truth fine detail into coarsened cells.

\begin{algorithm}[H]
 \caption{Training-time quadtree tokenization of a single reference frame}
 \label{alg:train-tokenize}
 \begin{algorithmic}[1]
 \Require uniform-grid field $x \in \mathbb{R}^{C \times H \times W}$, tolerance $\varepsilon$, adaptation channels $\mathcal{A} \subseteq \{1,\dots,C\}$, max depth $L$, refinement-dilation radius $\delta$
 \State Initialize $Q$ as a uniform tiling of $\Omega$ by root cells of size $(\tau_w, \tau_h)$, maximum depth $L$
 \State Compute the global per-channel range $R_c \gets \max_\Omega x_c - \min_\Omega x_c$ for each $c \in \mathcal{A}$
 \For{each channel $c \in \mathcal{A}$}
   \State Initialize $Q_c$ as a copy of $Q$ with all leaves at root level
   \For{$p = 1, \dots, L$} \Comment{iterative refinement passes}
     \For{each leaf $C_i$ of $Q_c$}
       \State Compute the local range $r_c(C_i) \gets \max_{C_i} x_c - \min_{C_i} x_c$
       \If{$r_c(C_i) > \varepsilon \cdot R_c$ \textbf{and} $\ell_i < L$}
         \State Split $C_i$ into four equal children
       \EndIf
     \EndFor
     \If{no cells were split in this pass} \textbf{break} \EndIf
   \EndFor
 \EndFor
 \State Merge the per-channel trees by union: a cell is refined in $Q_{\mathrm{merged}}$ iff it is refined in any $Q_c$
 \State Set $Q \gets Q_{\mathrm{merged}}$ and restore 2:1 balance by refining any leaf whose face neighbor is more than one level finer
 \For{$d = 1, \dots, \delta$}
   \State Collect coarser face neighbors of the current refined leaves
   \State Refine the collected leaves and restore 2:1 balance
 \EndFor
 \If{augmentation is enabled}
   \State Apply random resample augmentation to $Q$ (Appendix~\ref{app:resample}) with coarsen/refine ratios $(\rho_c, \rho_r)$
   \State Restore 2:1 balance
 \EndIf
 \State Fill each leaf of $Q$ by block-averaging the corresponding region of the source array $x$
 \State Collect leaves in Morton order: $X$ stacks the per-cell rasters, $\mathbf{c}$ stacks the per-cell centers
 \State \Return $(X, \mathbf{c})$
 \end{algorithmic}
 \end{algorithm}

The pipeline is staged so its expensive parts amortize across training. An offline \emph{topology pass} runs Algorithm~\ref{alg:train-tokenize} up to and including fine-region expansion and balance restoration once per training window, building each window's tree from its reference frame. The result is serialized as a difference against the trivially uniform coarse tree and written to a Parquet table. At each training step the cached tree is reconstructed in memory and the optional augmentation and data-fill steps execute per visit, with the uniform-grid source frames streamed from disk afresh on every step rather than bundled into the cache. This lazy fill keeps augmented topology inside the exact-conservative regime of Eq.~\eqref{eq:assign-block-avg} and reduces cache size to a small fraction of the raw dataset. Both passes are scheduled by the distributed dataset abstraction of Ray~\citep{moritz2018ray}: the topology pass parallelizes across an autoscaling task pool with per-task memory budgets sized for heavily-refined windows, and the fill pass runs with asynchronous prefetch that overlaps the CPU-bound block-average with the GPU forward pass.

\subsection{Three-phase regrid algorithm}
\label{app:regrid}
Algorithm~\ref{alg:regrid} gives inference-time regridding for default uniform-patch storage. Splitting a cell bilinearly upsamples each retained frame onto four children, and merging four siblings replaces each frame by its $2\times 2$ block mean, as detailed in Appendices~\ref{app:refine}--\ref{app:coarsen}. Multi-scale native storage uses the same three-phase structure with exact split/stitch value transfers (Appendix~\ref{app:multiscale-method}). All refinement indicators, channel ranges, and protection/coarsening tests use the newest retained frame. Topology changes transfer every frame of the retained input window. The balance-aware coarsening lookahead predicate used in Phase~3 is defined below.

\paragraph{Refinement criterion at inference.}
At inference the uniform-grid source field is unavailable and only the cellular state is accessible: a feature whose discontinuity aligns with a cell boundary leaves both adjacent cells looking smooth internally and would be missed by within-cell range alone. For channel $c$, let $r_c(C_i) = \max_{C_i} x_c - \min_{C_i} x_c$ denote the range within the leaf's raster. To detect variation between leaves, sample the cellular state one leaf-width from $C_i$'s center in each direction $d\in\{W,E,S,N\}$, using the width along that direction. Each sample $v_c^{(d)}(C_i)$ is obtained by bilinear interpolation in the highest-level cell containing the sample point. With $v_c(C_i)$ denoting the center value, define the jump indicator as $b_c(C_i) = \max_d |v_c(C_i) - v_c^{(d)}(C_i)|$. Phase~1 uses the combined indicator $e_c(C_i) = \max\!\big(r_c(C_i),\, b_c(C_i)\big)$.

Phase~1 refines a leaf when $e_c(C_i) > \varepsilon_c R_c$ for some $c \in \mathcal{A}$, where $\varepsilon_c$ is the inference tolerance for channel $c$. This retains the range-scaled threshold form used during training, but the tolerances may differ (Table~\ref{tab:regrid-inference}). A scalar tolerance sets $\varepsilon_c=\varepsilon$ for every selected channel. Phases~2 and~3 instead use the within-cell range $r_c$ alone, with the coarsen-side hysteresis threshold $\gamma \varepsilon_c R_c$ ($\gamma < 1$). For an internal node $P$ whose four children are leaves, $r_c(P)$ denotes the per-channel range over the union of the four children's rasters.

\begin{algorithm}[H]
 \caption{Inference-time regridding with uniform-patch storage}
 \label{alg:regrid}
 \begin{algorithmic}[1]
 \Require quadtree $Q$ carrying the retained $T_\text{in}$-frame window, per-channel tolerances $\{\varepsilon_c\}_{c\in\mathcal{A}}$, coarsen ratio $\gamma < 1$, protection radius $\delta$, adaptation channels $\mathcal{A}$, max level $L$, max passes per phase $p_{\max}$; per-channel global range $R_c$ is computed from the newest retained frame
 \Statex \hrulefill
 \Statex \textbf{Phase 1: Refine}
 \For{$p = 1, \dots, p_{\max}$}
   \For{each leaf $C_i$ of $Q$}
     \If{$\exists\, c \in \mathcal{A}:$ $e_c(C_i) > \varepsilon_c \cdot R_c$ \textbf{and} $\ell_i < L$}
       \State Split $C_i$ into four children and bilinearly upsample each retained frame onto them
     \EndIf
   \EndFor
   \State Restore 2:1 balance by recursively splitting any leaf whose neighbor is more than one level finer
   \If{no cells were split in this pass} \textbf{break} \EndIf
 \EndFor
 \Statex \hrulefill
 \Statex \textbf{Phase 2: Protect}
 \State Initialize all leaves as unprotected
 \If{$\delta > 0$}
   \For{each leaf $C_i$ such that $r_c(C_i) \ge \gamma \varepsilon_c R_c$ for some $c \in \mathcal{A}$}
     \State Mark $C_i$ as protected
   \EndFor
   \State Dilate the protected set by $\delta$ face-neighbor layers
   \For{$d = 1, \dots, \delta$}
     \State Collect protected leaves that are coarser than an adjacent protected leaf
     \If{no leaves are collected} \textbf{break} \EndIf
     \State Refine the collected leaves, propagating protection to their children
     \State Restore 2:1 balance
   \EndFor
 \EndIf
 \Statex \hrulefill
 \Statex \textbf{Phase 3: Coarsen} (skipped if coarsening is disabled for this step)
 \For{$p = 1, \dots, p_{\max}$}
   \State Collect internal nodes whose 4 children are all leaves
   \For{each collected node $P$}
     \If{no child is protected \textbf{and} lookahead of Eq.~\eqref{eq:coarsen-lookahead} holds
         \textbf{and} $\forall\, c \in \mathcal{A}:$ $r_c(P) < \gamma \cdot \varepsilon_c \cdot R_c$}
       \State Merge the four children of $P$ into $P$ by $2\times 2$ block averaging in each retained frame
     \EndIf
   \EndFor
   \If{no nodes were merged in this pass} \textbf{break} \EndIf
 \EndFor
 \State \Return $Q$ \Comment{topology and retained window updated in place}
 \end{algorithmic}
 \end{algorithm}

When considering whether to coarsen a node $P$ at level $\ell$ (i.e.\ remove its four leaf children and make $P$ a leaf at level $\ell$), the predicate must verify that no face neighbor of any child $C_j$ of $P$ (excluding siblings) is at level $\ge \ell + 2$. If such a neighbor exists, coarsening $P$ would violate the 2:1 balance constraint and would trigger a cascade of re-refinements. The predicate therefore performs a lookahead:
\begin{equation}
\begin{aligned}
   &\forall\, C_j \in \mathrm{children}(P),\quad
   \forall\, d \in \{W,E,S,N\}, \\
   &\forall\, C_k \in \mathrm{FaceNeighbors}(C_j,d)
   \setminus \mathrm{children}(P):\quad \ell_k < \ell + 2.
\end{aligned}
\label{eq:coarsen-lookahead}
\end{equation}
The lookahead preserves 2:1 balance during coarsening, while the protection buffer limits immediate coarsening around regions with appreciable within-cell variation.

\paragraph{Implementation.} Reported rollouts use a packed-array implementation of this algorithm with compiled CPU kernels (Section~\ref{sec:rollout}). If its output fails the 2:1-balance or leaf-overlap check, that regrid call is recomputed with a reference tree-based implementation of the same algorithm. Because protection and dilation admit more than one valid balanced result, the two implementations can select different topologies.

\section{Method (extended)}
\label{app:method}

This appendix collects architectural and algorithmic detail deferred from Section~\ref{sec:method}. The transformer backbone block is described in Appendix~\ref{app:backbone} and the multi-scale (native-resolution) variant of WAMRViT in Appendix~\ref{app:multiscale-method}. Quadtree primitives (refinement, coarsening, 2:1-balance restoration, AMR mapping, and the random-resample augmentation) are in Appendix~\ref{app:quadtree}; the 3D RoPE construction is in Appendix~\ref{app:rope-config} and the multi-scale tokenizer architecture in Appendix~\ref{app:multiscale-arch}; the inference-time regridding algorithm is in Appendix~\ref{app:regrid}.

\subsection{Backbone block}
\label{app:backbone}

Given the residual stream $h \in \mathbb{R}^{N \times d}$, first compute a shared normalized stream $\tilde h = \mathrm{LN}(h)$. The attention branch processes $\tilde h$ using RMSNorm on queries and keys (QK normalization~\citep{dehghani2023vit22b}), 3D RoPE, and scaled dot-product attention. In parallel, the MLP branch applies a linear map followed by SwiGLU to $\tilde h$, with intermediate width $d_{\mathrm{mlp}} = \eta\, d$. The two branch outputs are concatenated and projected back to the residual width using $W_{\mathrm{proj}} \in \mathbb{R}^{d \times (d + d_{\mathrm{mlp}})}$, then added to $h$. The parallel arrangement uses a single pre-norm rather than the two of the sequential layout. We use unit MLP expansion $\eta = 1$ throughout: at $d_{\mathrm{mlp}} = d$ the joint projection has the same $2d^2$ parameter count as two independent branch projections summed in the residual stream.

After the block stack a final LayerNorm is followed by a linear projection $W_{\mathrm{out}} \in \mathbb{R}^{d \times (C_{\mathrm{out}} T_{\mathrm{out}} p_h p_w)}$ that maps each token back to a per-leaf raster, reshaped to $(N, C_{\mathrm{out}}, T_{\mathrm{out}}, p_h, p_w)$ for adaptive mode or gathered into a dense $(C_{\mathrm{out}}, T_{\mathrm{out}}, H, W)$ grid for the uniform-grid baseline. The multi-scale variant of Appendix~\ref{app:multiscale-method} replaces both the per-block pre-norm and this post-stack norm with the level-conditioned AdaLN of Eq.~\eqref{eq:adaln} and uses a per-level decoder. The block stack is held identical across default uniform-patch WAMRViT and ViT-finest/mid, providing a shared-backbone comparison. The multi-scale variant differs as stated above.

\subsection{Multi-scale (native-resolution) variant}
\label{app:multiscale-method}

\begin{figure}[!htbp]
\centering
\includegraphics[width=0.42\linewidth,height=0.65\textheight,keepaspectratio]{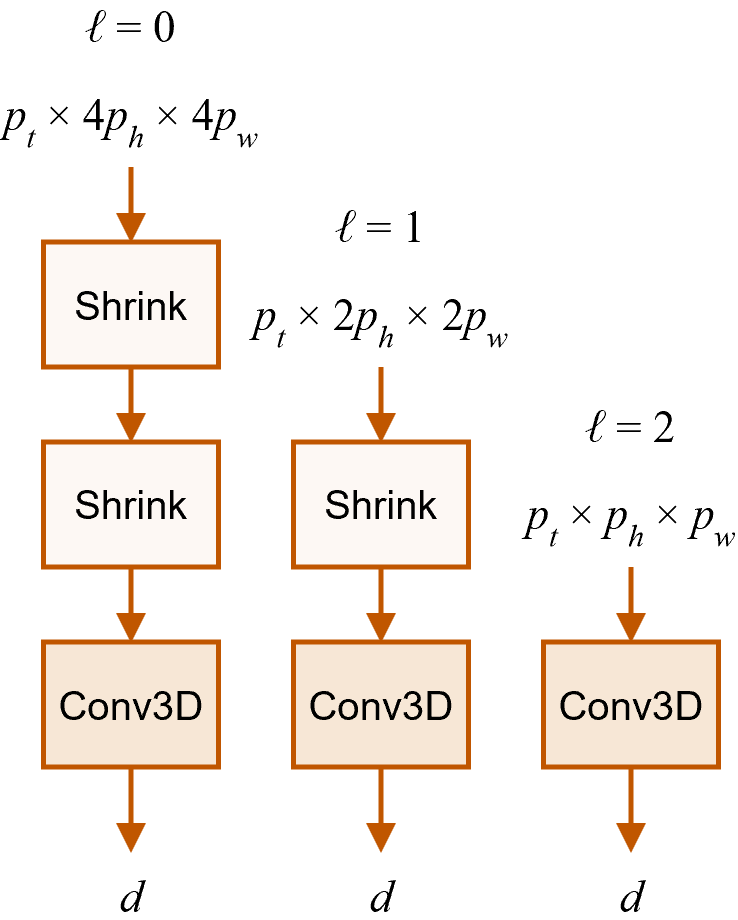}
\caption{Per-level patchify cascade of the multi-scale variant.}
\label{fig:multiscale-tokenizer}
\end{figure}

The default tokenizer downsamples coarse leaves by exact block averaging (Eq.~\eqref{eq:assign-block-avg}) to a shared $(C, p_h, p_w)$ raster, so each output pixel of a level-$\ell$ leaf summarizes $4^{L-\ell}$ source pixels, and sub-raster detail of coarse leaves cannot be recovered. With the motivation of a shared latent space~\citep{cheng2024multi} across refinement levels, the multi-scale variant retains each leaf's full source region and handles the change of resolution inside the model rather than at the data-fill step. The validated design requires a uniform-grid source and scatters predictions back to its finest grid for the full-field loss. ``Native-resolution'' here describes the retained per-leaf source region, not direct training on simulator-native AMR data.

\paragraph{Native value storage.}
In native value storage, the block averaging of Eq.~\eqref{eq:assign-block-avg} is replaced by a direct copy: each level-$\ell$ leaf retains its full $(C, p_h r_\ell, p_w r_\ell)$ source region, with native-resolution factor $r_\ell = 2^{L-\ell}$, so finest-level rasters are unchanged and only coarser leaves grow. The rest of Algorithm~\ref{alg:train-tokenize} (per-channel refinement and union, 2:1 balance, dilation, and optional resampling) acts on topology alone and is unchanged. For an input window, tensor conversion (Appendix~\ref{app:tensor}) stacks the $T_\text{in}$ frames into a per-level dictionary $\{\ell : (N_\ell, C, T_\text{in}, p_h r_\ell, p_w r_\ell)\}_{\ell=0}^{L}$ together with a leaf-to-bucket index recording each token's position within its level. This is the input of the per-level patch embedder (Appendix~\ref{app:multiscale-arch}). Rasters and partition formulas below are written per frame and applied independently across the window.

\paragraph{Refine and coarsen under native storage.}
Native storage reorganizes but does not destroy information at topology changes, so the bilinear refinement of Appendix~\ref{app:refine} and the block-mean coarsening of Appendix~\ref{app:coarsen} simplify to exact partition operations. Refinement of a level-$\ell$ leaf with raster $v \in \mathbb{R}^{C \times p_h r_\ell \times p_w r_\ell}$ splits $v$ along its spatial axes into four contiguous $(C, p_h r_\ell/2, p_w r_\ell/2)$ blocks and assigns them to the four children at level $\ell+1$, each of which is exactly the native-storage shape $(C, p_h r_{\ell+1}, p_w r_{\ell+1})$ since $r_{\ell+1} = r_\ell / 2$. Coarsening is its inverse: the four children are concatenated in their NW/NE/SW/SE layout (Eq.~\eqref{eq:stitch}) into a $(C, p_h r_\ell, p_w r_\ell)$ super-patch, which is stored verbatim at the parent rather than reduced by the $2{\times}2$ block-mean of Eq.~\eqref{eq:block-avg}. Both operations are bit-exact and conservative. The bilinear smoothing and the mean reduction of the uniform mode are not needed because the native shape already accommodates each child's pixel count.

\paragraph{Architecture and training objective.}
Each level's bucket is embedded by a per-level patchify cascade of $L-\ell$ learned shrink stages (Figure~\ref{fig:multiscale-tokenizer}) and decoded back to its native raster by a per-level expansion head, and level-conditioned AdaLN replaces the backbone's pre-norm and output norm. Appendix~\ref{app:multiscale-arch} specifies these components. Predictions are scattered onto the finest uniform grid and trained with the full-field loss of Appendix~\ref{app:multiscale-loss}.

\section{Scale-aware design choices}
\label{app:rope}

Two design choices govern how WAMRViT handles cells that differ in physical extent and refinement level: (i) the per-axis scales of the 3D rotary positional embedding, which place the level axis on a soft-bias footing relative to the spatial axes; and (ii) the optional multi-scale tokenizer for uniform-grid inputs, which retains each leaf at its source resolution rather than projecting all leaves to a shared finest patch shape. Both target the same mismatch between a fixed-shape transformer block stack and a token set whose elements span $L+1$ refinement levels and so differ in physical area by factors of up to $4^L$ within a single sample.

\subsection{3D RoPE configuration}
\label{app:rope-config}

\paragraph{Per-axis frequency partition.}
The head dimension $d_h$ is partitioned across the three axes, $d_h = d_x + d_y + d_\ell$, and each axis $a \in \{x, y, \ell\}$ owns an independent block of $d_a/2$ frequency planes in the axial 2D-RoPE style of \citet{heo2024rope}. On axis $a$, plane $j \in \{0, \dots, d_a/2 - 1\}$ has angular frequency
\begin{equation}
   \omega^{(a)}_j \;=\; \theta_{\mathrm{base}}^{-2j/d_a},
   \qquad
   \theta_{\mathrm{base}} = 10{,}000,
   \label{eq:rope-freq}
 \end{equation}
so each axis is anchored at a common fastest plane $\omega = 1$ at $j = 0$, with the slowest plane decaying as $\theta_{\mathrm{base}}^{-(d_a-2)/d_a}$. Larger $d_a$ extends the spectrum further into low frequency, while smaller $d_a$ truncates it. For the spatial axes, $p_a(C_i)=s_a\tilde{c}_a(C_i)$, $a\in\{x,y\}$, where $\tilde{c}_a\in[0,1]$ is the normalized center coordinate. For the level axis, $p_\ell(C_i)=s_\ell\ell(C_i)$, where $\ell(C_i)\in\{0,\dots,L\}$. RoPE applies a block-diagonal $2{\times}2$ rotation by phase $\omega^{(a)}_j \cdot p_a$ to each query/key on plane $j$ \citep{su2024roformer}. The per-axis rotations are concatenated along the head dimension before the attention dot product.

\paragraph{Spatial scales.}
The spatial scales use the shorter side of a finest-level leaf as a common length unit for both axes. Let $\lambda = \min(\tau_w, \tau_h) / 2^L$ be the shorter side of a finest-level leaf in domain units. Two finest-level neighbors along axis $a$ have normalized-coordinate separation $\Delta \tilde{c}_a = (\tau_a / 2^L)\,/\,W_a$ with $W_a \in \{W, H\}$ the domain extent on that axis. Setting $s_x = W/\lambda$ and $s_y = H/\lambda$ then yields a unit position step at the finest level along the smaller-tile direction, with the other axis inheriting a proportionally larger step when tiles are not square. Coarser cells inherit proportionally larger steps because their physical separation is larger. For square root tiles ($\tau_w = \tau_h$) this reduces to $s_x = W/\lambda = n_x \cdot 2^L$ and $s_y = H/\lambda = n_y \cdot 2^L$ with $n_x, n_y$ the root-tile counts, recovering the uniform ViT convention exactly.

\paragraph{Level axis as soft bias.}
The level axis carries the integer refinement level $\ell \in \{0, \dots, L\}$ rather than a continuous coordinate, and its scale is set to $s_\ell = 1$. The 2:1 balance constraint guarantees that adjacent leaves differ by at most one level, so cross-level attention is the regime in which refined regions communicate with their coarser neighborhoods rather than an edge case. The level coordinate distinguishes refinement levels explicitly within the positional encoding. Setting $s_\ell=0$ removes this contribution from RoPE, while increasing $s_\ell$ increases the relative rotary phases associated with a given level difference. The resulting attention scores depend on the learned queries and keys.

To examine the level contribution alone, consider an idealized comparison with equal query and key vectors, $q=k$, and zero spatial phase difference, varying only the level difference $\Delta\ell$. Because RoPE is a per-plane rotation and the dot product across planes is additive, the post-RoPE dot product (with uniform per-plane magnitudes $\|q_j\|^2 = 2/d_h$) is
\begin{equation}
   q^{\!\top} k(\Delta x{=}\Delta y{=}0,\, \Delta\ell)
   \;=\;
   \frac{d_x + d_y}{d_h} \;+\; \frac{2}{d_h}
   \sum_{j=0}^{d_\ell/2 - 1} \cos\bigl(\omega^{(\ell)}_j \cdot \Delta\ell\bigr),
   \label{eq:level-falloff}
 \end{equation}
normalized so that $\Delta\ell = 0$ yields $1$. With the PLI configuration of Table~\ref{tab:training-specs} ($d_h = 192$, $d_\ell = 20$, $\theta_{\mathrm{base}} = 10{,}000$), Eq.~\eqref{eq:level-falloff} evaluates to $\approx 0.994$ at $\Delta\ell = 1$: the post-RoPE dot product retains roughly $99.4\%$ of its same-level value. In this idealized comparison, the default $s_\ell=1$ produces only a small change in the dot product for adjacent levels. Larger scales change the cosine terms non-monotonically. Between adjacent quadtree levels, cell side lengths and perimeter differ by a factor of $2$, and area by a factor of $4$, irrespective of aspect ratio. These ratios motivate the geometric interpretations of $s_\ell=2$ and $s_\ell=4$. PLI uses a $2{:}1$ rectangular leaf $(p_h, p_w)=(10,5)$ and non-square spatial scales $(s_x,s_y)=(80,224)$, so its level increment occupies a different relative scale than on square-leaf AMReX. The forecasting effects of $s_\ell$ across datasets, metrics, evaluation scopes, and horizons are evaluated in Appendix~\ref{app:level-scale-ablation}.

\paragraph{Frequency-plane sizing.}
We use $d_\ell \ge 4(L+1)$, allocating at least two $2$-dimensional frequency planes per level. The configurations in Table~\ref{tab:training-specs} use $d_\ell=20$ at $L=3$ on PLI, $d_\ell=16$ at $L=2$ on TRL, and $d_\ell=24$ at $L=4$ on AMReX.

\subsection{Multi-scale tokenization architecture}
\label{app:multiscale-arch}

The multi-scale variant of Appendix~\ref{app:multiscale-method} produces a per-level token bucket $\{\ell : V_\ell \in \mathbb{R}^{N_\ell \times C \times T_\text{in} \times p_h r_\ell \times p_w r_\ell}\}_{\ell=0}^{L}$, where the native-resolution factor $r_\ell = 2^{L - \ell}$ is distinct from the RoPE level-axis scale $s_\ell$, rather than a single $(N, C, T_\text{in}, p_h, p_w)$ tensor. A per-level patch embedder converts each bucket to tokens of the shared embedding dimension $d$, and a per-level output head decodes back to native shape. We describe both below.

\paragraph{Per-level patchify cascade.}
Let $\Delta = L - \ell$ denote the number of shrink stages at level $\ell$. Each stage is a 3D convolution with kernel $(1,2,2)$ and stride $(1,2,2)$ followed by a learned SwiGLU block, halving both spatial extents and multiplying the channel count by a fixed factor $m$ (the channel multiplier). Starting from $C_0=C$, the cascade produces a tensor of shape $(C m^\Delta, T_\text{in}, p_h, p_w)$. Its spatial dimensions match those of uniform-mode leaves, with learned reduction replacing block-mean data-fill (Appendix~\ref{app:datafill}). A final per-level $\mathrm{Conv3d}(C m^\Delta, d, (p_t, p_h, p_w), \mathrm{stride}{=}(p_t, p_h, p_w))$ projects to the shared embedding dimension $d$. The finest level ($\Delta = 0$) has no shrink stages and reduces to the uniform-mode patch embedder. With $m=2$, each shrink stage halves the activation size. Coarser levels require more stages and larger per-level projections.

\paragraph{Per-level output head.}
The decoder begins with a per-level linear map $W_\ell \in \mathbb{R}^{d \times (C_\mathrm{out}m^\Delta T_\mathrm{out}p_hp_w)}$, where $\Delta=L-\ell$, followed by a reshape to $(N_\ell,C_\mathrm{out}m^\Delta,T_\mathrm{out},p_h,p_w)$. A cascade of $\Delta$ expansion stages then doubles both spatial dimensions and reduces the channel count by a factor $m$ at each stage, using a $3\times3$ convolution followed by factor-two pixel shuffle. Intermediate stages include a SwiGLU block. The final stage omits the activation. Expansion operates independently on each output frame and produces the native raster $(N_\ell,C_\mathrm{out},T_\mathrm{out},p_hr_\ell,p_wr_\ell)$. At the finest level, the cascade is empty and decoding reduces to the uniform-mode linear projection.

\paragraph{Backbone with level-conditioned norm.}
The transformer block stack is unchanged in shape (tokens of dimension $d$, sequence length $N = \sum_\ell N_\ell$ regardless of tokenization mode), but each block's pre-norm and the post-stack output norm are replaced by independent level-conditioned affine modulations in the AdaLN form of \citet{peebles2023dit}. For block index $b$, hidden state $h_i \in \mathbb{R}^d$ at token $i$, and per-token level index $\ell(i) \in \{0, \dots, L\}$,
\begin{equation}
   \mathrm{AdaLN}^{(b)}\!\bigl(h_i;\, \ell(i)\bigr)
   \;=\;
   \bigl(1 + \alpha^{(b)}_{\ell(i)}\bigr) \,\odot\, \mathrm{LN}(h_i)
   \;+\; \beta^{(b)}_{\ell(i)},
   \label{eq:adaln}
 \end{equation}
where each block owns a learned embedding $E^{(b)} \in \mathbb{R}^{(L+1) \times 2d}$ whose level-$\ell$ row splits into $\alpha^{(b)}_\ell,\, \beta^{(b)}_\ell \in \mathbb{R}^d$. All $E^{(b)}$ are initialized to zero, so every $\mathrm{AdaLN}^{(b)}$ reduces to plain $\mathrm{LN}$ at initialization. Its level-specific scale and shift parameters are then learned jointly with the rest of the model. We do not adopt the residual-branch gating that completes the AdaLN-Zero variant of \citet{peebles2023dit}. The level-conditioned norm is enabled only in the multi-scale tokenizer. On the uniform-patch tokenizer the same modulation regressed in preliminary experiments (not reported), suggesting that its gains are tied to the multi-scale token distribution.

\paragraph{Three roles for the level index.}
In the multi-scale variant, the refinement level enters the model in three ways: as the third coordinate of the 3D RoPE (Appendix~\ref{app:rope-config}); as the resolution index that fixes each token's native raster size and its patch-embedder depth; and as the conditioning index of the AdaLN modulation in Eq.~\eqref{eq:adaln}. RoPE encodes relative spatial position and refinement level in the attention mechanism. The multi-scale tokenizer keeps each leaf's raster at its source resolution, so the model sees the underlying field at native fidelity per level. The level-conditioned norm gives each level a learned scale-and-shift that absorbs systematic distributional differences (e.g.\ coarse leaves dominated by smooth content vs.\ fine leaves carrying sharp features), without requiring the backbone to recover that distinction from RoPE alone. The variant reported as ``WAMRViT (multi-scale)'' in Section~\ref{sec:results} adopts all three.

\section{Training and evaluation specifications}
\label{app:training-specs}

\subsection{Training specifications}
Table~\ref{tab:training-specs} lists the training hyperparameters used for the adaptive-WAMRViT runs reported in Section~\ref{sec:results}. The SwinV2 counterpart is listed in Table~\ref{tab:swin-specs}. Rows with a single value apply to all three datasets. All runs use Intel Xeon Platinum 8468 CPUs alongside the GPUs listed in the table.

\begin{table}[H]
\centering
\caption{Training specifications for the WAMRViT runs reported in Section~\ref{sec:results}. Topology resampling is disabled. AMReX builds its quadtree from the simulator hierarchy (Appendix~\ref{app:amrex}). Root tiles are given in finest-level cells. The AMReX tile equals $32\times32$ level-0 cells.}
\label{tab:training-specs}
\small
\setlength{\tabcolsep}{5pt}
\begin{tabular}{@{}llll@{}}
 \toprule
  & \textbf{PLI} & \textbf{TRL} & \textbf{AMReX} \\
 \midrule
 \multicolumn{4}{@{}l}{\emph{Backbone (shared)}} \\
 Layers / heads / per-head $d_h$ & \multicolumn{3}{c}{$6$ / $12$ / $192$} \\
 Transformer parameters & \multicolumn{3}{c}{$286.7$M} \\
 Complete parameters (uniform) & $290.2$M & $287.5$M & $407.1$M \\
 Complete parameters (multi-scale) & $339.0$M & $292.0$M & -- \\
 \midrule
 \multicolumn{4}{@{}l}{\emph{Tokenizer}} \\
 Input channels & $10$ & $4$ & $17$ \\
 Window $(T_\text{in} \to T_\text{out})$ & $2 \to 1$ ($p_t = 2$) & $4 \to 1$ ($p_t = 4$) & $2 \to 1$ ($p_t = 2$) \\
 Patch size $(p_h, p_w)$ & $(10, 5)$ & $(4, 4)$ & $(32, 32)$ \\
 Root tile $(\tau_h, \tau_w)$ & $(80, 40)$ & $(16, 16)$ & $(512, 512)$ \\
 Levels $(L+1)$ & $4$ & $3$ & $5$ \\
 Adaptation channels & $9$ of $10$ & $4$ of $4$ & HRR \\
 Tolerance $\varepsilon$ & $0.01$ & $0.1$ & $0.01$ \\
 Refinement dilation (hops) & \multicolumn{3}{c}{$2$} \\
 \midrule
 \multicolumn{4}{@{}l}{\emph{3D RoPE}} \\
 Axis head dims $(d_x, d_y, d_\ell)$ & $(86, 86, 20)$ & $(88, 88, 16)$ & $(84, 84, 24)$ \\
 Axis scales $(s_x, s_y, s_\ell)$ & $(80, 224, 1)$ & $(96, 32, 1)$ & $(768, 128, 1)$ \\
 \midrule
 \multicolumn{4}{@{}l}{\emph{Optimization}} \\
 Optimizer & \multicolumn{3}{c}{AdamW, clip $1.0$} \\
 Learning rate & \multicolumn{3}{c}{$10^{-4}$} \\
 LR schedule & \multicolumn{3}{c}{cosine, $1000$-step warmup, min $10^{-6}$} \\
 \midrule
 \multicolumn{4}{@{}l}{\emph{Training loop}} \\
 Micro-batch per GPU; precision & \multicolumn{3}{c}{$1$ (variable $N$); BF16 autocast} \\
 Effective batch & \multicolumn{3}{c}{$8$} \\
 Push-forward unroll $K$ & $1$ & $1$ & $4$ \\
 Epochs & $10$ & $200$ & $600$ \\
 EMA & \multicolumn{3}{c}{$0.9999$ decay, $1000$-step warmup, $\gamma_{\mathrm{inv}}=1.0$, power $2/3$} \\
 \midrule
 \multicolumn{4}{@{}l}{\emph{Hardware}} \\
 GPUs & \multicolumn{3}{c}{$8$ H100} \\
 \bottomrule
 \end{tabular}
\end{table}

\begin{table}[H]
\centering
\caption{SwinV2 specifications for the regular-grid architecture-family control. The four conventional shifted-window stages approximately match WAMRViT's transformer capacity. The refined FPN decoder uses two ConvNeXt blocks per non-bottleneck skip and a learned $3\times3$ correction after bilinear upsampling, inspired by Poseidon~\citep{herde2024poseidon}. On AMReX, the hierarchy is projected onto the uniform grid of AMR level $2$ ($1024\times6144$), four times coarser per direction than the finest level.}
\label{tab:swin-specs}
\small
\setlength{\tabcolsep}{5pt}
\begin{tabular}{@{}llll@{}}
 \toprule
  & \textbf{PLI} & \textbf{TRL} & \textbf{AMReX} \\
 \midrule
 \multicolumn{4}{@{}l}{\emph{Backbone and decoder}} \\
 Stage depths & \multicolumn{3}{c}{$[8,\,12,\,12,\,6]$} \\
 Heads per stage & \multicolumn{3}{c}{$[6,\,12,\,24,\,48]$} \\
 Transformer-stage parameters & \multicolumn{3}{c}{$286.5$M} \\
 Embedding dim / decoder channels & \multicolumn{3}{c}{$192$ / $256$} \\
 Decoder refinement blocks & \multicolumn{3}{c}{$2$} \\
 Complete model parameters & $294.2$M & $292.9$M & $334.5$M \\
 \midrule
 \multicolumn{4}{@{}l}{\emph{Tokenizer and attention}} \\
 Input channels & $10$ & $4$ & $17$ \\
 Window $(T_\text{in} \to T_\text{out})$ & $2 \to 1$ ($p_t = 2$) & $4 \to 1$ ($p_t = 4$) & $2 \to 1$ ($p_t = 2$) \\
 Patch size $(p_h, p_w)$ & $(10, 5)$ & $(4, 4)$ & $(32, 32)$ \\
 Image size $(H, W)$ & $(1120, 400)$ & $(128, 384)$ & $(1024, 6144)$ \\
 Attention window $w$ & $2$ & $2$ & $8$ \\
 \bottomrule
 \end{tabular}
\end{table}

\paragraph{Micro-batching for adaptive runs.}
\label{app:micro-batch}
Adaptive tokenization yields a sample-dependent token count, so two adaptive samples cannot be stacked into a single tensor. Two work-arounds exist: padding within a micro-batch to a common $N_{\max}$ with attention masked on the padding slots, or block-diagonal packing with a custom attention mask. We adopt neither and instead train at micro-batch one with data-parallel replication over eight GPUs, for an effective batch size of eight. The ViT-finest, ViT-mid, and SwinV2 baselines use the same configuration. Padded and packed alternatives are left to future work (Appendix~\ref{app:future-work}).

\paragraph{Objective function.}
All reported model variants use an absolute-$L_2$ objective, with per-leaf reduction for uniform-patch WAMRViT and full-field reduction for the uniform-grid baselines and multi-scale WAMRViT. Section~\ref{sec:objectives} defines both objectives. Their tensor layouts, numerical stabilizer, and push-forward recursion are detailed below.

Let $\hat{y}^{(k)},\, y^{(k)}$ denote the prediction and ground truth at push-forward step $k$. For uniform-patch WAMRViT, the loss tensor shape is $(N_t, C, T_\text{out}, p_h, p_w)$, with $N_t$ the leaf count of the current topology and per-leaf flattened volume $M_p = T_\text{out}\,p_h\,p_w$ entering the inner $L_2$ of Eq.~\eqref{eq:loss}. For the uniform-grid baselines (ViT-mid, ViT-finest, and SwinV2), the output head reassembles patches into the full field before the loss, so the loss tensor shape is $(1, C, T_\text{out}, H, W)$. Equation~\eqref{eq:loss-dense} suppresses the singleton leading axis and takes the norm over the full $T_\text{out}\times H\times W$ spatiotemporal volume of one channel. Both objectives adapt the $L_p$ loss of the NeuralOperator library \citep{li2021fno,kossaifi2025library} with $d\!=\!3$, $p\!=\!2$, adding a numerical stabilizer $\epsilon = 10^{-4}$ inside the square root of the inner $L_2$. We use the \emph{absolute} variant of $L_2$ rather than the \emph{relative} default of the same library. The relative variant normalizes by $\|y\|_2$ and degrades on regions where the ground truth is near zero.

For uniform-patch WAMRViT, push-forward training averages the per-leaf losses over the $K$ unrolled steps,
\begin{equation}
   \mathcal{L}(\theta) \;=\; \frac{1}{K}\sum_{k=1}^{K} \ell\!\bigl(\hat y^{(k)},\, y^{(k)}\bigr),
   \qquad
   \hat y^{(k)} = f_\theta\!\bigl(\hat y^{(k-T_\text{in}):(k-1)}\bigr),
 \end{equation}
with $\hat y^{(j)} = x_{t+j}$ for $j \in \{-T_\text{in}+1,\dots,0\}$, so each gradient step sees $K$ autoregressive applications of $f_\theta$. The window passed to the next unrolled step is detached, so each step's loss backpropagates only through its own model application. For the uniform-grid baselines and multi-scale variant, the same $K^{-1}$ average instead uses $\ell_{\mathrm{full}}$, evaluated after full-field reassembly or the full-field scatter described below, respectively. $K$ equals $1$ on the PLI and TRL runs and $4$ on the AMReX run (Table~\ref{tab:training-specs}).

\paragraph{Multi-scale variant: full-field loss.}
\label{app:multiscale-loss}
The multi-scale (native-resolution) variant of Appendix~\ref{app:multiscale-method} emits per-level predictions $\{\hat y^{(k)}_\ell \in \mathbb{R}^{N_\ell \times C \times T_\text{out} \times p_h r_\ell \times p_w r_\ell}\}_{\ell=0}^{L}$ with native-resolution factor $r_\ell = 2^{L-\ell}$, rather than a single tensor at a shared raster size. We do not invoke Eq.~\eqref{eq:loss} on each per-level bucket separately. Instead, the per-level predictions and targets are first scattered onto the finest uniform grid via a differentiable advanced-index assignment, producing $\hat y^{(k)}_u, y^{(k)}_u \in \mathbb{R}^{1 \times C \times T_\text{out} \times H \times W}$, and the full-field loss $\ell_{\mathrm{full}}$ of Eq.~\eqref{eq:loss-dense} is then applied to this pair. Multi-scale WAMRViT therefore uses the same full-field objective as the uniform-grid baselines. The scatter introduces no learnable parameters and admits gradients back into the per-level buckets through the assignment, so the multi-scale model trains end-to-end against the full-field loss.

\paragraph{Training and rollout orchestration.}
Training and rollout both run on the Ray distributed runtime \citep{moritz2018ray} and share the data pipeline of Appendix~\ref{app:train-tokenize}. Rollout disables topology resampling. A PyTorch DDP trainer launches one data-parallel worker per GPU, with checkpointing, EMA tracking, and logging owned by rank zero.

\subsection{Training and inference wall-clock time}
\label{app:walltime}

The per-call token count $\bar N$ that drives WAMRViT's forward time is set at inference by the regrid hyperparameters in Table~\ref{tab:regrid-inference}: the tolerance $\varepsilon$ controls refinement and coarsening, the adaptation channels restrict which fields the detector watches, and the protection radius prevents flagged neighborhoods from immediate coarsening (Appendix~\ref{app:regrid}). Inference timings cover $20$ model calls per starting window. Adaptive models additionally perform four complete regrid events at an interval of four calls. Total inference time in Table~\ref{tab:walltime} is reported as mean $\pm$ population standard deviation across the ten matched starting windows. Peak memory reports the maximum PyTorch-allocated CUDA memory and excludes non-PyTorch allocations. For TRL, the timing run is an engineering throughput benchmark, distinct from its one-step accuracy protocol. Both TRL adaptive variants use tolerance $0.1$ in all four channels for initial mesh construction and regridding, as in the full-test accuracy and qualitative evaluations. All WAMRViT models start from identical topologies, but their meshes can diverge after predictions.

\begin{table}[H]
\centering
\caption{Inference regrid settings for Table~\ref{tab:walltime}. PLI and TRL tolerances are per selected channel. AMReX uses a scalar tolerance. ``CPU array'' denotes compiled, parallel CPU kernels that update both the topology and the leaf values. For AMReX, ``GPU values'' means that only the topology is computed on the CPU. The leaf values stay on the GPU, where the resulting split/merge records are applied (Section~\ref{sec:rollout}). All regrids use coarsen ratio $\gamma=0.25$ and at most $p_{\max}=10$ passes per phase (Algorithm~\ref{alg:regrid}).}
\label{tab:regrid-inference}
\scriptsize
\setlength{\tabcolsep}{3pt}
\begin{tabular}{@{}llllll@{}}
 \toprule
 Dataset & Variant & Adaptation channels & Tolerance $\varepsilon$ & Radius $\delta$ & Execution \\
 \midrule
 PLI & WAMRViT (uniform)     & $3$ of $10$ & $[0.1,\,0.1,\,0.3]$           & $2$ & CPU array \\
     & WAMRViT (multi-scale) & $3$ of $10$ & $[0.1,\,0.1,\,0.3]$           & $2$ & CPU array \\
 TRL & WAMRViT (uniform)     & $4$ of $4$  & $[0.1,\,0.1,\,0.1,\,0.1]$     & $2$ & CPU array \\
     & WAMRViT (multi-scale) & $4$ of $4$  & $[0.1,\,0.1,\,0.1,\,0.1]$     & $2$ & CPU array \\
 AMReX & WAMRViT (uniform)   & HRR          & $0.01$                          & $2$ & CPU array + GPU values \\
 \bottomrule
\end{tabular}
\end{table}

Table~\ref{tab:walltime} separates predictor, complete-regrid, and end-to-end scopes. Regridding accounts for $8.0\%/13.9\%$ of uniform/multi-scale PLI inference, $17.0\%/5.4\%$ on TRL, and $17.2\%$ on AMReX. Including those events, uniform/multi-scale WAMRViT is $3.0\times/2.6\times$ faster than ViT-finest on PLI and approximately $1.2\times$ faster for both TRL variants. Multi-scale TRL takes $0.6\%$ longer than uniform WAMRViT, so their mean totals are nearly tied under matched tolerances. On AMReX, the WAMRViT forward pass is faster than ViT-mid, but complete regridding makes total time $8.6\%$ higher. SwinV2 is fastest and uses the least peak allocated VRAM on all three datasets. Parameter matching therefore does not imply runtime or memory matching.

AMReX adaptive training remains dominated by data-pipeline materialization rather than accelerator compute. The reported uniform-patch path stores every leaf as a $17\times32\times32$ raster, producing much larger streaming blocks than PLI or TRL. For uniform-grid sources, the validated multi-scale variant computes its full-field loss after scattering predictions onto the finest source grid. Applying this objective to AMReX would require materializing the $4096\times24576$ grid, so we do not use that variant for the native-AMReX experiment. The reported AMReX training time includes materialization of the uniform-patch data blocks. Length-bucketed batching, block-diagonal packing, and a non-materializing native objective remain future work (Appendix~\ref{app:future-work}).

\subsection{Evaluation metrics}
\label{app:metrics}

Full-field evaluation is performed on PLI and TRL; AMReX uses the per-level cell-space metrics defined below. For full-field evaluation, let $\hat{y}, y \in \mathbb{R}^{C \times H \times W}$ denote the prediction and ground truth at a given rollout step on the common finest uniform grid, with $N_p = H \cdot W$ cells and $C$ physical channels. Uniform-patch WAMRViT leaves are expanded to this grid by nearest-neighbor upsampling, whereas multi-scale leaves already store their region at this resolution and are placed directly. All metrics are computed per channel $c$ and then averaged over channels.

\paragraph{RMSE.}
The root-mean-squared error for channel $c$ is
\begin{equation}
   \mathrm{RMSE}_c(\hat{y}, y)
   \;=\;
   \sqrt{\frac{1}{N_p} \sum_{j=1}^{N_p} \bigl(\hat{y}_{c,j} - y_{c,j}\bigr)^2}.
 \end{equation}

\paragraph{VRMSE.}
The variance-scaled root-mean-squared error normalizes $\mathrm{RMSE}_c$ by the spatial standard deviation of the ground truth on the same frame, as in \citet{wang2026iap} and \citet{ohana2024well}. We follow The Well's benchmark implementation, using Bessel-corrected spatial variance:
\begin{equation}
   \mathrm{VRMSE}_c(\hat{y}, y)
   \;=\;
   \sqrt{\frac{\tfrac{1}{N_p}\sum_{j=1}^{N_p} \bigl(\hat{y}_{c,j} - y_{c,j}\bigr)^2}
              {\sigma_c^2(y) + \varepsilon}},
   \qquad
   \sigma_c^2(y) = \frac{1}{N_p-1} \sum_{j=1}^{N_p} \bigl(y_{c,j} - \bar{y}_c\bigr)^2,
 \end{equation}
where $\bar{y}_c$ is the spatial mean of channel $c$ and $\varepsilon = 10^{-7}$ guards against division by zero on constant frames. Variance scaling reduces sensitivity to channel magnitude, but near-zero target variance can still produce large relative errors. The numerical stabilizer does not prevent such channels from dominating the aggregate.

\paragraph{Per-level cell-space variant.}
The per-level RMSE and VRMSE reported in Table~\ref{tab:perlevel} and Figure~\ref{fig:rollout} are computed directly on the quadtree, without projection to a uniform grid, as an equal per-leaf average. On AMReX's finest leaves, regular-grid predictions are mapped to the finer evaluation raster by piecewise-constant prolongation (nearest-neighbor repetition). Let $\mathcal{N}_\ell$ index the leaves at level $\ell$, and let each leaf contain $M=p_hp_w$ pixels under uniform-patch storage or $M=p_hp_wr_\ell^2$ pixels under native multi-scale storage, where $r_\ell=2^{L-\ell}$. Let $\hat y_{i,c},y_{i,c}\in\mathbb{R}^M$ denote the predicted and ground-truth values of channel $c$ on leaf $i$. Both storage modes use the same metric definitions, evaluated over their retained raster. At the finest level, $r_L=1$, and the raster sizes coincide. The per-leaf squared error and per-leaf ground-truth variance are
\[
   \mathrm{MSE}_{i,c}
   \;=\;
   \frac{1}{M} \sum_{j=1}^{M} \bigl(\hat y_{i,c,j} - y_{i,c,j}\bigr)^2,
   \qquad
   \sigma_{i,c}^{2}(y)
   \;=\;
   \frac{1}{M-1} \sum_{j=1}^{M} \bigl(y_{i,c,j} - \bar y_{i,c}\bigr)^2,
 \]
and the reported per-level metrics are the mean of per-leaf metrics across the $|\mathcal{N}_\ell|$ leaves at level $\ell$,
\begin{equation}
   \mathrm{RMSE}_{c,\ell}
   \;=\;
   \frac{1}{|\mathcal{N}_\ell|}
   \sum_{i \in \mathcal{N}_\ell}
   \sqrt{\mathrm{MSE}_{i,c}},
   \qquad
   \mathrm{VRMSE}_{c,\ell}
   \;=\;
   \frac{1}{|\mathcal{N}_\ell|}
   \sum_{i \in \mathcal{N}_\ell}
   \sqrt{\frac{\mathrm{MSE}_{i,c}}{\sigma_{i,c}^{2}(y) + \varepsilon}}.
   \label{eq:perlevel-metrics}
 \end{equation}
The VRMSE denominator is the \emph{per-leaf} ground-truth variance $\sigma_{i,c}^{2}(y)$, computed over the $M$ pixels of leaf $i$ with the same Bessel correction, not the frame-level $\sigma_c^2(y)$. This normalization measures each leaf's error relative to its local spatial variation, and the level average gives every leaf equal weight. It is therefore a local-contrast diagnostic rather than a guarantee of balanced contributions across fields: nearly constant leaves can produce very large ratios, particularly for sparse HRR on AMReX. We retain per-field RMSE and VRMSE breakdowns to expose this limitation.

\section{Additional results}
\label{app:additional_rollout}

This appendix provides first-step RMSE, aggregated and per-field rollout curves on PLI and AMReX combustion, the push-forward, random-resample, and level-scale ablations, and the PLI mesh-policy sensitivity study. Rollout curves and PLI/AMReX ablations use a fixed seed. Qualitative snapshots are in Appendix~\ref{app:qualitative-overview}.

\subsection{First-step RMSE summary}
\label{app:perlevel-rmse}

Table~\ref{tab:perlevel-rmse} reports first-step RMSE for the same evaluations as Table~\ref{tab:perlevel}. The IAP$^{*}$ column is omitted because its paper does not report RMSE. Multi-scale WAMRViT has the lowest full-field RMSE on PLI and TRL. Evaluated over the regions refined to level $L$ by its own topology, its RMSE is $22.8\%$ below the ViT-finest baseline evaluated on the same regions on PLI and $1.4\%$ below on TRL. These evaluations agree with the ViT-finest values listed in Table~\ref{tab:perlevel} at the reported precision at the first step. Refined regions can diverge during subsequent regridding. At the PLI finest level, the ViT-finest and uniform-WAMRViT RMSE means differ by less than $0.1\%$ and their seed spreads overlap. Uniform WAMRViT is $60.8\%/60.2\%$ below ViT-mid/SwinV2 on AMReX.

\begin{table}[H]
\centering
\caption{\textbf{First-rollout-step RMSE} (channel mean), counterpart of Table~\ref{tab:perlevel}. Entries are mean $\pm$ sample standard deviation over three training seeds. Training depth is $K=1$ for PLI/TRL and $K=4$ for AMReX. Finest-level metrics are evaluated on the regions refined to level $L$ in the initial adaptive topology. AMReX omits ViT-finest because its level-$4$ grid ($4096\times24576$) cannot be materialized. ViT-mid and SwinV2 use $p=32$.}
\label{tab:perlevel-rmse}
\scriptsize
\setlength{\tabcolsep}{2.6pt}
\begin{tabular}{@{}lccccc@{}}
 \toprule
 & \makecell{WAMRViT\\(uniform)} & \makecell{WAMRViT\\(multi-scale)} & \makecell{ViT\\(finest)} & \makecell{ViT\\(mid)} & SwinV2 \\
 \midrule
 PLI, full-field & \makecell{0.0110\\{\scriptsize $\pm$6.232E-5}} & \makecell{\textbf{4.963E-3}\\{\scriptsize $\pm$2.198E-5}} & \makecell{\underline{6.092E-3}\\{\scriptsize $\pm$2.338E-4}} & \makecell{6.932E-3\\{\scriptsize $\pm$4.526E-5}} & \makecell{0.0111\\{\scriptsize $\pm$6.474E-5}} \\
 PLI, finest ($\ell\!=\!3$) & \makecell{2.475E-3\\{\scriptsize $\pm$8.896E-5}} & \makecell{\textbf{1.909E-3}\\{\scriptsize $\pm$1.167E-5}} & \makecell{\underline{2.473E-3}\\{\scriptsize $\pm$1.338E-4}} & \makecell{2.858E-3\\{\scriptsize $\pm$3.869E-5}} & \makecell{5.702E-3\\{\scriptsize $\pm$4.940E-5}} \\
 \midrule
 TRL, full-field & \makecell{0.5185\\{\scriptsize $\pm$4.802E-3}} & \makecell{\textbf{0.4979}\\{\scriptsize $\pm$8.661E-4}} & \makecell{\underline{0.5069}\\{\scriptsize $\pm$1.483E-4}} & \makecell{0.5972\\{\scriptsize $\pm$3.193E-3}} & \makecell{0.7047\\{\scriptsize $\pm$3.396E-3}} \\
 TRL, finest ($\ell\!=\!2$) & \makecell{\textbf{0.3283}\\{\scriptsize $\pm$3.064E-3}} & \makecell{\underline{0.3333}\\{\scriptsize $\pm$1.103E-3}} & \makecell{0.3382\\{\scriptsize $\pm$2.818E-4}} & \makecell{0.4007\\{\scriptsize $\pm$2.046E-3}} & \makecell{0.5193\\{\scriptsize $\pm$7.857E-4}} \\
 \midrule
 AMReX, finest ($\ell\!=\!4$) & \makecell{\textbf{1.202E10}\\{\scriptsize $\pm$8.293E8}} & / & / & \makecell{3.064E10\\{\scriptsize $\pm$1.818E8}} & \makecell{\underline{3.022E10}\\{\scriptsize $\pm$1.498E8}} \\
 \bottomrule
\end{tabular}
\end{table}

\subsection{Aggregated rollout curves on PLI}
\label{app:perfield-pli}

All panels are at push-forward $K\!=\!1$, averaged over all evaluation windows, sharing the legend of the PLI panels of Figure~\ref{fig:rollout}. Only the ViT-finest series paired with uniform-patch WAMRViT is plotted. Its cell-space errors use uniform WAMRViT's evolving topology, while multi-scale WAMRViT retains its own topology. Paired numerical comparisons with multi-scale WAMRViT use its separately evaluated ViT-finest reference, which is not plotted. Intermediate-level ($\ell\!\in\!\{1,2\}$) breakdowns are omitted for readability.

\begin{figure}[H]
\centering
\begin{subfigure}{0.48\linewidth}
\includegraphics[width=\linewidth]{figs/pli_pf1_agg__uniform_VRMSE.png}
\caption{Full-field VRMSE.}
\end{subfigure}\hfill
\begin{subfigure}{0.48\linewidth}
\includegraphics[width=\linewidth]{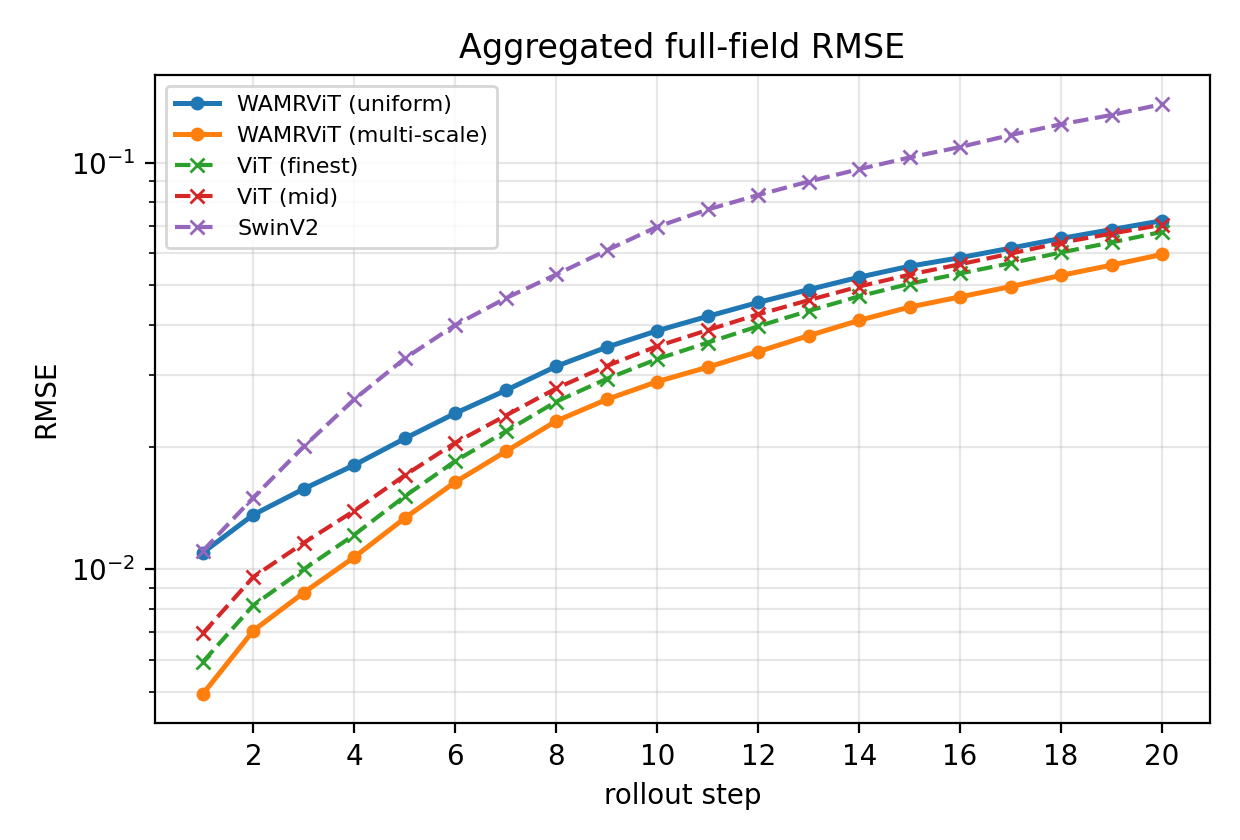}
\caption{Full-field RMSE.}
\end{subfigure}
\caption{\textbf{Aggregated full-field error on PLI}, $K\!=\!1$.}
\label{fig:app-pli-agg-full}
\end{figure}

\begin{figure}[H]
\centering
\begin{subfigure}{0.48\linewidth}
\includegraphics[width=\linewidth]{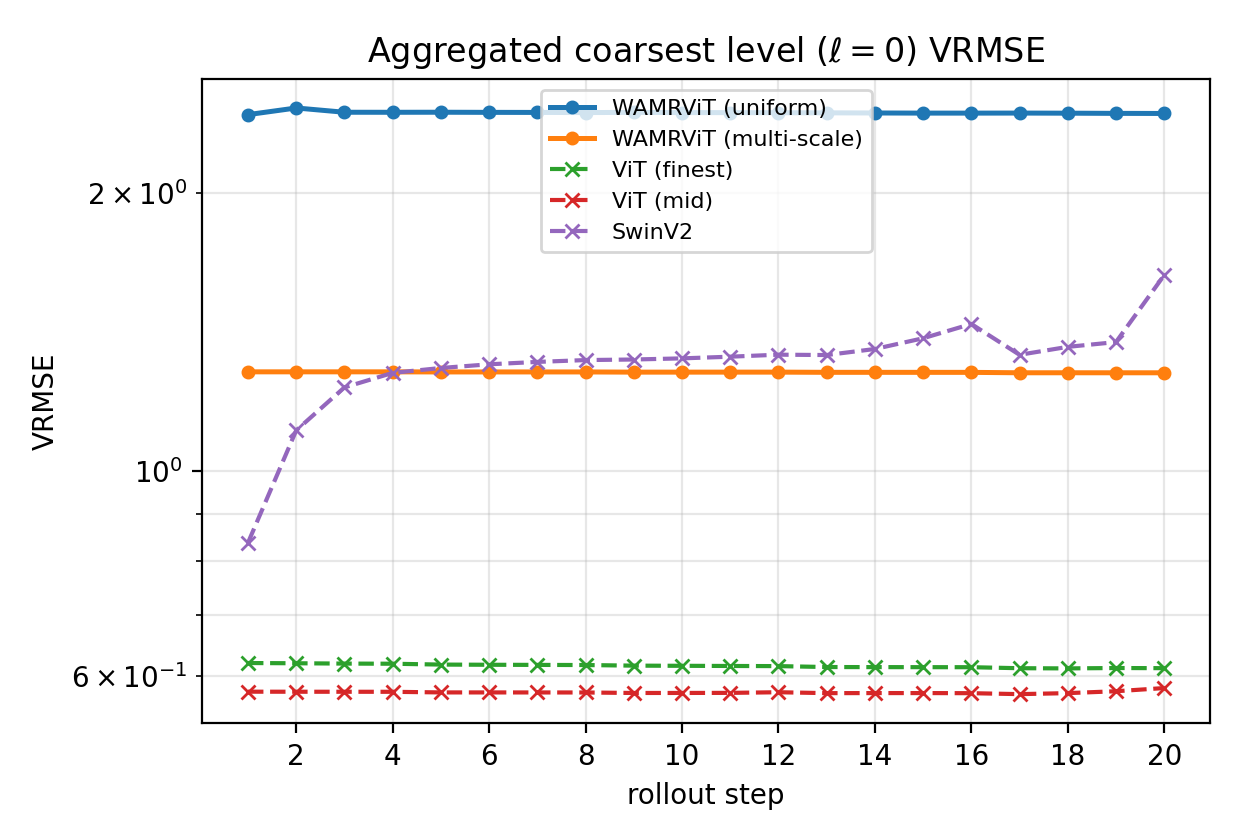}
\caption{$\ell\!=\!0$ VRMSE.}
\end{subfigure}\hfill
\begin{subfigure}{0.48\linewidth}
\includegraphics[width=\linewidth]{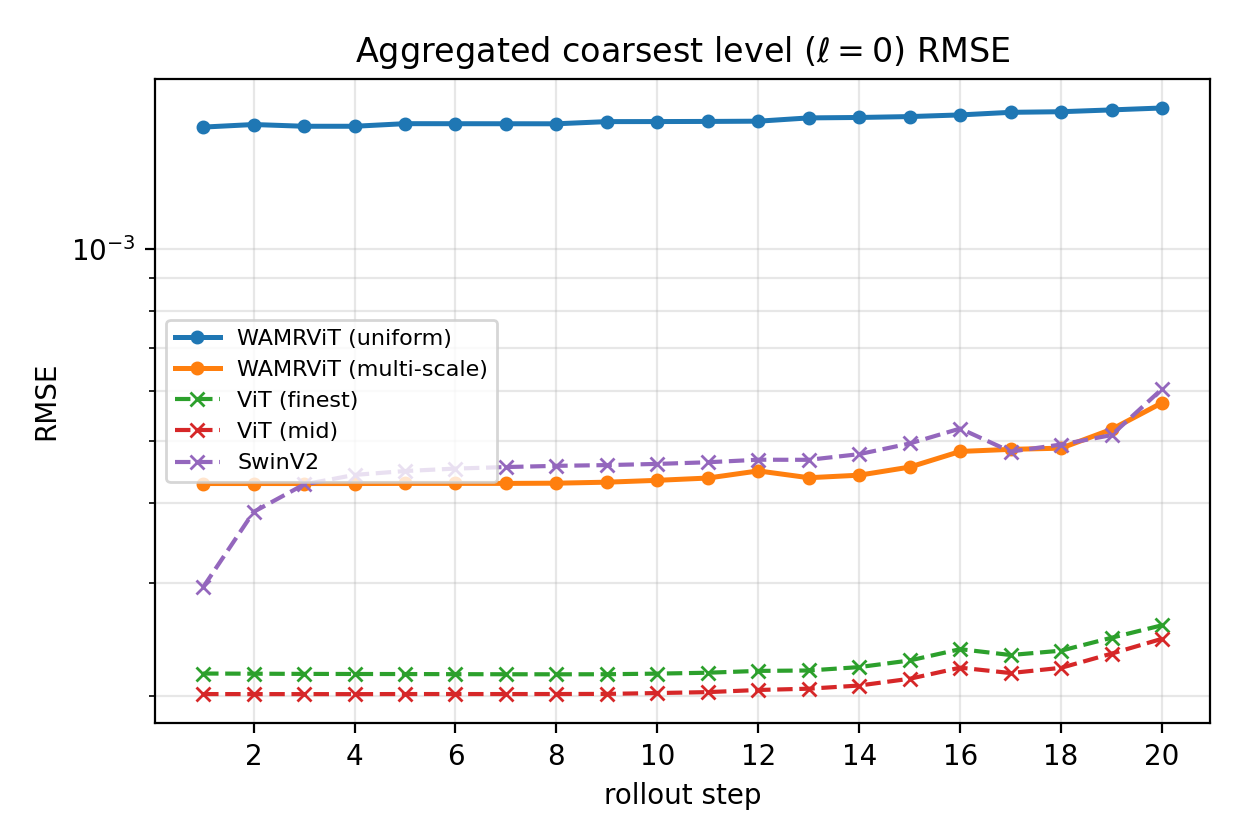}
\caption{$\ell\!=\!0$ RMSE.}
\end{subfigure}
\caption{\textbf{Aggregated coarsest-level error on PLI}, $K\!=\!1$. Each coarsest leaf spans $8^2$ times the finest-leaf area. Uniform-patch WAMRViT stores $50$ pixels per leaf, whereas multi-scale WAMRViT retains an $80\times40$ raster. The displayed regular-grid controls are block-mean projected onto uniform WAMRViT's leaf rasters.}
\label{fig:app-pli-agg-coarse}
\end{figure}

\begin{figure}[H]
\centering
\begin{subfigure}{0.48\linewidth}
\includegraphics[width=\linewidth]{figs/pli_pf1_agg__level_3_VRMSE.png}
\caption{Finest ($\ell\!=\!3$) VRMSE.}
\end{subfigure}\hfill
\begin{subfigure}{0.48\linewidth}
\includegraphics[width=\linewidth]{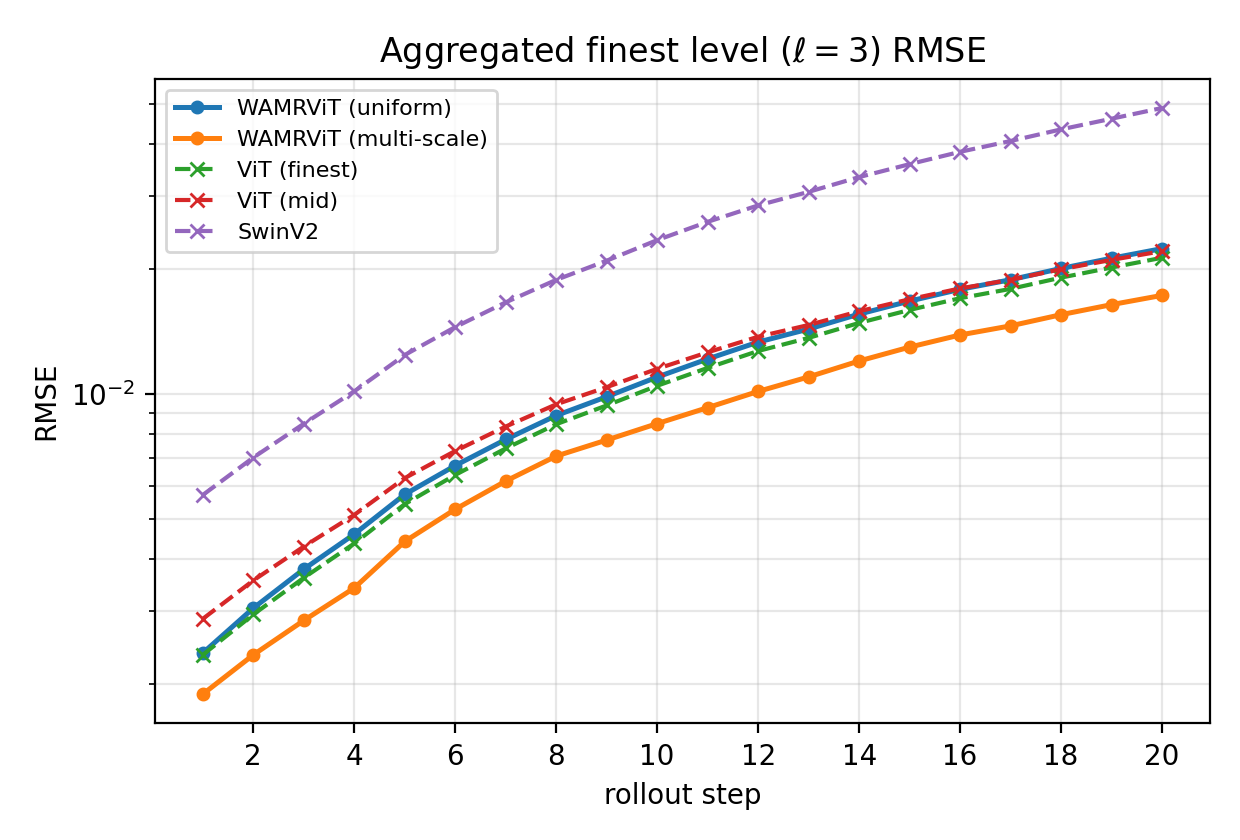}
\caption{Finest ($\ell\!=\!3$) RMSE.}
\end{subfigure}
\caption{\textbf{Aggregated finest-level error on PLI}, $K\!=\!1$.}
\label{fig:app-pli-agg-finest}
\end{figure}

\subsection{Push-forward ablation on PLI}
\label{app:pf-ablation}

\begin{figure}[H]
\centering
\begin{subfigure}{0.48\linewidth}
\includegraphics[width=\linewidth]{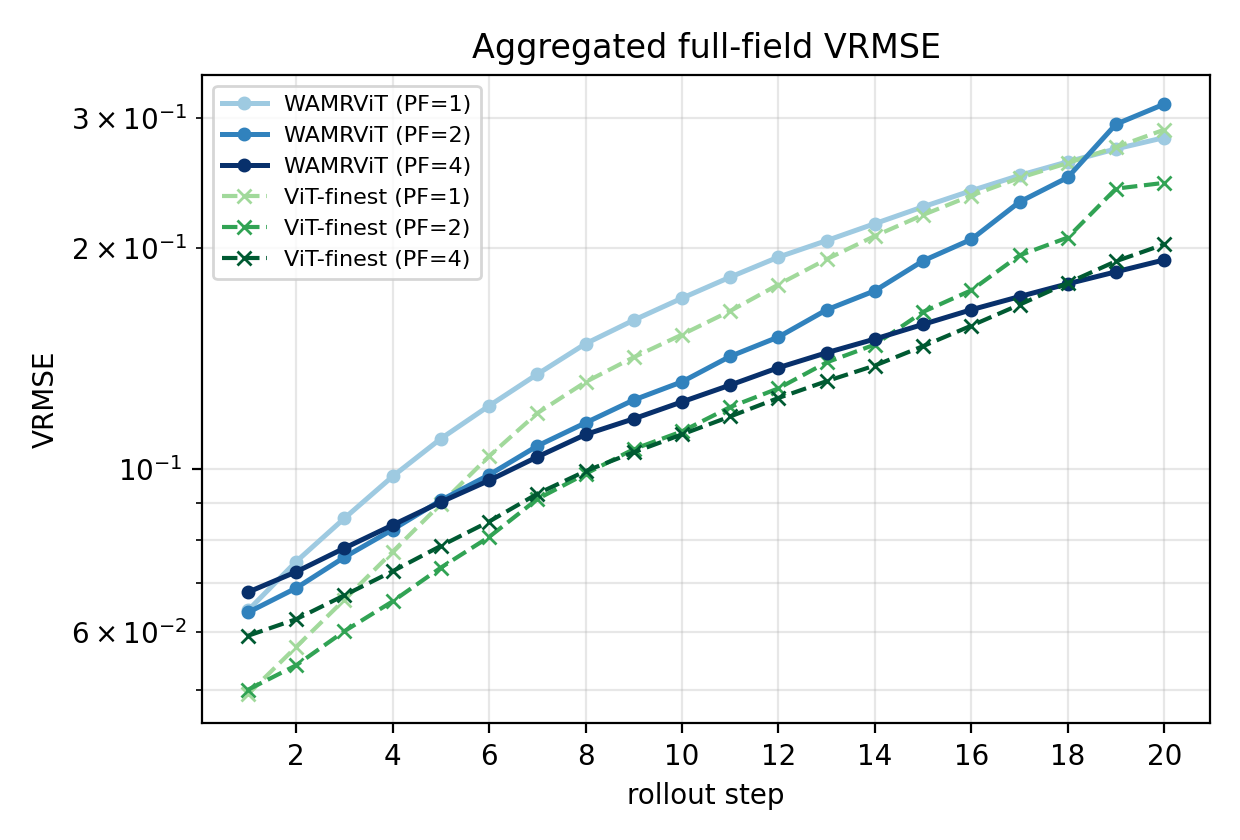}
\caption{Full-field VRMSE.}
\end{subfigure}\hfill
\begin{subfigure}{0.48\linewidth}
\includegraphics[width=\linewidth]{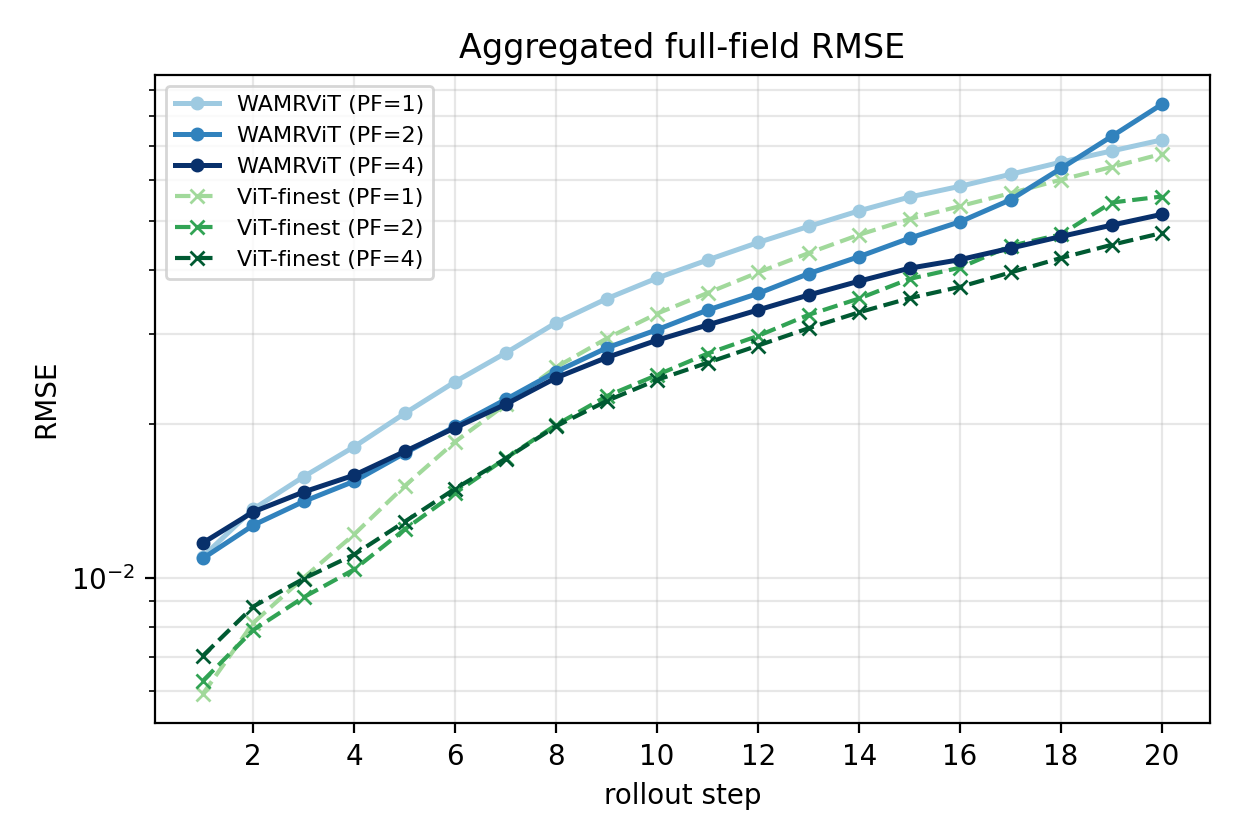}
\caption{Full-field RMSE.}
\end{subfigure}

\vspace{0.4em}

\begin{subfigure}{0.48\linewidth}
\includegraphics[width=\linewidth]{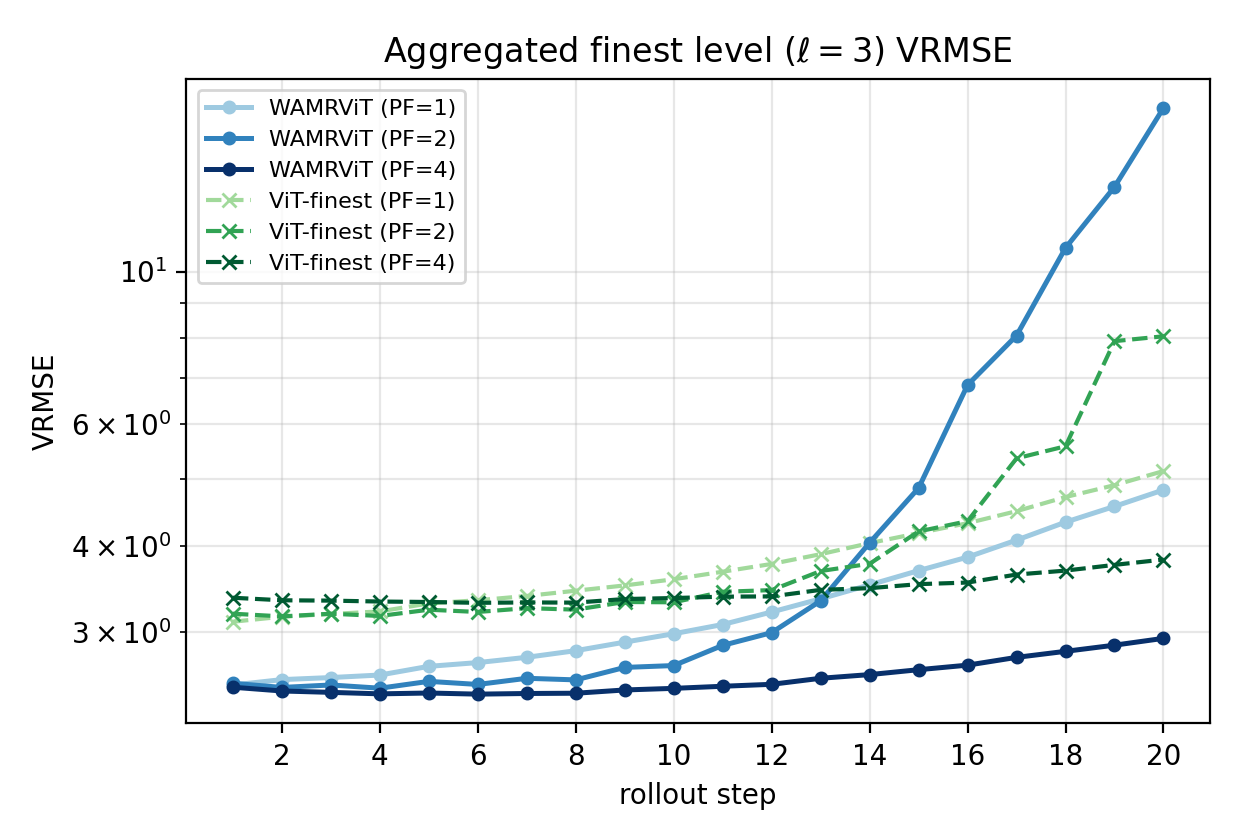}
\caption{Finest ($\ell\!=\!3$) VRMSE.}
\end{subfigure}\hfill
\begin{subfigure}{0.48\linewidth}
\includegraphics[width=\linewidth]{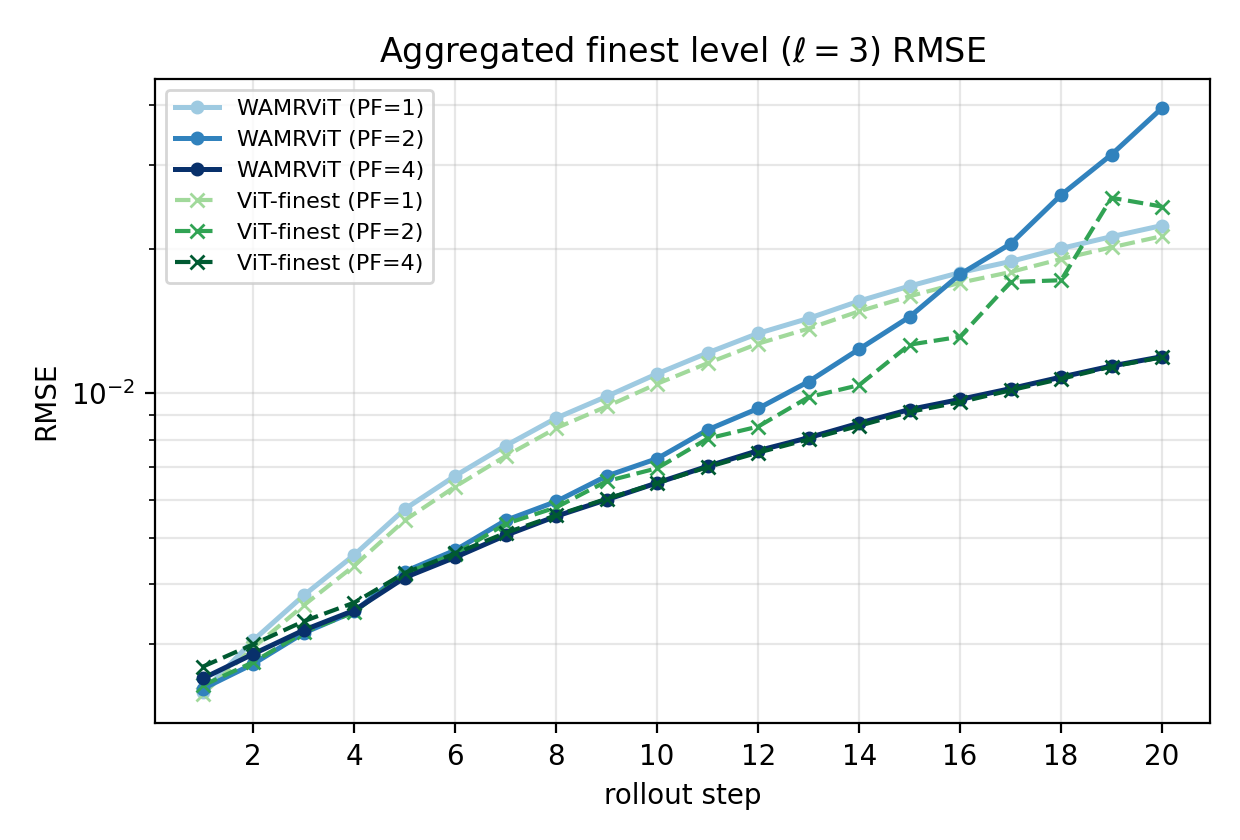}
\caption{Finest ($\ell\!=\!3$) RMSE.}
\end{subfigure}

\caption{\textbf{Push-forward ablation on PLI.} WAMRViT (solid, blue family) and the ViT-finest baseline (dashed, green family) at $K \in \{1, 2, 4\}$, with light-to-dark shading for $K = 1 \to 4$ within each family. Top row: full-field aggregate. Bottom row: finest refinement level ($\ell = 3$).}
\label{fig:app-pli-pf-ablation}
\end{figure}

WAMRViT (uniform tokenization) and the ViT-finest baseline are each trained at three push-forward depths $K \in \{1, 2, 4\}$ on PLI, with all other hyperparameters held fixed to the values in Table~\ref{tab:training-specs} (which uses $K = 1$ as the PLI main-text setting). First-step VRMSE numbers are summarized in Table~\ref{tab:pf-step1}. Aggregated full-field and finest-level rollout curves averaged over all evaluation windows are shown in Figure~\ref{fig:app-pli-pf-ablation}.

\begin{table}[H]
\centering
\caption{First-step VRMSE on PLI under the push-forward sweep, channel-mean over the ten PLI fields.}
\label{tab:pf-step1}
\begin{tabular}{@{}lcccc@{}}
\toprule
 & \multicolumn{2}{c}{full-field} & \multicolumn{2}{c}{finest ($\ell\!=\!3$)} \\
\cmidrule(lr){2-3}\cmidrule(lr){4-5}
$K$ & WAMRViT & ViT-finest & WAMRViT & ViT-finest \\
\midrule
$1$ & \underline{0.0642} & \textbf{0.0494} & \underline{2.5107} & \textbf{3.1023} \\
$2$ & \textbf{0.0638} & \underline{0.0501} & 2.5252 & \underline{3.1863} \\
$4$ & 0.0680 & 0.0593 & \textbf{2.4921} & 3.3610 \\
\bottomrule
\end{tabular}
\end{table}

At full field, $K=4$ improves both models at long horizon: at rollout step $20$, WAMRViT and ViT-finest reach $0.1925$ and $0.2021$ VRMSE, respectively. Averaged over all $20$ steps at $K=4$, however, WAMRViT's full-field VRMSE is $7.0\%$ higher than ViT-finest's. At the finest level, WAMRViT is $6.1\%$ below ViT-finest at rollout step $20$ for $K=1$ and $23.2\%$ below it for $K=4$. The intermediate $K=2$ adaptive run is unstable at this scope. The effect of push-forward depth depends on the evaluation scope and horizon. At step $20$, WAMRViT's finest-level advantage holds for $K=1$ and $K=4$, but reverses at $K=2$.

\subsection{Random-resample ablation}
\label{app:resample-ablation}

WAMRViT (uniform tokenization) is trained with and without the random-resample augmentation of Appendix~\ref{app:resample} on both PLI and AMReX combustion, all other hyperparameters held fixed to the values in Table~\ref{tab:training-specs}.

\begin{figure}[H]
\centering
\begin{subfigure}{0.48\linewidth}
\includegraphics[width=\linewidth]{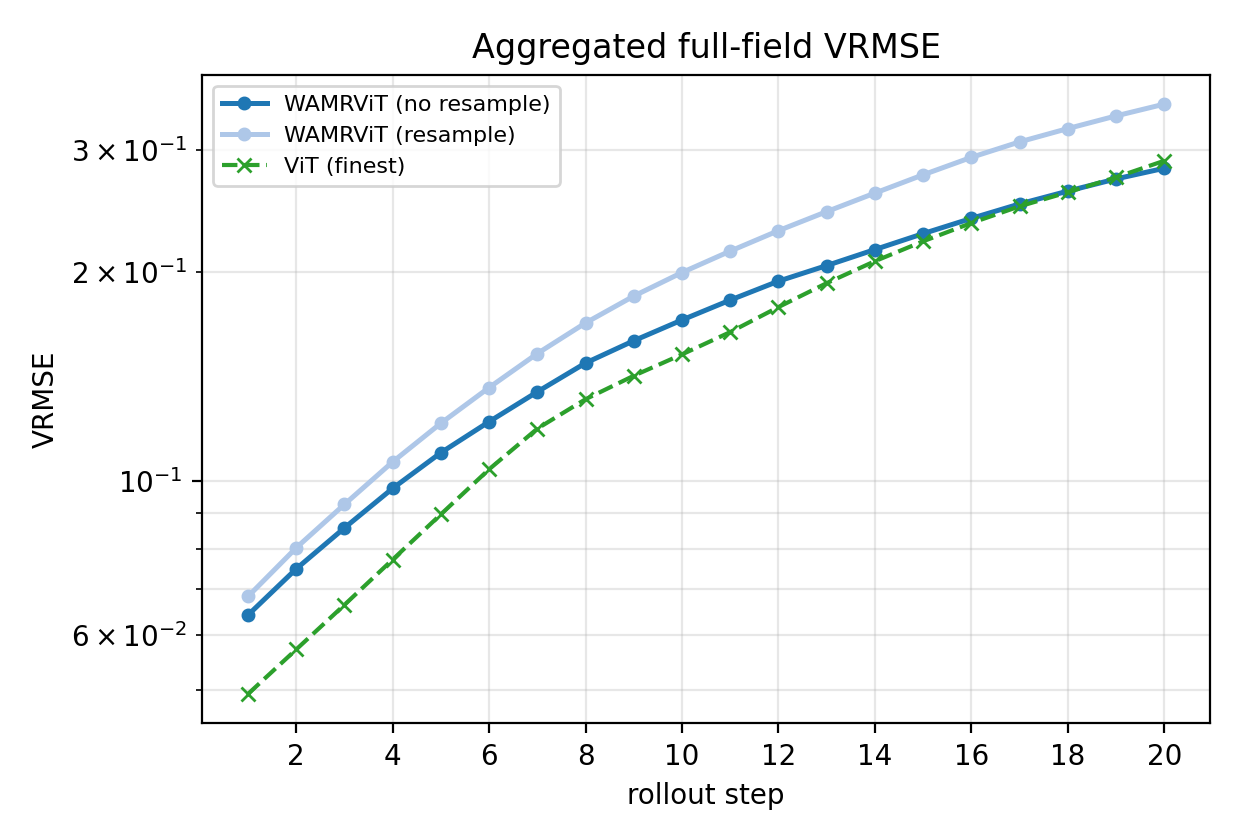}
\caption{Full-field VRMSE.}
\end{subfigure}\hfill
\begin{subfigure}{0.48\linewidth}
\includegraphics[width=\linewidth]{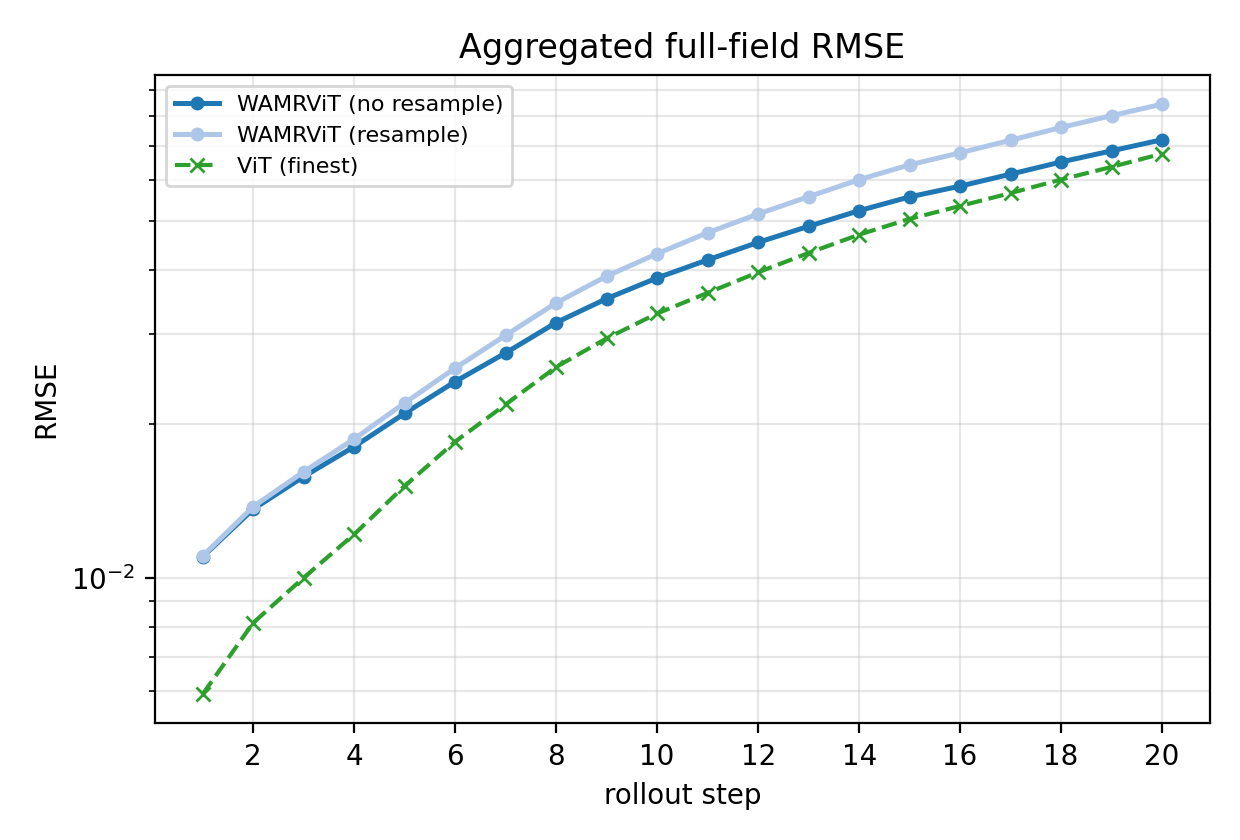}
\caption{Full-field RMSE.}
\end{subfigure}

\vspace{0.4em}

\begin{subfigure}{0.48\linewidth}
\includegraphics[width=\linewidth]{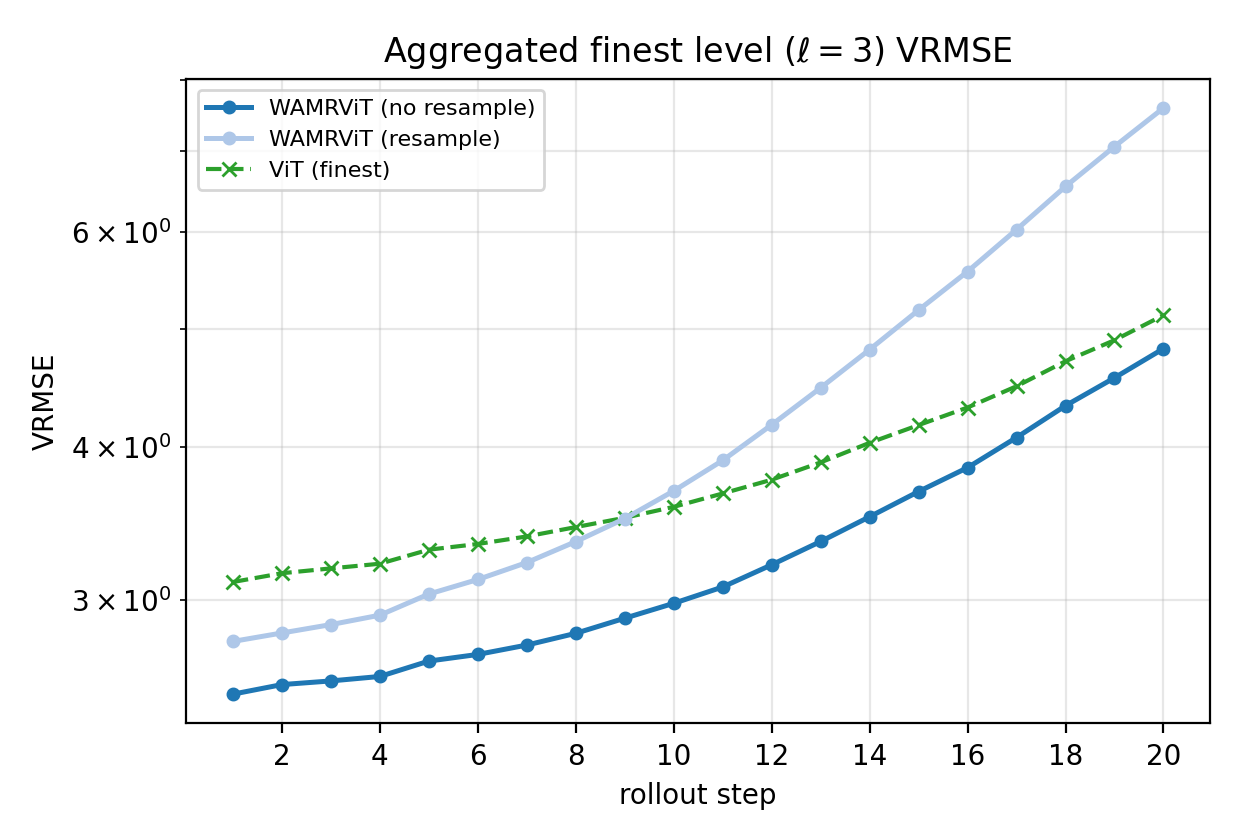}
\caption{Finest ($\ell\!=\!3$) VRMSE.}
\end{subfigure}\hfill
\begin{subfigure}{0.48\linewidth}
\includegraphics[width=\linewidth]{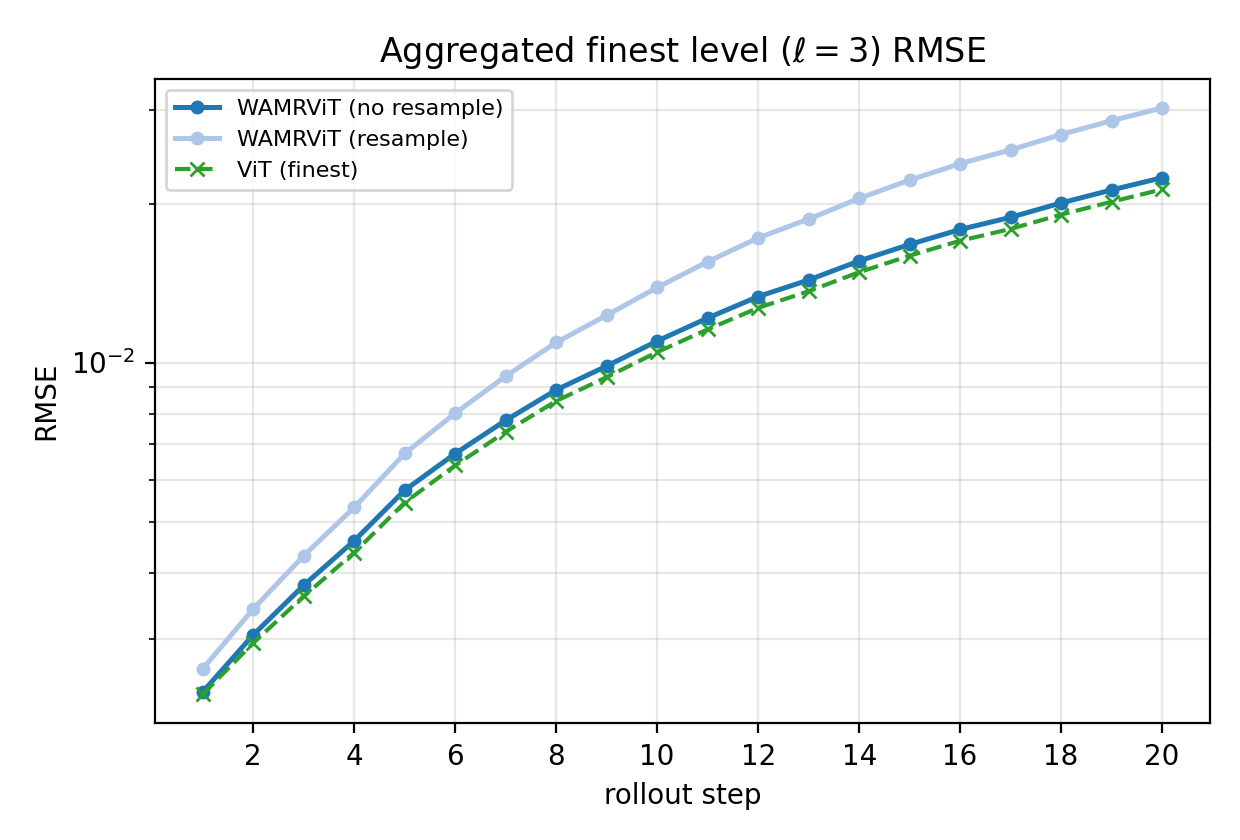}
\caption{Finest ($\ell\!=\!3$) RMSE.}
\end{subfigure}

\caption{\textbf{Random-resample ablation on PLI.} WAMRViT (uniform tokenization) trained without the augmentation (solid, dark blue, the main-text default) vs.\ with the augmentation of Appendix~\ref{app:resample} (solid, light blue), at $K\!=\!1$, against the ViT-finest baseline (dashed, green). Top row: full-field aggregate. Bottom row: finest refinement level ($\ell\!=\!3$).}
\label{fig:app-pli-resample-ablation}
\end{figure}

\paragraph{PLI.} First-step VRMSE numbers are summarized in Table~\ref{tab:resample-pli-step1}. Aggregated full-field and finest-level rollout curves averaged over all evaluation windows are shown in Figure~\ref{fig:app-pli-resample-ablation}.

\begin{table}[H]
\centering
\caption{First-step VRMSE on PLI under the random-resample augmentation, channel-mean over the ten PLI fields.}
\label{tab:resample-pli-step1}
\begin{tabular}{@{}lcc@{}}
\toprule
Variant & full-field & finest ($\ell\!=\!3$) \\
\midrule
WAMRViT (no resample, default) & \textbf{0.0642} & \textbf{2.5107} \\
WAMRViT (with resample)        & 0.0683 & 2.7734 \\
\bottomrule
\end{tabular}
\end{table}

On PLI, resampling degrades both scopes. Relative to the default, its first-step VRMSE is $6.4\%$ higher at full field and $10.5\%$ higher at the finest level. Its rollout-mean values are $18.5\%$ and $32.2\%$ higher, respectively. The corresponding RMSE curves give the same ordering. Thus this augmentation does not provide an accuracy gain on PLI at $K=1$.

\begin{figure}[H]
\centering
\begin{subfigure}{0.48\linewidth}
\includegraphics[width=\linewidth]{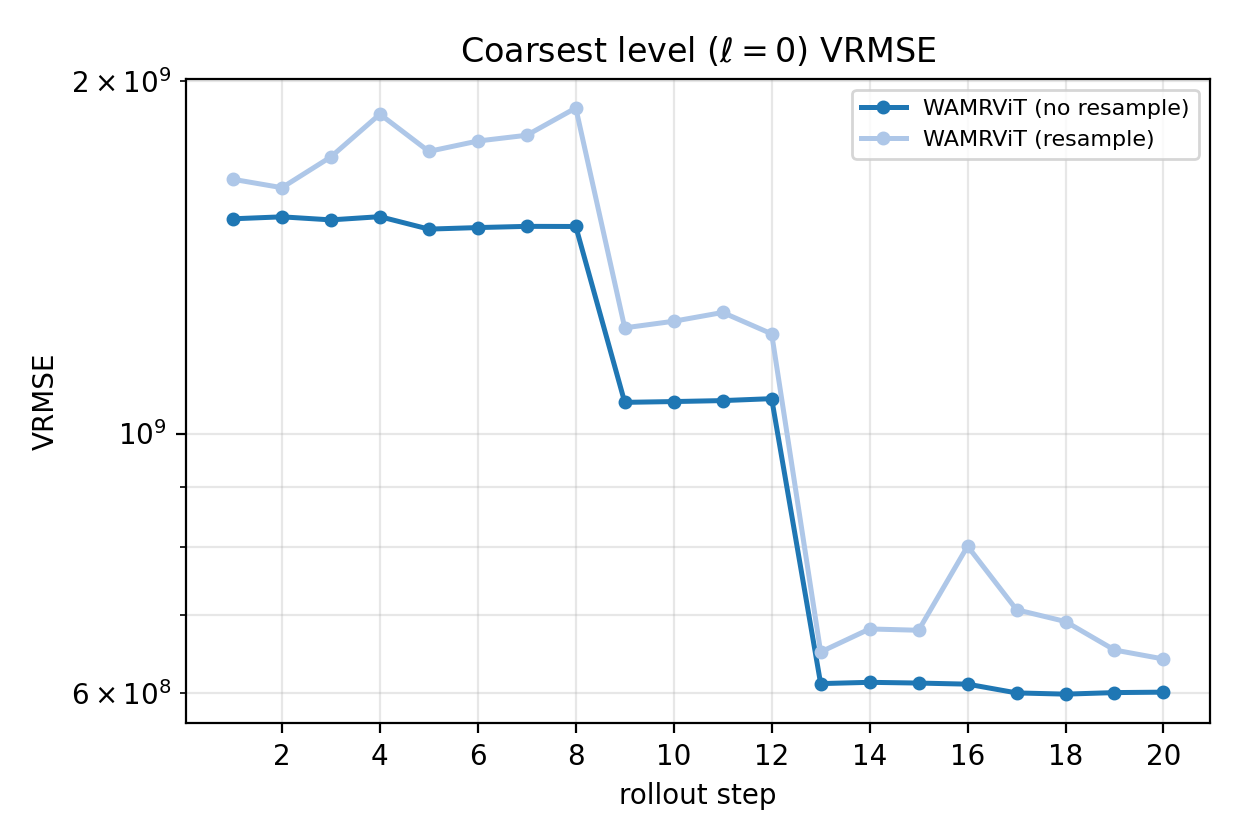}
\caption{Coarsest ($\ell\!=\!0$) VRMSE.}
\end{subfigure}\hfill
\begin{subfigure}{0.48\linewidth}
\includegraphics[width=\linewidth]{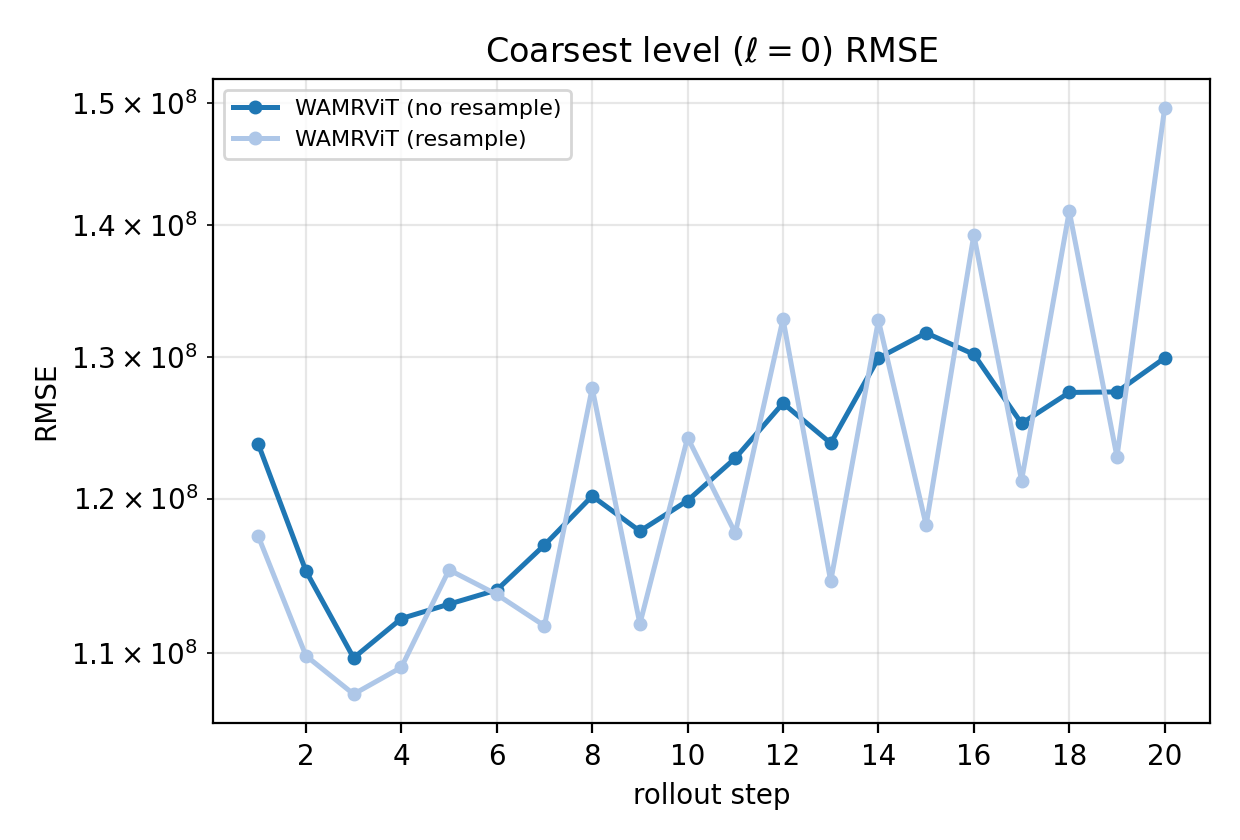}
\caption{Coarsest ($\ell\!=\!0$) RMSE.}
\end{subfigure}

\vspace{0.4em}

\begin{subfigure}{0.48\linewidth}
\includegraphics[width=\linewidth]{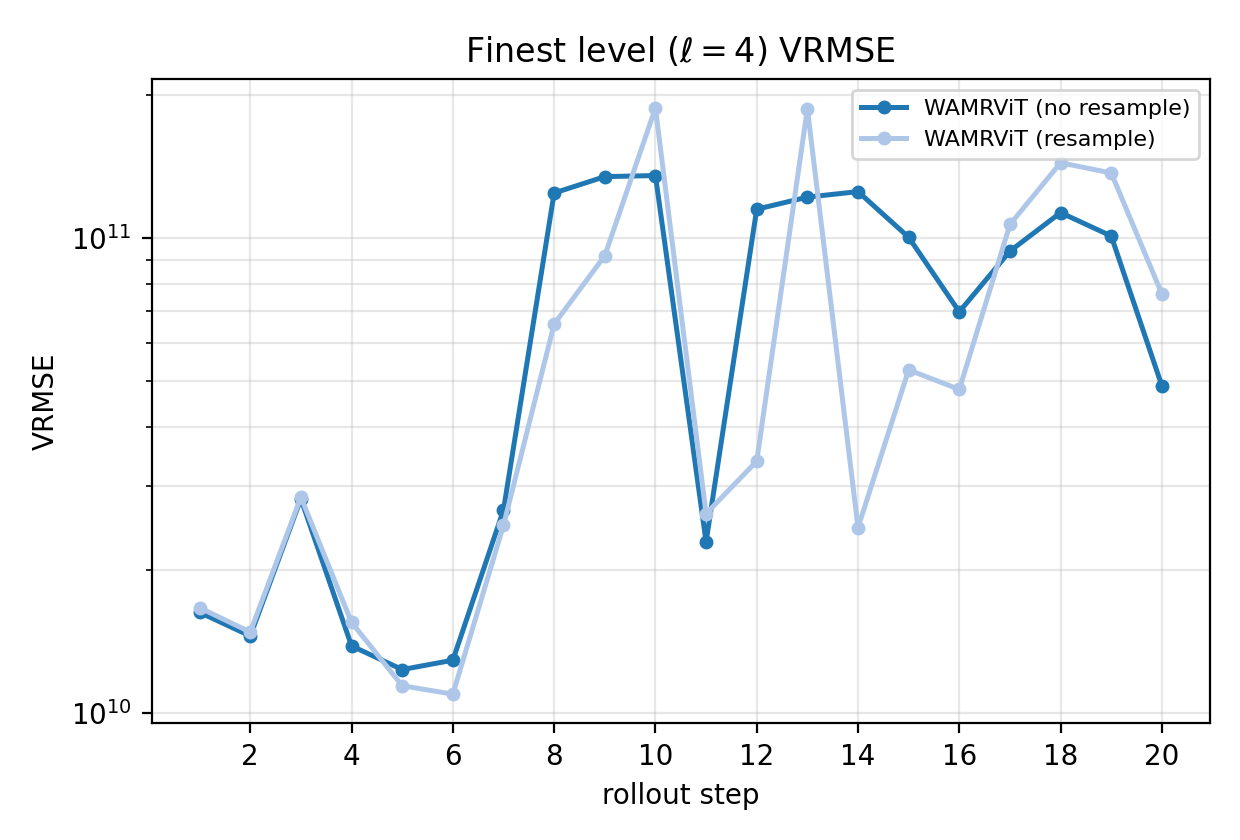}
\caption{Finest ($\ell\!=\!4$) VRMSE.}
\end{subfigure}\hfill
\begin{subfigure}{0.48\linewidth}
\includegraphics[width=\linewidth]{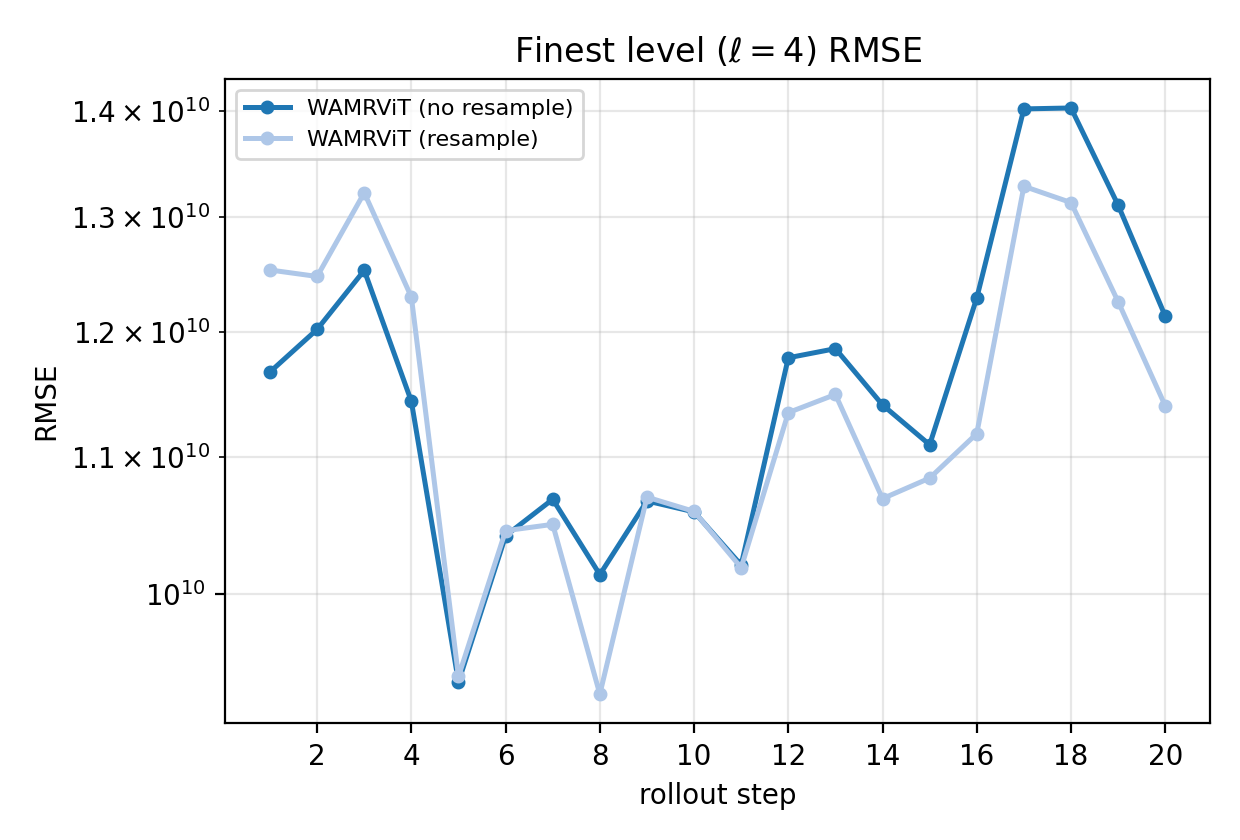}
\caption{Finest ($\ell\!=\!4$) RMSE.}
\end{subfigure}

\caption{\textbf{Random-resample ablation on AMReX combustion.} WAMRViT (uniform tokenization) trained without the augmentation (solid, dark blue, the no-resampling control) vs.\ with the augmentation (solid, light blue), at $K\!=\!4$. Top row: coarsest level ($\ell\!=\!0$). Bottom row: finest refinement level ($\ell\!=\!4$).}
\label{fig:app-amrex-resample-ablation}
\end{figure}

\paragraph{AMReX combustion.} First-step VRMSE numbers are summarized in Table~\ref{tab:resample-amr-step1}. Coarsest- and finest-level aggregated rollout curves averaged over all evaluation windows are shown in Figure~\ref{fig:app-amrex-resample-ablation}.

\begin{table}[H]
\centering
\caption{First-step VRMSE on AMReX combustion under the random-resample augmentation, channel-mean over the seventeen combustion fields. AMReX has no full-field row (Section~\ref{sec:datasets}). Resampling increases coarsest- and finest-level VRMSE by $8.1\%$ and $2.3\%$, respectively.}
\label{tab:resample-amr-step1}
\begin{tabular}{@{}lcc@{}}
\toprule
Variant & coarsest ($\ell\!=\!0$) & finest ($\ell\!=\!4$) \\
\midrule
WAMRViT (no resample, default) & \textbf{1.525E9} & \textbf{1.631E10} \\
WAMRViT (with resample)        & 1.648E9 & 1.668E10 \\
\bottomrule
\end{tabular}
\end{table}

On AMReX, resampling has resolution-, metric-, and horizon-dependent effects. Averaged over all $20$ rollout steps, it lowers finest-level VRMSE by $8.8\%$ and RMSE by $1.8\%$, but raises coarsest-level VRMSE by $15.7\%$. At the finest level, first-step VRMSE and RMSE increase by $2.3\%$ and $7.4\%$, respectively. Final-step VRMSE increases by $56.1\%$ even though RMSE decreases by $6.1\%$. The augmentation trades modest average finest-level gains against worse early predictions and coarsest-level VRMSE.

\subsection{RoPE level-axis scale ablation}
\label{app:level-scale-ablation}

\begin{figure}[H]
\centering
\begin{subfigure}{\linewidth}
\centering
\includegraphics[width=\linewidth]{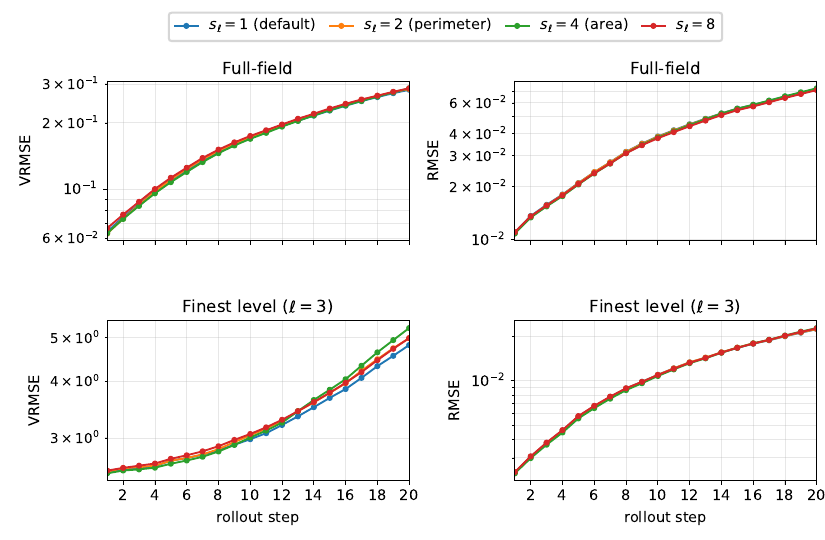}
\caption{Uniform-patch WAMRViT.}
\label{fig:app-pli-level-scale-uniform}
\end{subfigure}

\vspace{0.4em}
\begin{subfigure}{\linewidth}
\centering
\includegraphics[width=\linewidth]{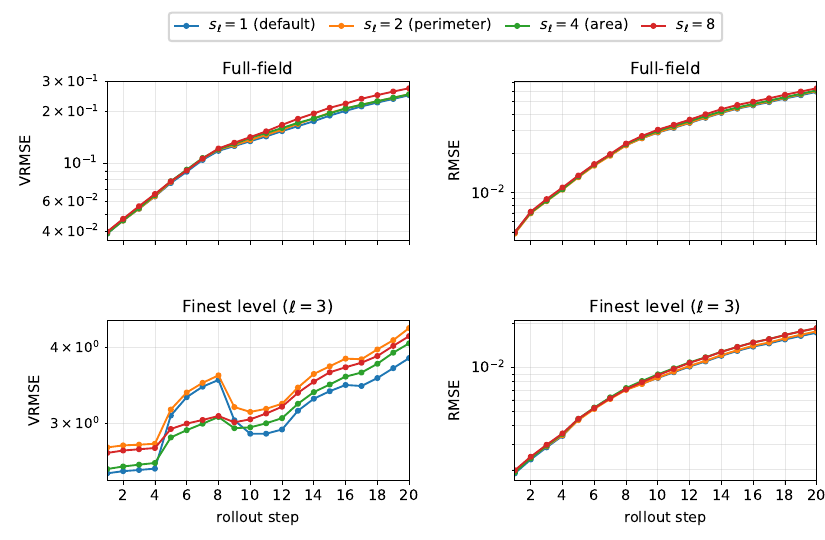}
\caption{Multi-scale WAMRViT.}
\label{fig:app-pli-level-scale-multi}
\end{subfigure}
\caption{\textbf{RoPE level-axis scale ablation on PLI.} Single-seed rollouts for (a) uniform-patch and (b) multi-scale WAMRViT at $s_\ell\in\{1,2,4,8\}$, trained without topology resampling. Within each group, rows show full-field (top) and finest-level ($\ell=3$, bottom) errors; columns show VRMSE (left) and RMSE (right). Colors denote the same scales in both groups. Finest-level metrics use each rollout's evolving topology.}
\label{fig:app-pli-level-scale-ablation}
\end{figure}

\begin{figure}[H]
\centering
\begin{subfigure}{0.48\linewidth}
\includegraphics[width=\linewidth]{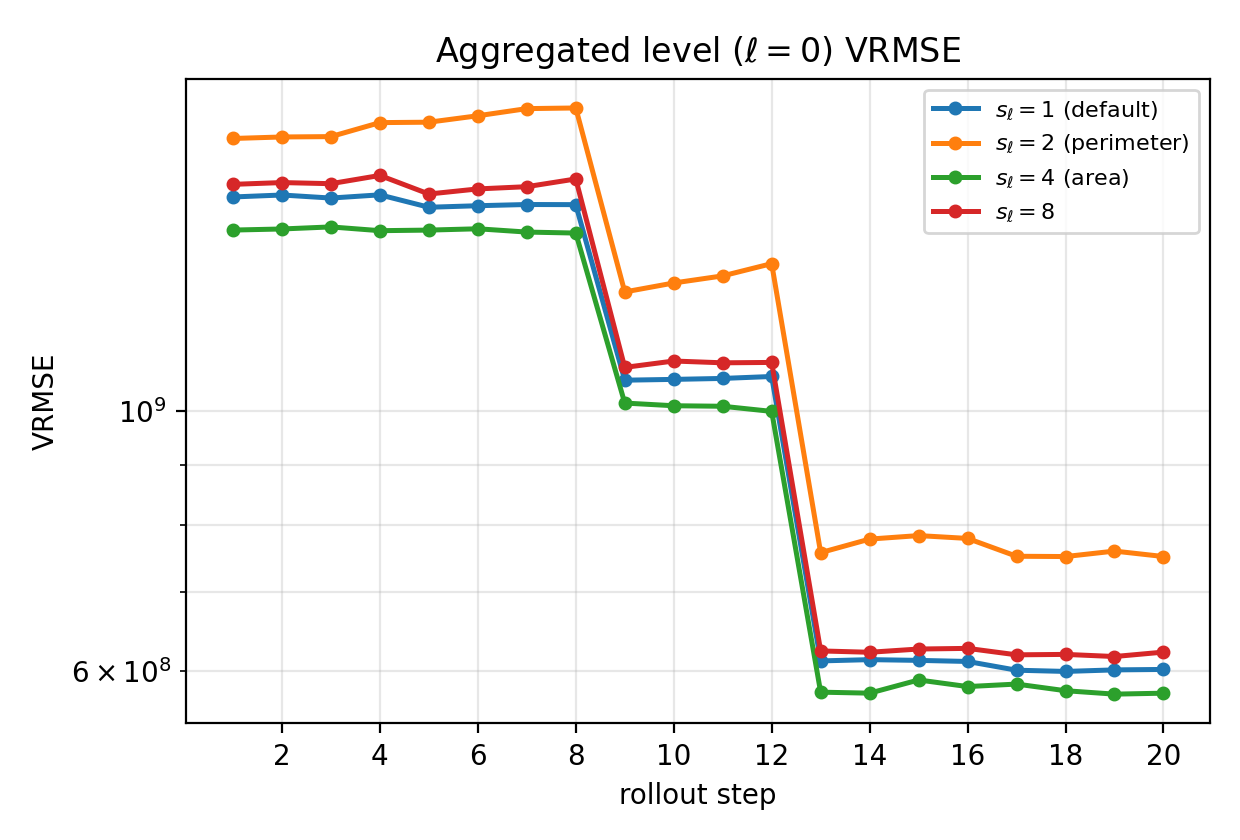}
\caption{Coarsest ($\ell\!=\!0$) VRMSE.}
\end{subfigure}\hfill
\begin{subfigure}{0.48\linewidth}
\includegraphics[width=\linewidth]{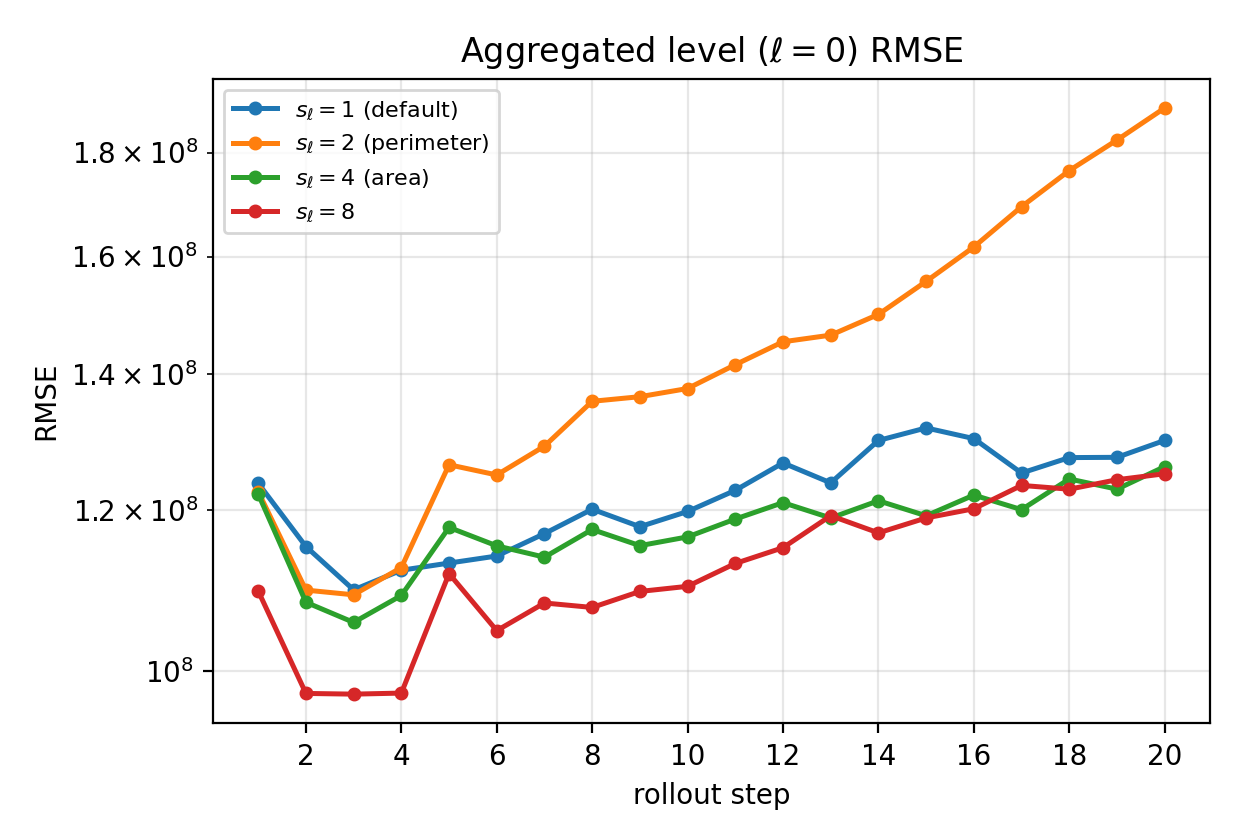}
\caption{Coarsest ($\ell\!=\!0$) RMSE.}
\end{subfigure}

\vspace{0.4em}

\begin{subfigure}{0.48\linewidth}
\includegraphics[width=\linewidth]{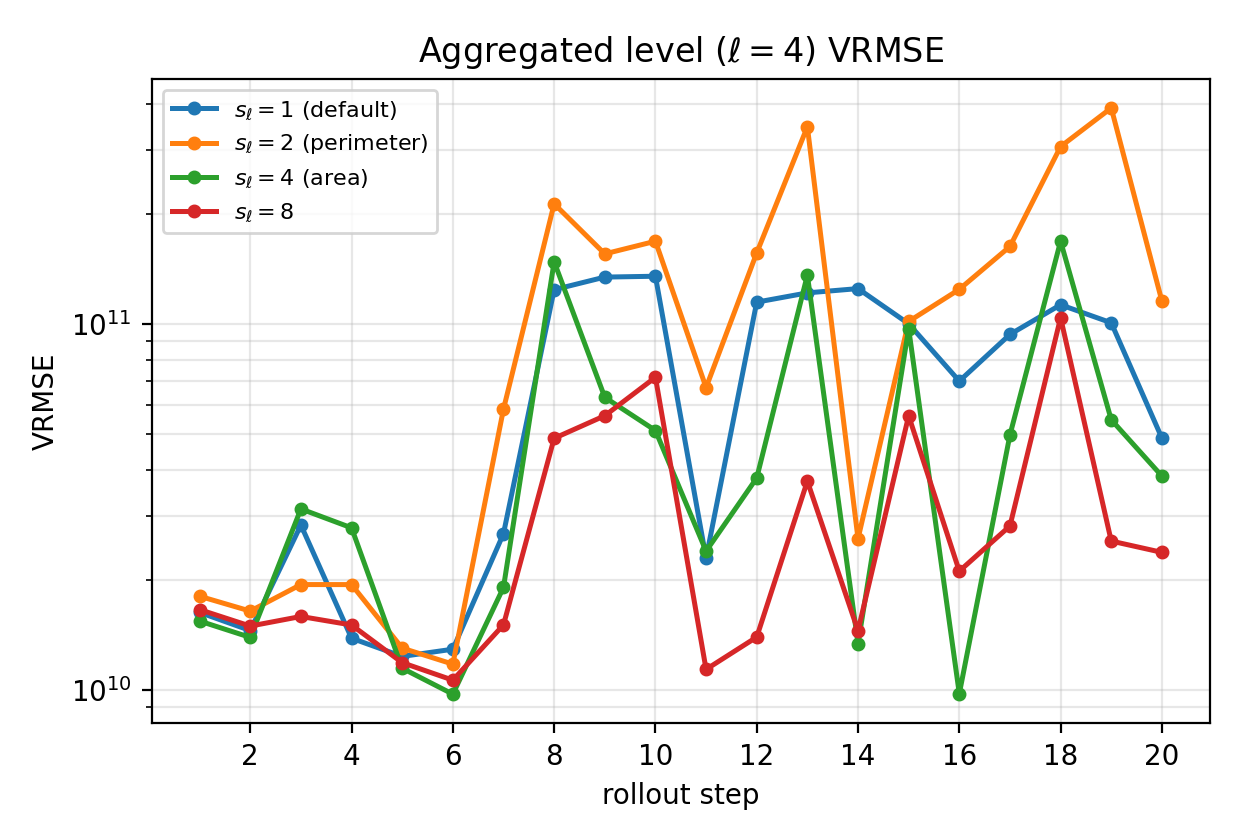}
\caption{Finest ($\ell\!=\!4$) VRMSE.}
\end{subfigure}\hfill
\begin{subfigure}{0.48\linewidth}
\includegraphics[width=\linewidth]{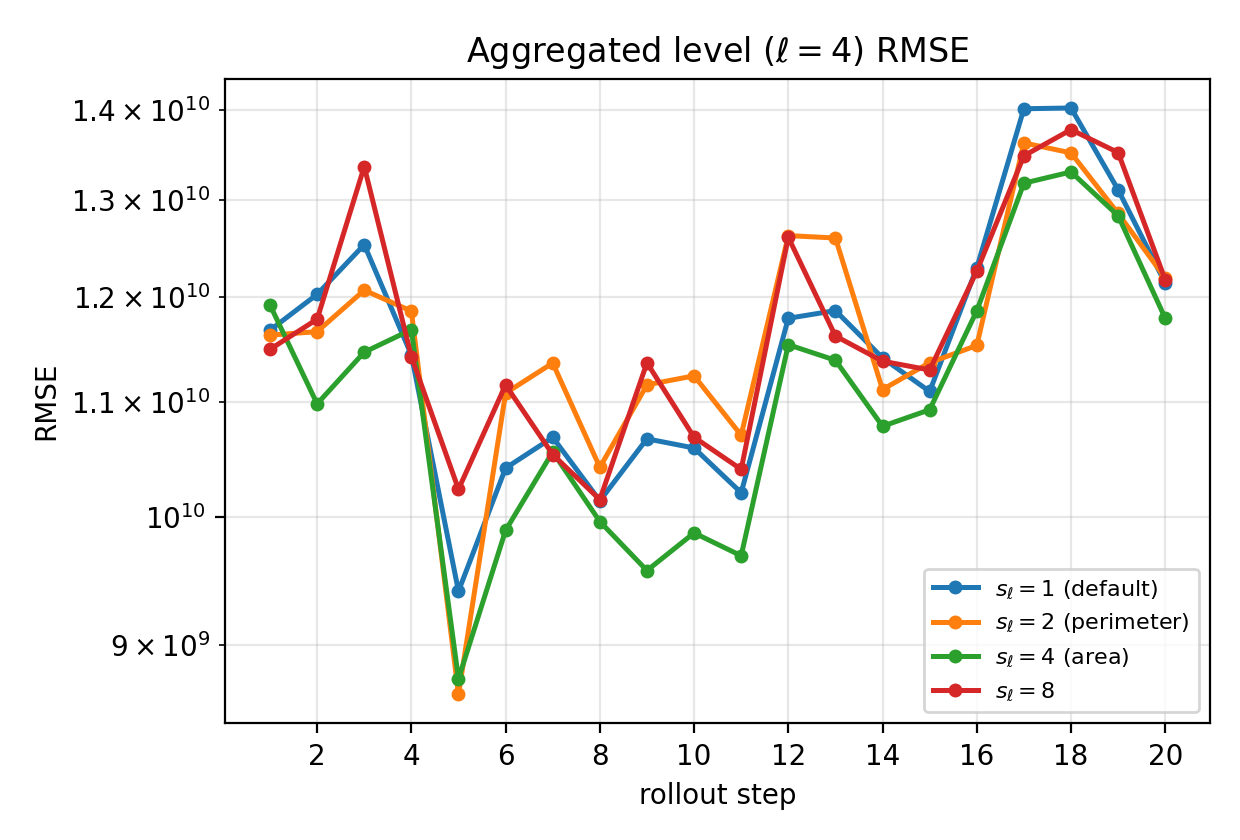}
\caption{Finest ($\ell\!=\!4$) RMSE.}
\end{subfigure}

\caption{\textbf{RoPE level-axis scale ablation on AMReX combustion.} WAMRViT rollouts at $s_\ell \in \{1,2,4,8\}$. Top row: coarsest level ($\ell=0$). Bottom row: finest level ($\ell=4$). Full-field materialization is omitted.}
\label{fig:app-amrex-level-scale-ablation}
\end{figure}

We compare separately trained models at level-axis RoPE scales $s_\ell \in \{1, 2, 4, 8\}$: uniform-patch WAMRViT on PLI, TRL, and AMReX combustion, and multi-scale WAMRViT on PLI. Topology resampling is disabled, and all other hyperparameters are held fixed within each variant (Table~\ref{tab:training-specs}). Appendix~\ref{app:rope-config} gives the geometric interpretation of these scales. First-step VRMSE is reported in Table~\ref{tab:level-scale-step1}. Figure~\ref{fig:app-pli-level-scale-ablation} compares the two PLI variants, and Figure~\ref{fig:app-amrex-level-scale-ablation} shows the uniform-patch AMReX sweep.

For uniform-patch WAMRViT on PLI (Figure~\ref{fig:app-pli-level-scale-uniform}), $s_\ell=2$ has the lowest first-step full-field VRMSE and $s_\ell=4$ has the lowest first-step finest-level VRMSE, but the differences from the default are small. Across the rollout, $s_\ell=4$ has the lowest mean full-field VRMSE, while the default $s_\ell=1$ has the lowest mean finest-level VRMSE. RMSE produces another ordering. On TRL, $s_\ell=8$ leads at full field and $s_\ell=2$ at the finest level in the one-step test (Table~\ref{tab:level-scale-step1}). These uniform-patch results do not support a monotone effect or a single best scale on PLI. The geometric labels remain useful interpretations of the scale, but the preferred value depends on the evaluation scope and horizon.

\paragraph{Multi-scale WAMRViT on PLI.}
We additionally train multi-scale WAMRViT at $s_\ell\in\{2,4,8\}$, retaining native-resolution patchification, level-conditioned AdaLN, and full-field loss, and compare with the existing $s_\ell=1$ model under the same fixed-seed protocol. Evaluation uses the same $761$ held-out windows over $20$ rollout steps. Table~\ref{tab:level-scale-step1} reports first-step VRMSE, and Figure~\ref{fig:app-pli-level-scale-multi} shows the rollout curves. Relative to $s_\ell=1$, scales $2$, $4$, and $8$ increase rollout-mean full-field VRMSE by $2.1\%$, $2.7\%$, and $8.1\%$, and RMSE by $0.6\%$, $2.2\%$, and $5.6\%$, respectively. Scale $2$ nevertheless lowers first-step full-field RMSE by $1.7\%$, illustrating that a short-horizon improvement need not persist over rollout. The default therefore has the lowest rollout-mean full-field errors among the tested scales in this single-seed multi-scale sweep.

The different scale sensitivity of the two variants may partly reflect the additional level conditioning in multi-scale WAMRViT. Its AdaLN layers apply learned level-specific affine transformations throughout the backbone, providing an explicit level signal alongside RoPE and potentially changing sensitivity to the rotary level scale.

On AMReX, $s_\ell=4$ has the lowest first-step VRMSE at both coarsest and finest levels and the lowest rollout-mean coarsest-level VRMSE. Under RMSE, $s_\ell=8$ is lowest at the first step on both levels and in the coarsest-level rollout mean. At the finest level, $s_\ell=4$ also has the lowest rollout-mean RMSE, reducing mean VRMSE and RMSE by $28.6\%$ and $4.1\%$ relative to $s_\ell=1$. Both metrics improve in all seventeen fields when averaged over the rollout. Scale $s_\ell=8$ has the lowest rollout-mean finest-level VRMSE, a $57.2\%$ reduction, but its RMSE is $1.4\%$ higher than the default. Scale $s_\ell=2$ raises the corresponding means by $74.1\%$ and $0.7\%$, respectively. HRR dominates the channel-mean finest-level VRMSE and causes sharp step-to-step changes, so the per-field curves remain necessary for interpretation. Together with PLI and TRL, this sweep shows that level scale is configuration-, scope-, metric-, and horizon-dependent. No single geometric choice is uniformly preferred.


\subsection{Mesh-policy sensitivity on PLI}
\label{app:pli-mesh-sensitivity}

We evaluate uniform-patch and multi-scale WAMRViT with fixed checkpoints on the same held-out PLI trajectories, forecasting $20$ steps with regridding every $R=4$ steps. The default uses a neighborhood radius $\delta=2$, initial tolerance $0.01$, and runtime tolerances $[0.1,0.1,0.3]$ for the two velocity components and density. We vary $\delta\in\{1,3\}$ at fixed tolerances, or scale both initial and runtime tolerances by $0.5$ or $1.5$ at $\delta=2$. Each change applies to initial mesh construction and subsequent regridding. Model weights are unchanged.

Table~\ref{tab:pli-mesh-sensitivity} compares full-field predictions reconstructed onto the same finest physical grid against the original uniform-grid ground truth, rather than evaluating each model's own refined regions. Errors are averaged over the ten PLI fields and all $20$ rollout steps.

\begin{table}[H]
\centering
\caption{\textbf{PLI mesh-policy sensitivity.} Full-field rollout-mean errors and token counts for the fixed checkpoints. Token counts $\bar N$ are averaged over the same evaluation windows and rollout steps.}
\label{tab:pli-mesh-sensitivity}
\small
\setlength{\tabcolsep}{4pt}
\begin{tabular}{@{}lcccccc@{}}
\toprule
 & \multicolumn{3}{c}{WAMRViT (uniform)} & \multicolumn{3}{c}{WAMRViT (multi-scale)} \\
\cmidrule(lr){2-4}\cmidrule(lr){5-7}
Mesh policy & $\bar N$ & RMSE & VRMSE & $\bar N$ & RMSE & VRMSE \\
\midrule
 Default & 3,324 & \textbf{0.0403} & \textbf{0.1749} & 3,336 & \textbf{0.0306} & \textbf{0.1405} \\
 Radius $\delta=1$ & 2,717 & 0.0486 & 0.2238 & 2,739 & 0.0414 & 0.1950 \\
 Radius $\delta=3$ & 3,981 & 0.0427 & 0.1836 & 3,998 & 0.0329 & 0.1476 \\
 Tolerances $\times0.5$ & 3,745 & 0.0408 & 0.1780 & 3,717 & 0.0316 & 0.1435 \\
 Tolerances $\times1.5$ & 3,269 & \underline{0.0404} & \underline{0.1753} & 3,299 & \underline{0.0308} & \underline{0.1428} \\
 \bottomrule
\end{tabular}
\end{table}

The default has the lowest rollout-mean RMSE and VRMSE among the tested settings for both models. Both models were trained only with the default initial tolerance and neighborhood radius, so the degradation under these perturbations is consistent with a training-to-test mesh-distribution shift. Reducing $\delta$ to $1$ removes about $18\%$ of tokens but increases mean VRMSE by $28.0\%$ for uniform-patch WAMRViT and $38.8\%$ for the multi-scale variant. RMSE increases by $20.6\%$ and $35.2\%$, respectively. Increasing $\delta$ to $3$ adds about $20\%$ more tokens but also raises both errors. Thus, denser refinement alone does not guarantee improved forecasting with fixed weights.

Tolerance scaling changes either rollout-mean metric by at most $3.3\%$ across the two models, although individual fields can be more sensitive: halving the tolerances raises multi-scale case-material volume-fraction RMSE by $16.3\%$. Multi-scale WAMRViT retains lower channel-mean full-field RMSE and VRMSE than uniform-patch WAMRViT at every rollout step under each matched policy. The aggregate sensitivity is therefore stronger for the tested neighborhood changes than for tolerance scaling, without reversing the ordering of the two variants.

\begingroup
\raggedbottom
\setlength{\intextsep}{10pt}
\setlength{\abovecaptionskip}{6pt}
\setlength{\belowcaptionskip}{0pt}

\subsection{Per-field rollout curves on PLI}

The PLI panels of Figure~\ref{fig:rollout} aggregate across physical channels (mean over the ten PLI fields). The figures below show the same rollout broken out by field. All are at push-forward $K\!=\!1$, averaged over all evaluation windows. Adaptive tracks are solid and uniform-grid baselines dashed (uniform-grid predictions are block-mean-projected onto the adaptive quadtree for the per-level figures). For readability, we show four field groups: a velocity component ($U$-velocity), bulk density, summed material energy, and the arithmetic mean of the six per-channel volume-of-fluid error curves. The three scopes match the aggregated figures (full-field, coarsest, finest). Intermediate levels are omitted.

\begin{figure}[H]
\centering
\includegraphics[width=\linewidth]{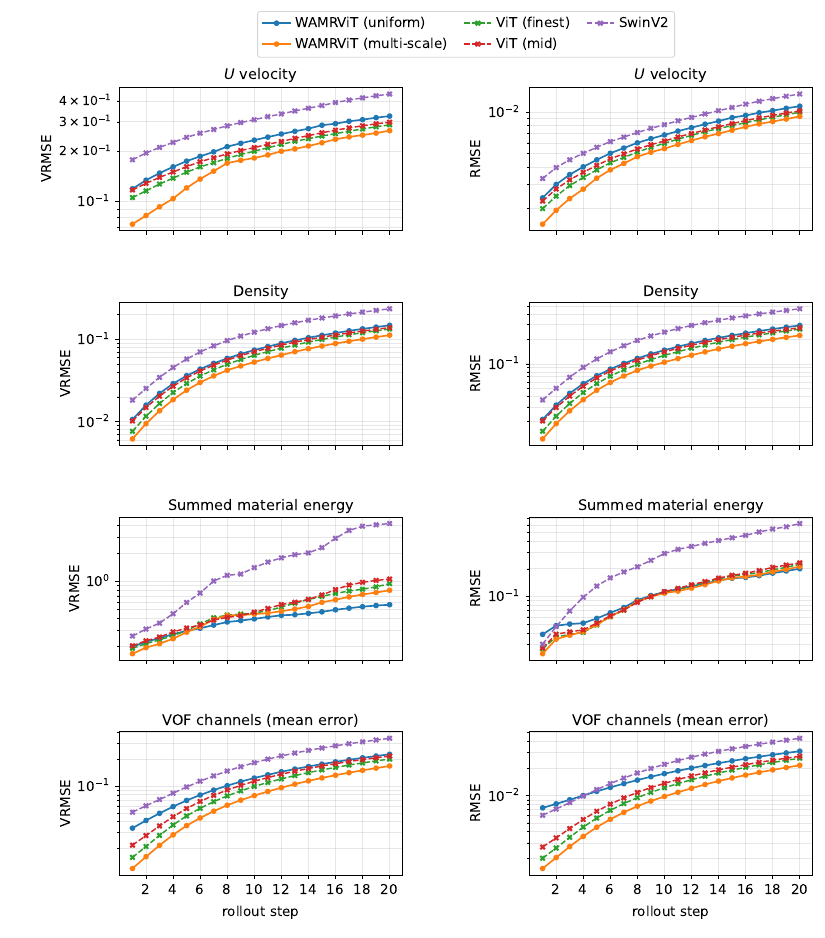}
\caption{\textbf{PLI full-field errors}, $K=1$. Rows show the four field groups described above. Columns show VRMSE (left) and RMSE (right).}
\label{fig:app-pli-uniform-metrics}
\label{fig:app-pli-uniform-vrmse}
\label{fig:app-pli-uniform-rmse}
\end{figure}

\clearpage
\begin{figure}[H]
\centering
\includegraphics[width=\linewidth]{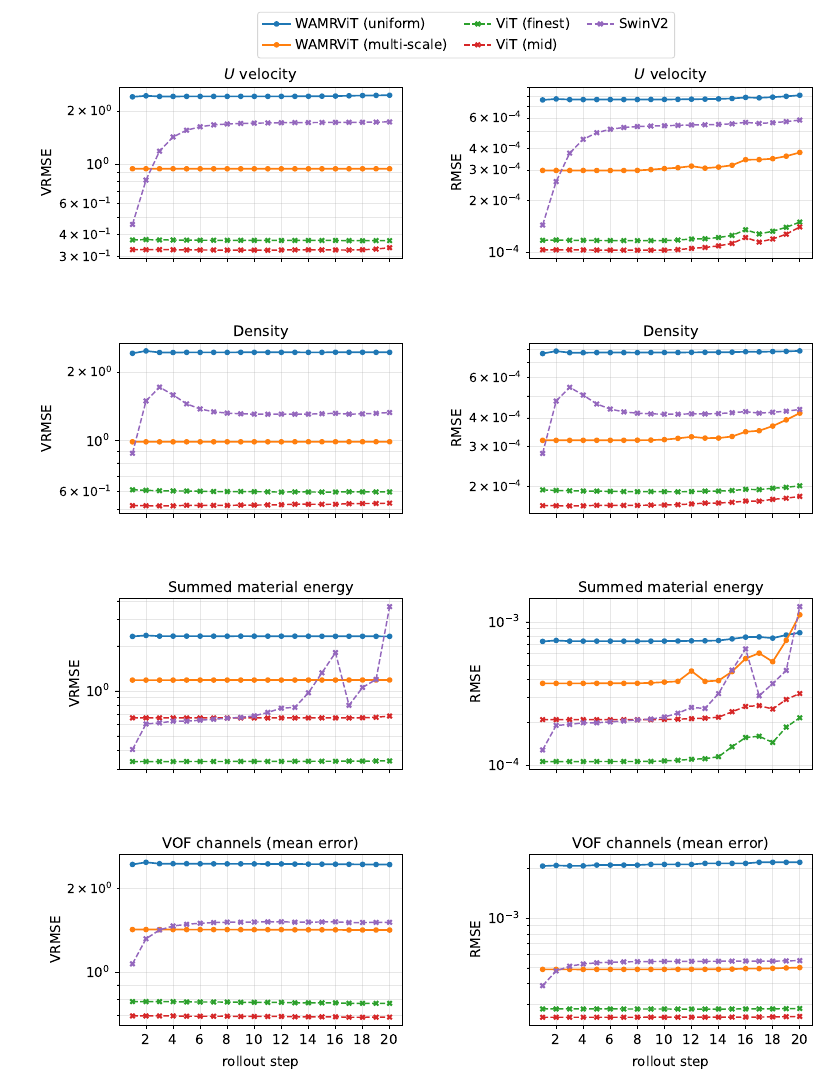}
\caption{\textbf{PLI coarsest-level errors} ($\ell=0$), $K=1$. Rows show the same four field groups. Columns show VRMSE (left) and RMSE (right).}
\label{fig:app-pli-l0-metrics}
\label{fig:app-pli-l0-vrmse}
\label{fig:app-pli-l0-rmse}
\end{figure}

\clearpage
\begin{figure}[H]
\centering
\includegraphics[width=\linewidth]{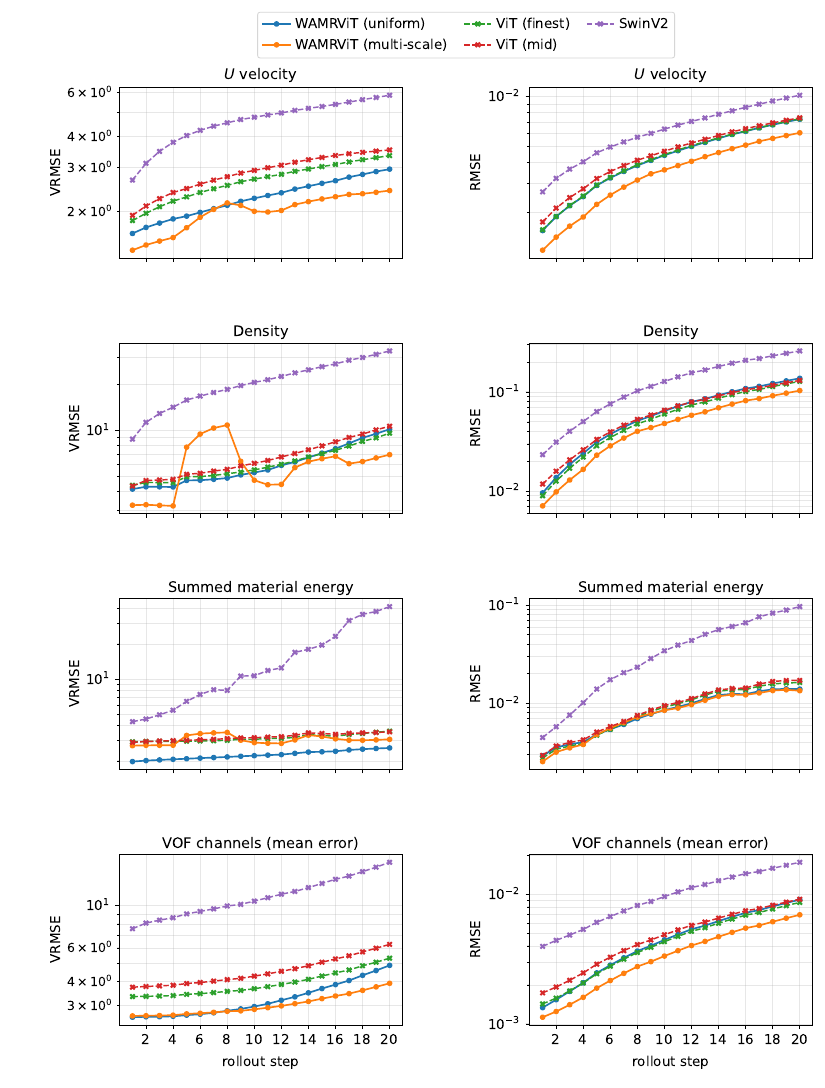}
\caption{\textbf{PLI finest-level errors} ($\ell=3$), $K=1$. Rows show the same four field groups. Columns show VRMSE (left) and RMSE (right).}
\label{fig:app-pli-l3-metrics}
\label{fig:app-pli-l3-vrmse}
\label{fig:app-pli-l3-rmse}
\end{figure}

\clearpage
\subsection{Aggregated rollout curves on AMReX combustion}
\label{app:perfield-amrex}

All panels use checkpoints trained at $K=4$, with the three-way legend of the AMReX panels in Figure~\ref{fig:rollout}: WAMRViT (uniform), ViT (mid), and SwinV2. The $p=32$ regular controls use the feasible intermediate projection rather than a full-domain uniform-grid representation at the native finest resolution. Intermediate levels $\ell\in\{1,2,3\}$ are omitted for readability.

\begin{figure}[H]
\centering
\includegraphics[width=\linewidth]{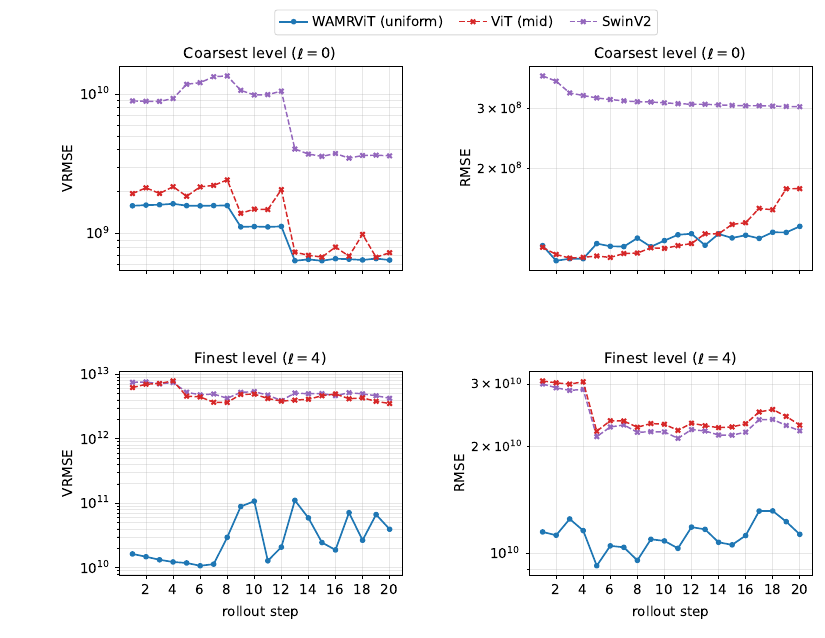}
\caption{\textbf{Aggregated AMReX combustion errors}, $K=4$. Rows show the coarsest ($\ell=0$) and finest ($\ell=4$) levels. Columns show VRMSE (left) and RMSE (right), averaged over the $17$ channels. As on PLI, the regular-grid predictions are block-mean-downsampled onto the adaptive coarsest-level leaves and can benefit from error-cancelling averages at this scope. The finest row corresponds to the AMReX panels of Figure~\ref{fig:rollout}.}
\label{fig:app-amrex-agg-metrics}
\label{fig:app-amrex-agg-coarse}
\label{fig:app-amrex-agg-finest}
\end{figure}

\clearpage
\subsection{Per-field rollout curves on AMReX combustion}
\label{app:perfield-amrex-fields}

The AMReX panels of Figure~\ref{fig:rollout} aggregate across seventeen combustion channels. The figures below separate six representative fields: heat-release rate (HRR), temperature $T$, Mach number, pressure, density $\rho$, and total energy density $\rho E$. They use the same WAMRViT, ViT-mid, and SwinV2 legend and the coarsest ($\ell=0$) and finest ($\ell=4$) scopes. Density-weighted species mass fractions are omitted for readability. All panels use checkpoints trained at $K=4$.

\begin{figure}[H]
\centering
\includegraphics[width=\linewidth]{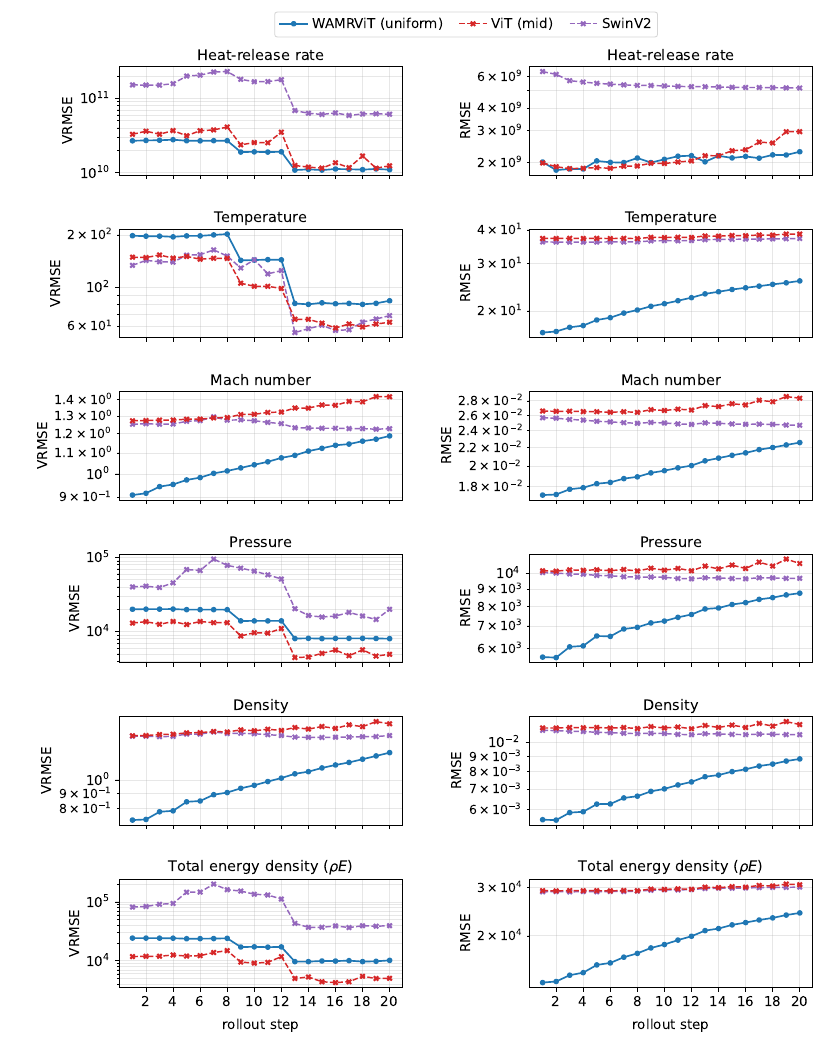}
\caption{\textbf{AMReX coarsest-level errors} ($\ell=0$), $K=4$. Each row corresponds to one of the six representative fields. Columns show VRMSE (left) and RMSE (right).}
\label{fig:app-amrex-l0-metrics}
\label{fig:app-amrex-l0-vrmse}
\label{fig:app-amrex-l0-rmse}
\end{figure}

\clearpage
\begin{figure}[H]
\centering
\includegraphics[width=\linewidth]{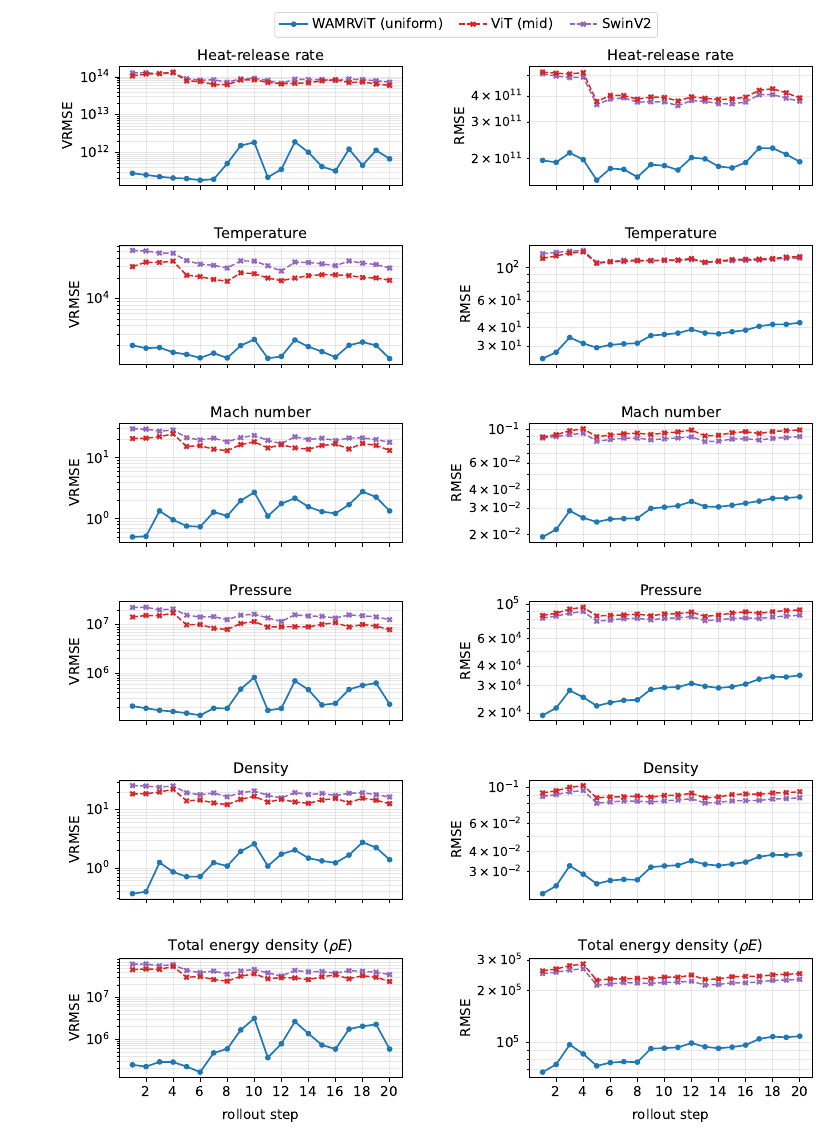}
\caption{\textbf{AMReX finest-level errors} ($\ell=4$), $K=4$. Rows show the same six fields. Columns show VRMSE (left) and RMSE (right).}
\label{fig:app-amrex-l4-metrics}
\label{fig:app-amrex-l4-vrmse}
\label{fig:app-amrex-l4-rmse}
\end{figure}

\clearpage
\endgroup

\section{Extended Related Work}
\label{app:related-work}

\paragraph{Adaptive patching for ViT-based PDE surrogates.}
AMR-Transformer \citep{xu2025amrtransformer} decides splits via four velocity-field criteria. Two image-domain methods use hand-crafted scores: the Adaptive Patch Framework (APF) \citep{zhang2024apf}, for high-resolution segmentation, uses a Canny edge detector, and APT \citep{choudhury2025apt} scores patches by entropy. MATEY \citep{zhang2024matey} is the closest in spirit: it is explicitly AMR-inspired, computes a per-patch variance, refines patches above a global threshold to a single finer size, uses two resolution levels in its reported experiments, positions itself as a multiscale adaptive transformer for spatiotemporal PDE forecasting, and handles variable token counts via padding masks. Iterative Adaptive Patchification \citep[IAP;][]{wang2026iap} fixes a target token count and iteratively splits the top-scoring patches under a Sobel-based gradient score until the budget is reached. Overtone \citep{mukhopadhyay2025controllable} does not choose where to refine. It trains a single ViT to run at any of a discrete patch-size set $\{4, 8, 16\}$ on uniform-grid inputs via convolutional kernel/stride modulation, cycling sizes across rollout steps to disrupt grid-aligned artifacts of fixed-patch tokenization. The partition itself remains spatially uniform within any forward pass and the per-step size is set externally. These approaches motivate adaptive patching on uniform-grid data. WAMRViT combines four capabilities: (i)~tokens carry an explicit 2:1-balanced quadtree hierarchy with level-aware 3D RoPE and mean-conservative coarsening, in contrast to recursive trees without a balance constraint (AMR-Transformer, APF, IAP) or MATEY's evaluated two-level partition; (ii)~variable token counts are handled by micro-batch-one training with data-parallel replication (Section~\ref{sec:training}) rather than by padding masks or a fixed token budget; (iii)~we evaluate changing-topology $20$-step rollouts on PLI and AMReX; and (iv)~we forecast directly on native multi-level AMReX data. To our knowledge, the novelty is the integration of variable-token prediction, repeated balanced-topology updates, and state transfer across those updates in one ML-based AMR surrogate, rather than any one component in isolation.

\paragraph{Mesh-free neural surrogates.}
Mesh-free methods represent the solution as a coordinate MLP $u_\theta(x, t)$ trained against PDE residuals or a variational principle. Physics-Informed Neural Networks \citep{raissi2019pinn} and the Deep Galerkin Method \citep{sirignano2018dgm} are canonical, with the DeepXDE library \citep{lu2021deepxde} consolidating these formulations and introducing residual-based adaptive refinement that densifies collocation points where the PDE residual is large. This concentrates compute in spirit similar to our wavelet-driven token refinement, but applied to where the optimizer evaluates the loss rather than to where the model places tokens. The methodological gap with our setting is consequential. Mesh-free methods typically solve a separate optimization per initial condition, amortizing only over conditions supplied as network inputs, and the long autoregressive rollouts we report are uncommon in this family because temporal causality must be enforced through the loss rather than through model architecture, so per-instance training cost compounds with rollout horizon. WAMRViT instead learns a single autoregressive operator amortized across trajectories, concentrates compute via adaptive tokenization rather than collocation density, and maintains a moving topology over the rollout via inference-time regridding.

\paragraph{Neural operators and PDE ViTs.}
FNO \citep{li2021fno}, DeepONet \citep{lu2021deeponet}, UPT \citep{alkin2024upt}, and CViT \citep{wang2025cvit} provide the broader context in which our tokenization sits. FNO and CViT consume a single uniform-resolution input field at inference time, with any hierarchical or multi-resolution structure internal to the operator (e.g.\ in the kernel representation of FNO) rather than expressed at the discretization. CViT encodes gridded inputs using fixed spatial patches, while its cross-attention decoder predicts at queried coordinates. Both WAMRViT variants use a spatially adaptive partition whose leaf areas can differ by factors of up to $4^L$ within one sample ($L=3$ on PLI and $L=4$ for uniform-patch AMReX; Table~\ref{tab:training-specs}). The transformer processes these leaves as one sequence under level-aware 3D RoPE rather than a multi-resolution cascade. The variant named ``multi-scale'' additionally retains each leaf's full source-resolution raster from a uniform-grid source. Graph-based and point-cloud operators such as UPT reach discretization-level flexibility on unstructured meshes. We instead use a 2:1-balanced quadtree, constructed from uniform-grid fields or initialized from a simulator hierarchy and subsequently adjusted as described in Appendix~\ref{app:amrex}. This structure provides explicit face-neighbor relations and mean-conservative merging of sibling rasters. A separate line of work embeds learned operators inside classical FE solvers rather than replacing them. NOEM \citep{ouyang2026noem}, building on earlier FE-NO coupling \citep{yin2022interfacing}, pre-trains neural operators to approximate boundary-to-solution maps on user-chosen subdomains and reconciles FE and NO degrees of freedom by energy minimization at inference. These methods use static partitions, and NOEM targets steady variational problems, in contrast to our dynamic wavelet refinement over autoregressive rollouts.

\section{Future work}
\label{app:future-work}

Throughput in the variable-$N$ regime admits several concrete follow-ups: padded micro-batching with zero-attention masks (in the spirit of MATEY~\citep{zhang2024matey}), length-sorted bucketing, block-diagonal flex-attention packing, and dynamic-shape GPU kernels each address a specific bottleneck. The current backend computes topology updates with packed, parallel CPU kernels and, for AMReX, applies the resulting split/merge records to leaf values kept on the GPU (Section~\ref{sec:rollout}). Moving neighbor lookup and balancing onto the GPU would remove the remaining synchronization and topology-transfer cost. The topology-cache machinery of Appendix~\ref{app:train-tokenize} composes with these changes. A native multi-scale objective that avoids materializing the finest uniform grid would also make the multi-scale variant applicable to simulator-native AMR data (Appendix~\ref{app:walltime}). A learned, multi-field generalization of the current non-learned wavelet score could also replace the global $\varepsilon$ and per-channel union with a content-aware detector. Its accuracy or token reduction would need to outweigh added inference and CPU--GPU synchronization costs.

WAMRViT factors cleanly into an adaptive quadtree tokenizer and a transformer backbone joined by a standard token interface, so the backbone is interchangeable without altering the tokenizer or the inference-time regrid. In large-patch regimes such as AMReX combustion ($p_h{=}p_w{=}32$), where the convolutional patch embedder carries a disproportionate share of the model's representational capacity, two spectral substitutions warrant investigation: an FNO-style spectral patch embedder~\citep{li2021fno} in place of the convolutional one, and AFNO-style Fourier token mixing~\citep{guibas2021adaptive} in place of self-attention. Either substitution leaves the tokenizer and rollout machinery untouched, isolating the architectural change. A further extension is a controllable variant of WAMRViT in the spirit of \citet{mukhopadhyay2025controllable}, exposing the wavelet refinement tolerance as an explicit compute-budget control distinct from the patch-size selector they study at inference.

The level-scale ablation (Appendix~\ref{app:level-scale-ablation}) shows neither a monotone trend nor a dataset-independent optimum, suggesting that a single hand-set $s_\ell$ is a restrictive parametrization of refinement level. Learning the level scale jointly with the model, or encoding level differences through a learned scale-aware attention bias, could remove the need to tune it per dataset.

\end{document}